\documentclass[prodmode,acmtocl]{acmsmall} 
\pdfoutput=1

\usepackage{comment}
\usepackage{amsmath}
\usepackage{amsfonts}
\usepackage{amssymb}
\usepackage{longtable}
\usepackage{caption}
\usepackage[subrefformat=parens,labelformat=simple]{subfig}

\usepackage{url}
\usepackage{multirow}
\usepackage{cases}
\usepackage{tikz}
\usepackage{pgfplots}
\usepackage{pgfplotstable}
\usepackage{filecontents}
\usetikzlibrary{plotmarks}
\usetikzlibrary{patterns}
\ifdefined\TIKZEXTERNALIZE
  \usepgfplotslibrary{external} 
  \tikzsetexternalprefix{tikzfigures/}
\fi

\def\folhappens{\mathit{happens}}
\def\folholdsAt{\mathit{holdsAt}}

\def\folinitiatedAt{\mathit{initiatedAt}}
\def\folterminatedAt{\mathit{terminatedAt}}
\def\folclose{\mathit{close}}

\def\folOM{\mathit{orientationMove}}

\def\CEConditions{\mathit{Conditions[X,\, Y,\, T]}}

\newcommand{\pHoldsAt}[2]{\folholdsAt(\mathit{#1},\, #2)}
\newcommand{\pHappens}[2]{\folhappens(\mathit{#1},\, #2)}
\newcommand{\pInitiatedAt}[2]{\folinitiatedAt(\mathit{#1},\, #2)}
\newcommand{\pTerminatedAt}[2]{\folterminatedAt(\mathit{#1},\, #2)}

\newcommand{\pClose}[4]{\folclose(\mathit{#1},\, \mathit{#2},\, #3,\, #4)}
\newcommand{\pOrientationMove}[3]{\folOM(\mathit{#1},\, \mathit{#2},\, #3)}

\def\PEC{\mathit{MLN}{\textendash}\mathit{EC}}

\def\ECLP{\mathit{EC_{crisp}}}
\def\CRF{\mathit{l}{\textendash}\mathit{CRF}}

\def\moving{\textit{moving}}
\def\meeting{\textit{meeting}}

\newcommand{\pec}[1]{\PEC_{\mathit{#1}}}
\def\MLNEC{$\PEC$}
\def\LCRF{$\CRF$}
\def\CREC{$\ECLP$}

\def\HI{\pec{HI}}
\def\SI{\pec{SI}}

\def\SIEq{\pec{SI^{eq}}} 
\def\SIEqWeak{\pec{SI^{eq}_{weak}}}

\def\SIT{\pec{SI^{h}}}
\def\SITWeak{\pec{SI^{h}_{weak}}}
\def\SII{\pec{SI^{\neg h}}}
\def\SIIWeak{\pec{SI^{\neg h}_{weak}}}

\newenvironment{mysplit}%
  {\arraycolsep 0pt \begin{array}{l}}%
  {\end{array}}

\DeclareMathOperator*{\argmax}{\arg\!\max}

\title{Probabilistic Event Calculus for Event Recognition}

\author{ANASTASIOS SKARLATIDIS$^{1,2}$, GEORGIOS PALIOURAS$^{1}$, ALEXANDER ARTIKIS$^{1}$ {and} GEORGE A. VOUROS$^{2}$
\affil{$^1$Institute of Informatics and Telecommunications, NCSR ``Demokritos'', \\
$^2$Department of Digital Systems, University of Piraeus}}

\begin{abstract}
Symbolic event recognition systems have been successfully applied to a variety of application domains, extracting useful information in the form of events, allowing experts or other systems to monitor and respond when significant events are recognised. In a typical event recognition application, however, these systems often have to deal with a significant amount of uncertainty. In this paper, we address the issue of uncertainty in logic-based event recognition by extending the Event Calculus with probabilistic reasoning.  Markov Logic Networks are a natural candidate for our logic-based formalism. However, the temporal semantics of the Event Calculus introduce a number of challenges for the proposed model. We show how and under what assumptions we can overcome these problems. Additionally, we study how probabilistic modelling changes the behaviour of the formalism, affecting its key property, the \textit{inertia} of fluents. Furthermore, we demonstrate the advantages of the probabilistic Event Calculus through examples and experiments in the domain of activity recognition, using a publicly available dataset for video surveillance.
\end{abstract}




\acmformat{Anastasios Skarlatidis, Georgios Paliouras, Alexander Artikis and George A. Vouros. 2013. Probabilistic Event Calculus for Event Recognition.}

\begin{document} \raggedbottom

\begin{bottomstuff}
This work has been partially funded by the European Commission, in the context of the PRONTO project (FP7-ICT 231738). This paper is a significantly updated and extended version of \citeN{anskarl2011RML}. 
	
Author's addresses: A. Skarlatidis {and} G. Paliouras {and} A. Artikis, Institute of Informatics and Telecommunications, NCSR ``Demokritos'', Athens 15310, Greece 
G. Vouros, Department of Digital Systems, University of Piraeus, Piraeus 18534, Greece
\end{bottomstuff}

\maketitle

\section{Introduction} \label{sec:Intro}

%
%
Symbolic event recognition systems have received attention in a variety of application domains, such as health care monitoring, public transport management, telecommunication network monitoring and activity recognition \cite{luckham2002power,etzion2010event,LuckhamEPB,LBER_KER12}. The aim of these systems is to extract useful information, in the form of events, by processing time-evolving data that comes from various sources (e.g.~various types of sensor, surveillance cameras, network activity logs, etc.). The extracted information can be exploited by other systems or human experts, in order to monitor an environment and respond to the occurrence of significant events. The input to a symbolic event recognition system consists of a stream of time-stamped symbols, called simple, derived events (SDEs). Consider, for example, a video tracking system detecting that someone is walking for a sequence of video frames. Based on such time-stamped input SDE observations, the symbolic event recognition system recognises composite events (CEs) of interest. For instance, that some people have started to move together. The recognition of a CE may be associated with the occurrence of various SDEs and other CEs involving multiple entities, e.g.~people, vehicles, etc. CEs therefore, are relational structures over other sub-events, either CEs or SDEs. 

Statistical approaches, e.g.~probabilistic graphical models, employ Machine Learning techniques, in order to learn the situations under which CEs must be recognised from annotated examples. Such methods are data driven and they are completely dependent upon the examples of the training set. On the other hand, background knowledge (e.g.~knowledge expressed by domain experts) may describe situations that do not appear in the training data or are difficult to be collected and annotated. The majority of statistical approaches employ models with limited capabilities in expressing relation among entities. As a result, the definition of CEs and the use of background knowledge is very hard. Logic-based approaches, such as the Event Calculus \cite{kowalski1986logic,artikis2010logic}, can naturally and compactly represent relational CE structures. Based on their formal and declarative semantics, they provide solutions that allow one to easily incorporate and exploit background knowledge. In contrast to statistical methods, however, they cannot handle uncertainty which naturally exists in many real-world event recognition applications.

%
%
Event recognition systems often have to deal with data that involves a significant amount of uncertainty [\citeNP{ShetNRD07};~\citeNP{artikis2010logic};~\citeNP[Section 11.2]{etzion2010event}; \citeNP{gal2011event}]: (a) Low-level detection systems often cannot detect all SDEs required for CE recognition, e.g.~due to a limited number of sensing sources. Logical definitions of CEs, therefore, have to be constructed upon a limited and often insufficient dictionary of SDEs. (b) Partial and noisy observations result in incomplete and erroneous SDE streams. For example, a sensor may fail for some period of time and stop sending  information, interrupting the detection of a SDE. Similarly, noise in the signal transmission may distort the observed values. (c) Inconsistencies between SDE streams and CE annotations introduce further uncertainty. When Machine Learning algorithms are used, similar patterns of SDEs may be inconsistently annotated. As a result, CE definitions and background knowledge, either learnt from data or derived by domain experts not strictly follow the annotation. Under such situations of uncertainty, the  performance of an Event Recognition system may be seriously compromised.

%
%
In the presence of some of the aforementioned types of uncertainty, e.g.~partial SDE streams and inconsistent annotations, the CE definitions of a logic-based Event Recognition system cannot capture perfectly the conditions under which a CE occurs. Based on such imperfect CE definitions, the aim of this work is to recognise CEs of interest under uncertainty. In particular, we propose a probabilistic version of the Event Calculus that employs Markov Logic Networks (MLNs) \cite{Domingos2009markov}. The Event Calculus is a formalism for representing events and their effects. Beyond the advantages stemming from the fact that it is a logic-based formalism with clear semantics, one of the most interesting properties of the Event Calculus is that it handles the persistence of CEs with domain-independent axioms. On the other hand, MLNs are a generic statistical relational framework that combines the expressivity of first-order logic with the formal probabilistic properties of undirected graphical models --- see \citeN{BrazAR08}, \citeN{RaedtK10} and \citeN{blockeel2011statistical} for surveys on logic-based relational probabilistic models. By combining the Event Calculus with MLNs, we present a principled and powerful probabilistic logic-based method for event recognition.

\vspace{1cm}

In particular the contributions of this work are the following:  
\begin{itemize}

\item A probabilistic version of the Event Calculus for the task of event recognition. The method inherits the domain-independent properties of the Event Calculus and supports the probabilistic recognition of CEs with imperfect definitions. 

 \item Efficient representation of the Event Calculus axioms and CE definitions in MLNs. The method employs a discrete variant of the Event Calculus and translates the entire knowledge base into compact Markov networks, in order to avoid the combinatorial explosion caused by the expressivity of the logical formalism.
 
 \item A thorough study of the behaviour of CE persistence. Under different conditions of interest, the method can model various types of CE persistence, ranging from deterministic to purely probabilistic.

\end{itemize}

%
%
To demonstrate the benefits of the proposed approach, the method is evaluated in the real-life event recognition task of human activity recognition. The method is compared against its crisp predecessor, as well as a purely statistical model based on linear-chain Conditional Random Fields. The definitions of CEs are domain-dependent rules that are given by humans and expressed using the language of the Event Calculus. The method processes the rules in the knowledge base and produces Markov networks of manageable size and complexity. Each rule can be associated with a weight value, indicating a degree of confidence in it. Weights are automatically estimated from a training set of examples.  The input to the recognition system is a sequence of SDEs expressed as a narrative of ground predicates. Probabilistic inference is used to recognise CEs. 

%
%
The remainder of the paper is organised as follows. First, in Section \ref{sec:example}, we present the target activity recognition application, in order to introduce a running example for the rest of the paper. In Section \ref{sec:EventCalculus} we present the axiomisation of the proposed probabilistic version of the Event Calculus for the task of Event Recognition. In Section \ref{sec:MLN} we briefly present Markov Logic Networks. Then, in Section \ref{sec:CompactMN} we present representational simplifications and transformations that we employ, in order to produce compact ground Markov Networks. In Section \ref{sec:inertia}, we study the behaviour of the probabilistic formalism. In Section \ref{sec:experiments} we demonstrate the benefits of probabilistic modelling, through experiments in the real-life activity recognition application. Finally in Sections  \ref{sec:related_work} and \ref{sec:conclusions}, we present related work and outline directions for further research.

\section{Running Example: Activity Recognition} \label{sec:example}

To demonstrate our method, we apply it to video surveillance in public spaces using the publicly available benchmark dataset of the CAVIAR project\footnote{\url{http://homepages.inf.ed.ac.uk/rbf/CAVIARDATA1}}. The aim is to recognise activities that take place between multiple persons, by exploiting information about observed individual activities. The dataset comprises $28$ surveillance videos, where each frame is annotated by human experts from the CAVIAR team on two levels. The first level contains simple, derived events (SDEs) that concern activities of individual persons or the state of objects. The second level contains composite event (CE) annotations, describing the activities between multiple persons and/or objects, e.g.~people \meeting\ and \moving\ together, \textit{leaving an object}, etc. In this paper, we focus on the recognition of the \meeting\ and \moving\ CEs, for which the dataset contains a sufficient amount of training examples.

%
%
The input to our method is a stream of SDEs, representing people \textit{walking}, \textit{running}, staying \textit{active}, or \textit{inactive}. We do not process the raw video data in order to recognise such individual activities. Instead we use the SDEs provided in the CAVIAR dataset. Thus, the input stream of SDEs is represented by a narrative of time-stamped predicates. The first and the last time that a person or an object is tracked are represented by the SDEs \textit{enter} and \textit{exit}. Additionally, the coordinates of tracked persons or objects are preprocessed and represented by predicates that express qualitative spatial relations, e.g.~two persons being relatively close to each other. Examples of these predicates are presented in the following sections.

The definitions of the \meeting\ and \moving\ CEs in the Event Calculus were developed in \cite{artikis2009behaviour}. These definitions take the form of common-sense rules and describe the conditions under which a CE starts or ends. For example, when two persons are \text{walking} together with the same orientation, then \moving\ starts being recognised. Similarly, when the same persons walk away from each other, then \moving\ stops being recognised.

Based on the input stream of SDEs and the CE definitions, the aim is to recognise instances of the two CEs of interest. The CE definitions are imperfect, since under the presence of uncertainty they cannot capture perfectly all the conditions under which a CE occurs. Furthermore, the definitions are derived from experts and may not strictly follow the annotation. As a result, CE definitions do not lead to perfect recognition of the CEs. 

\section{The Event Calculus} \label{sec:EventCalculus}

The Event Calculus, originally introduced by \citeN{kowalski1986logic}, is a many-sorted first-order predicate calculus for reasoning about events and their effects.  A number of different dialects have been proposed using either logic programming or classical logic \cite{shanahan1999event,miller2002some,mueller2008event}. Most Event Calculus dialects share the same ontology and core domain-independent axioms. The ontology consists of \textit{time-points}, \textit{events} and \textit{fluents}. The underlying time model is often linear and may represent time-points as real or integer numbers. A \textit{fluent} is a property whose value may change over time. When an \textit{event} occurs it may change the value of a fluent. The core domain-independent axioms define whether a fluent holds or not at a specific time-point. Moreover, the axioms incorporate the common sense \emph{law of inertia}, according to which fluents persist over time, unless they are affected by the occurrence of some event. 

We base our model on an axiomisation of a discrete version of the Event Calculus in classical first-order logic. The Discrete Event Calculus (DEC) has been proved to be logically equivalent to the Event Calculus when the domain of time-points is limited to integers \cite{mueller2008event}. DEC\footnote{\url{http://decreasoner.sourceforge.net}} is composed of  twelve domain-independent axioms. However, for the task of event recognition, we focus only on the domain-independent axioms that determine the influence of events to fluents and the inertia of fluents. We do not consider the predicates and axioms stating when a fluent is not subject to inertia ($\mathit{releases}$ and $\mathit{releasedAt}$), as well as its discrete change based on some domain-specific mathematical function ($\mathit{trajectory}$ and $\mathit{antiTrajectory}$). Furthermore, we adopt a similar representation to that of \citeN{artikis2010logic}, where predicates stating the initiation and termination of fluents are only defined in terms of fluents and time-points. Table \ref{tbl:prob-ec} summarises the elements of the proposed Event Calculus (\MLNEC). Variables (starting with an upper-case letter) are assumed to be universally quantified unless otherwise indicated. Predicates, functions and constants start with a lower-case letter.  

\begin{table}[h]
\tbl{The \MLNEC\ predicates \label{tbl:prob-ec}}{%
\begin{tabular}{|l|l|}
\hline
\multicolumn{1}{|c|}{Predicate} & \multicolumn{1}{c|}{Meaning} \\\hline
$\pHappens{E}{T}$ & Event $\mathit{E}$ occurs at time-point $T$  \\\hline
$\pHoldsAt{F}{T}$ & Fluent $F$ holds at time-point $T$ \\\hline
$\pInitiatedAt{F}{T}$ & Fluent $F$ is initiated at time-point $T$ \\\hline
$\pTerminatedAt{F}{T}$ & Fluent $F$ is terminated at time-point $T$ \\\hline
\end{tabular}
} 
\end{table}

The $\PEC$ axioms that determine when a fluent holds are defined as follows:
\begin{align}
& \begin{mysplit}
\label{axiom:fol_dec7a_holdsAt}
 \pHoldsAt{F}{T{+}1} \Leftarrow\\
 \qquad \qquad  \pInitiatedAt{F}{T}
\end{mysplit} \\
%
& \begin{mysplit}
\label{axiom:fol_dec7a_holdsAt_inertia}
\pHoldsAt{F}{T{+}1} \Leftarrow\ \\
  \qquad \qquad \pHoldsAt{F}{T}\ \wedge \\
  \qquad \qquad \neg \pTerminatedAt{F}{T}
\end{mysplit}
\end{align}
Axiom \eqref{axiom:fol_dec7a_holdsAt} defines that if a fluent $F$ is initiated at time $T$, then it holds at the next time-point. Axiom \eqref{axiom:fol_dec7a_holdsAt_inertia} specifies that a fluent continues to hold unless it is terminated.

The axioms that determine when a fluent does not hold are defined similarly:
\begin{align}
& \begin{mysplit}
\label{axiom:fol_dec7a_not_holdsAt}
 \neg \pHoldsAt{F}{T{+}1} \Leftarrow\\
 \qquad \qquad \qquad \pTerminatedAt{F}{T}
\end{mysplit} \\
%
& \begin{mysplit}
\label{axiom:fol_dec7a_not_holdsAt_inertia}
 \neg \pHoldsAt{F}{T{+}1} \Leftarrow\ \\
 \qquad \qquad \qquad \neg \pHoldsAt{F}{T}\ \wedge \\
 \qquad \qquad \qquad \neg \pInitiatedAt{F}{T}
\end{mysplit}
\end{align}
According to axiom \eqref{axiom:fol_dec7a_not_holdsAt}, if a fluent $F$ is terminated at time $T$ then it does not hold at the next time-point. Axiom \eqref{axiom:fol_dec7a_not_holdsAt_inertia} states that a fluent continues not to hold unless it is initiated.

The predicates $\folhappens$, $\folinitiatedAt$ and $\folterminatedAt$ are defined only in a domain-dependent manner. $\folhappens$ expresses the input evidence, determining the occurrence of a SDE at a specific time-point. A stream of observed SDEs, therefore, is represented in the \MLNEC\ as a narrative of ground $\folhappens$ predicates. As an example, consider the following fragment of a narrative:
\begin{align}
&\begin{mysplit} \nonumber \label{EC:Narrative_Example}
 \ldots\\
 \pHappens{walking(id_1)}{99} \\  
 \pHappens{walking(id_2)}{99} \\
 \pHappens{walking(id_1)}{100} \\  
 \pHappens{walking(id_2)}{100} \\
 \ldots \\
 \pHappens{active(id_1)}{500} \\  
 \pHappens{active(id_2)}{500} \\
 \ldots
 \end{mysplit}
\end{align}

\noindent According to the above narrative, it has been observed that two persons $\mathit{id_1}$ and $\mathit{id_2}$ are \textit{walking}, e.g.~at time-points $99$ and $100$, and later at time-point $500$ they are \textit{active}, e.g.~they are moving their arms but staying at the same position.

The predicates $\folinitiatedAt$ and $\folterminatedAt$ specify under which circumstances a fluent --- representing a CE --- is to be initiated or terminated at a specific time-point. The domain-dependent rules of the \MLNEC, i.e.~the initiation and/or termination of some $fluent_1$ over some domain-specific entities $X$ and $Y$ take the following general form:
\begin{equation}
\label{dsa:fol_general_form}
\begin{mysplit}
  \pInitiatedAt{fluent_{1}(X,Y)}{T} \Leftarrow \\
  \qquad \qquad \pHappens{event_{i}(X)}{T}\ \wedge\ \dots\ \wedge \\
  \qquad \qquad \pHoldsAt{fluent_{j}(X)}{T}\ \wedge\  \dots\ \wedge \\
  \qquad \qquad \CEConditions 
\vspace{0.5em} \\
  \pTerminatedAt{fluent_{1}(X,Y)}{T} \Leftarrow \\
  \qquad \qquad \pHappens{event_{k}(X)}{T}\ \wedge\ \dots\ \wedge \\
  \qquad \qquad \pHoldsAt{fluent_{l}(X)}{T}\ \, \wedge\  \dots\ \wedge \\
  \qquad \qquad \CEConditions \\
 \end{mysplit}
\end{equation}
\noindent In this work we consider finite domains of time-points, events and fluents, that are represented by the finite sets $\mathcal{T}$, $\mathcal{E}$ and $\mathcal{F}$, respectively. All individual entities that appear in a particular event recognition task, e.g.~persons, objects, etc., are represented by the constants of the finite set $\mathcal{O}$. 
$\CEConditions$ in \eqref{dsa:fol_general_form} is a set of predicates that introduce further constraints in the definition, referring to time $T \in \mathcal{T}$ and entities $X,\, Y \in \mathcal{O}$. The predicates $\folhappens$ and $\folholdsAt$, as well as those appearing in $\mathit{Conditions[X,Y,T]}$, may also be negated. The initiation and termination of a fluent can be defined by more than one rule, each capturing a different initiation and termination case. With the use of $\folhappens$ predicates, we can define a CE over SDE observations. Similarly, with the $\folholdsAt$ predicate we can define a CE over other CE, in order to create hierarchies of CE definitions. In both $\folinitiatedAt$ and $\folterminatedAt$ rules, the use of $\folhappens$, $\folholdsAt$ and $\CEConditions$ is optional and varies according to the requirements of the target event recognition application.

In our example application, for instance, the \moving\ activity of two persons is terminated when both of them are \textit{active}. This termination case can be represented using the following rule:
\begin{equation}
 \begin{mysplit} \label{EC:DSA_Example}
 \pTerminatedAt{moving(ID_1,\, ID_2)}{T} \Leftarrow\\
  \qquad \qquad \pHappens{active(ID_1)}{T} \ \wedge\\
  \qquad \qquad \pHappens{active(ID_2)}{T} 
 \end{mysplit}
\end{equation}

Based on a narrative of SDEs and a knowledge base composed of domain-dependent CE definitions (e.g.~rule \eqref{EC:DSA_Example}) and the domain-independent Event Calculus axioms, we can infer whether a fluent holds or not at any time-point. When a fluent holds at a specific time-point, then the corresponding CE is considered to be recognised. For example,  the \moving\ CE between persons $\mathit{id_1}$ and $\mathit{id_2}$ is recognised at time-point $100$ by inferring that $\pHoldsAt{moving(id_1,\, id_2)}{100}$ is \textit{True}. Similarly, the \moving\ CE for the same persons is not recognised at time-point $501$ by inferring that $\pHoldsAt{moving(id_1,\, id_2)}{501}$ is \textit{False}.

Consider the following definition of the \meeting\ CE between two persons in our running example.
\begin{align}
&\begin{mysplit} \label{dsa:init_meet1}
  \pInitiatedAt{meeting(ID_1,\, ID_2)}{T} \Leftarrow \\
  \qquad \qquad \qquad \pHappens{active(ID_1)}{T}\ \wedge \\
  \qquad \qquad \qquad \neg \pHappens{running(ID_2)}{T}\ \wedge \\
  \qquad \qquad \qquad \pClose{ID_1}{ID_2}{25}{T}
 \end{mysplit} \\
%
&\begin{mysplit}  \label{dsa:init_meet2}
  \pInitiatedAt{meeting(ID_1,\, ID_2)}{T} \Leftarrow \\
  \qquad \qquad \qquad \pHappens{inactive(ID_1)}{T}\ \wedge \\
  \qquad \qquad \qquad \neg \pHappens{running(ID_2)}{T}\ \wedge \\
  \qquad \qquad \qquad \neg \pHappens{active(ID_2)}{T}\ \wedge \\
  \qquad \qquad \qquad \pClose{ID_1}{ID_2}{25}{T}
 \end{mysplit} 
 \end{align} 
 
 \begin{align}
&\begin{mysplit} \label{dsa:term_meet1}
 \pTerminatedAt{meeting(ID_1,\, ID_2)}{T} \Leftarrow \qquad \\
  \qquad \qquad \qquad \pHappens{walking(ID_1)}{T}\ \wedge \\
  \qquad \qquad \qquad \neg \pClose{ID_1}{ID_2}{34}{T}
\end{mysplit} \\
%
&\begin{mysplit} \label{dsa:term_meet2}
 \pTerminatedAt{meeting(ID_1,\, ID_2)}{T} \Leftarrow \qquad  \\
  \qquad \qquad \qquad \pHappens{running(ID_1)}{T}
\end{mysplit} \\
%
&\begin{mysplit} \label{dsa:term_meet3}
 \pTerminatedAt{meeting(ID_1,\, ID_2)}{T} \Leftarrow \qquad  \\
  \qquad \qquad \qquad \pHappens{exit(ID_1)}{T}
\end{mysplit}
\end{align}
The predicate $\folclose$ expresses a spatial constraint stating that the distance between persons $\mathit{ID_1}$ and $\mathit{ID_2}$ at time $T$ must be below a specified threshold in pixels, e.g.~$25$ pixels. According to rules \eqref{dsa:init_meet1} and \eqref{dsa:init_meet2}, the \meeting\ activity is initiated when the people involved interact with each other, i.e.~at least one of them is active or inactive, the other is not running, and the measured distance between them is at most $25$ pixels. The \meeting\ CE is terminated either when people walk away from each other (rule \ref{dsa:term_meet1}), or someone is running (rule \ref{dsa:term_meet2}), or has exited the scene (rule \ref{dsa:term_meet3}).

The definition of the CE that people are \moving\ together is represented as follows:
\begin{align}
&\begin{mysplit} \label{dsa:init_move}
  \pInitiatedAt{moving(ID_1,\, ID_2)}{T} \Leftarrow \\
  \qquad \qquad \qquad \pHappens{walking(ID_1)}{T}\ \wedge \\
  \qquad \qquad \qquad \pHappens{walking(ID_2)}{T}\ \wedge \\
  \qquad \qquad \qquad \pOrientationMove{ID_1}{ID_2}{T}\ \wedge \\
  \qquad \qquad \qquad \pClose{ID_1}{ID_2}{34}{T} \\
 \end{mysplit} \\
%
&\begin{mysplit} \label{dsa:term_move1}
   \pTerminatedAt{moving(ID_1,\, ID_2)}{T} \Leftarrow \qquad \qquad \\
    \qquad \qquad \qquad \pHappens{walking(ID_1)}{T}\ \wedge \\
    \qquad \qquad \qquad \neg \pClose{ID_1}{ID_2}{34}{T} \\
  \end{mysplit} \\
%
&\begin{mysplit} \label{dsa:term_move2}
   \pTerminatedAt{moving(ID_1,\, ID_2)}{T} \Leftarrow \qquad \qquad \\
    \qquad \qquad \qquad \pHappens{active(ID_1)}{T}\ \wedge \\
    \qquad \qquad \qquad \pHappens{active(ID_2)}{T} \\
  \end{mysplit} \\
%
&\begin{mysplit}
\label{dsa:term_move3}
   \pTerminatedAt{moving(ID_1,\, ID_2)}{T} \Leftarrow \qquad \qquad \\
    \qquad \qquad \qquad \pHappens{active(ID_1)}{T}\ \wedge \\
    \qquad \qquad \qquad \pHappens{inactive(ID_2)}{T} \\
  \end{mysplit} \\
%
&\begin{mysplit}
\label{dsa:term_move4}
   \pTerminatedAt{moving(ID_1,\, ID_2)}{T} \Leftarrow \qquad \qquad \\
    \qquad \qquad \qquad \pHappens{running(ID_1)}{T} \\
  \end{mysplit} \\
&\begin{mysplit}
\label{dsa:term_move5}
   \pTerminatedAt{moving(ID_1,\, ID_2)}{T} \Leftarrow \qquad \qquad \\
    \qquad \qquad \qquad \pHappens{exit(ID_1)}{T}
  \end{mysplit}
\end{align}
The predicate $\folOM$ is a spatial constraint, stating that the orientation of two persons is almost the same (e.g.~the difference is below $45$ degrees). According to rule \eqref{dsa:init_move}, the \moving\ CE is initiated when two persons $ID_1$ and $ID_2$ are walking close to each other (their distance is at most $34$ pixels) with almost the same orientation. The \moving\ CE is terminated under several cases: (a) As specified by rule \eqref{dsa:term_move1}, when people walk away from each other, i.e.~they have a distance larger than $34$ pixels. (b) When none is actually moving, i.e.~both are staying active,  or (c) one is active while the other is inactive, represented by rules \eqref{dsa:term_move2} and \eqref{dsa:term_move3}. (d) Finally, when one of them is running or exiting the scene, represented by rules \eqref{dsa:term_move4} and \eqref{dsa:term_move5}, respectively.  

\section{Markov Logic Networks} \label{sec:MLN}

Although the Event Calculus can compactly represent complex event relations, it does not handle uncertainty adequately. A knowledge base of Event Calculus axioms and composite event (CE) definitions is defined by a set of first-order logic formulas. Each formula imposes a (hard) constraint over the set of possible worlds, that is, Herbrand interpretations. A missed or an erroneous simple, derived event (SDE) detection can have a significant effect on the event recognition results. For example, an initiation may be based on an erroneously detected SDE, causing the recognition of a CE with absolute certainty.

We employ the framework of Markov Logic Networks\footnote{Systems implementing MLN reasoning and learning algorithms can be found at the following addresses:\linebreak 
{ 
\url{http://alchemy.cs.washington.edu}
\linebreak 
\url{http://research.cs.wisc.edu/hazy/tuffy}
\linebreak
\url{http://code.google.com/p/thebeast}
\linebreak
\url{http://ias.cs.tum.edu/probcog-wiki} 
}} (MLNs) \cite{Domingos2009markov} in order to soften these constraints and perform probabilistic inference. In MLNs, each formula $F_i$ is represented in first-order logic and is associated with a weight value $w_i \in \mathbb{R}$. The higher the value of weight $w_i$, the stronger the constraint represented by formula $F_i$. In contrast to classical logic, all worlds in MLNs are possible with a certain probability. The main idea behind this is that the probability of a world increases as the number of formulas it violates decreases. A knowledge base in MLNs may contain both hard and soft-constrained formulas. Hard-constrained formulas are associated with an infinite weight value and capture the knowledge which is assumed to be certain. Therefore, an acceptable world must at least satisfy the hard constraints. Soft constraints capture imperfect knowledge in the domain, allowing for the existence of worlds in which this knowledge is violated.

Formally, a knowledge base $L$ of weighted formulas, together with a finite domain of constants $\mathcal{C}$, is transformed into a ground Markov network $M_{L,\mathcal{C}}$. In our case, $L$ consists of Event Calculus axioms and CE definitions, and $\mathcal{C}{=}\mathcal{T}\cup \mathcal{O} \cup \mathcal{E} \cup \mathcal{F}$.  All formulas are converted into \textit{clausal form} and each clause is ground according to the domain of its distinct variables. The nodes in $M_{L,\mathcal{C}}$ are Boolean random variables, each one corresponding to a possible grounding of a predicate that appears in $L$. The predicates of a ground clause form a clique in $M_{L,\mathcal{C}}$. Each clique is associated with a corresponding weight $w_i$ and a Boolean \emph{feature}, taking the value 1 when the ground clause is true and 0 otherwise. The ground $M_{L,\mathcal{C}}$ defines a probability distribution over possible worlds and is represented as a log-linear model. 

In event recognition we aim to recognise CEs of interest given the observed streams of SDEs. For this reason we focus on discriminative MLNs \cite{singla2005discriminative}, that are akin to Conditional Random Fields \cite{LaffertyMP01CRF,sutton2006introduction}. Specifically, the set of random variables in $M_{L,\mathcal{C}}$ can be partitioned into two subsets. The former is the set of evidence random variables $X$, formed by a narrative of input ground $\folhappens$ predicates and spatial constraints. The latter is the set of random variables $Y$ that correspond to groundings of query $\folholdsAt$ predicates, as well as groundings of any other hidden/unobserved predicates. The joint probability distribution of a possible assignment of $Y{=}\mathbf{y}$, conditioned over a given assignment of $X{=}\mathbf{x}$, is defined as follows:
\begin{equation}
\begin{mysplit} \label{eq:mln_full_cond_PD}
P(Y{=}\mathbf{y}\, |\, X{=}\mathbf{x}) = \dfrac{1}{Z(\mathbf{x})} exp \left( \sum\limits_{i{=}1}^{|F_c|} w_i n_i(\mathbf{x,y}) \right)
\end{mysplit}
\end{equation}

\noindent  The vectors $\mathbf{x}\in\mathcal{X}$ and $\mathbf{y}\in\mathcal{Y}$ represent a possible assignment of evidence $X$ and query/hidden variables $Y$, respectively. $\mathcal{X}$ and $\mathcal{Y}$ are the sets of possible assignments that the evidence $X$ and query/hidden variables $Y$ can take. $F_c$ is the set of clauses produced from the knowledge base $L$ and the domain of constants $\mathcal{C}$. The scalar value $w_i$ is the weight of the \textit{i}-th clause and $n_i(\mathbf{x,y})$ is the number of satisfied groundings of the \textit{i}-th clause in $\mathbf{x}$ and $\mathbf{y}$. $Z(\mathbf{x})$ is the partition function, that normalises over all possible assignments $\mathbf{y' \in \mathcal{Y}}$ of query/hidden variables given the assignment $\mathbf{x}$, that is, $Z(\mathbf{x}) = \sum_{\mathbf{y'} \in \mathcal{Y}}exp( \sum_{i}^{|F_c|} w_i n_i(\mathbf{x,y'}) )$. 

Equation \eqref{eq:mln_full_cond_PD} represents a single exponential model for the joint probability of the entire set of query variables that is globally conditioned on a set of observables. Such a conditional model can have a much simpler structure than a full joint model, e.g.~a Bayesian Network. By modelling the conditional distribution directly, the model is not affected by potential dependencies between the variables in $X$ and can ignore them. The model also makes independence assumptions among the random variables $Y$, and defines by its structure the dependencies of $Y$ on $X$. Furthermore, conditioning on a specific assignment $\mathbf{x}$, given by the observed SDEs, reduces significantly the number of possible worlds and inference becomes much more efficient \cite{singla2005discriminative,minka2005discriminative,sutton2006introduction}.

Still, directly computing equation \eqref{eq:mln_full_cond_PD} is intractable, because the value of $Z(\mathbf{x})$ depends on the relationship among all clauses in the knowledge base. For this reason, a variety of efficient inference algorithms have been proposed in the literature, based on local search and sampling \cite{poon2006sound,singla2006memory,biba2011engineering}, variants of Belief Propagation \cite{singla2008lifted,kersting2009counting}, Integer Linear Programming \cite{riedel2008improving,huynh2009max}, etc. 

In this work we consider two types of inference, i.e.~marginal inference and maximum a-posteriori inference (MAP). The former type of inference computes the conditional probability that CEs hold given a narrative of observed SDEs, i.e.~$P(\pHoldsAt{CE}{T}{=}\mathit{True}\, |\, \mathit{SDE})$. In other words, this probability value measures the confidence that the CE is recognised. Since it is \#P-complete to compute this probability, we employ the state-of-the-art sampling algorithm \textit{MC-SAT} \cite{poon2006sound} to approximate it. The algorithm combines Markov Chain Monte Carlo sampling with satisfiability testing and even in large state spaces with deterministic dependencies (e.g.~hard-constrained formulas) it can  approximate this probability efficiently. The latter type of inference identifies the most probable assignment among all $\folholdsAt$ instantiations that are consistent with the given narrative of observed SDEs, i.e.~$\underset{\tiny \folholdsAt}{\argmax}\ P(\pHoldsAt{CE}{T}\, |\, \mathit{SDE})$. In MLNs this task reduces to finding the truth assignment of all $\folholdsAt$ instantiations that maximises the sum of weights of satisfied ground clauses. This is equivalent to the weighted maximum satisfiability problem. The problem is NP-hard in general and in order to find an approximate solution efficiently we employ the LP-relaxed Integer Linear Programming method proposed by \citeN{huynh2009max}.

The weights of the soft-constrained clauses in MLNs can be estimated from training data, using supervised learning techniques. When the goal is to learn a model that recognises CEs with some confidence (i.e.~probability), then the most widely adopted approach is to minimise the negative Conditional Log-Likelihood (CLL) function --- derived from equation \eqref{eq:mln_full_cond_PD}. This can be achived by using either \textit{first-order} or \textit{second-order} optimisation methods \cite{singla2005discriminative,lowd2007efficient}. First-order methods apply standard gradient descent optimisation techniques, e.g.~the voted perceptron algorithm \cite{collins2002discriminative,singla2005discriminative}, while second-order methods pick a search direction based on the quadratic approximation of the target function. As stated by \citeN{lowd2007efficient}, second-order methods are more appropriate for MLN training, as they do not suffer from the problem of \textit{ill-conditioning}. In a training set some clauses may have a significantly greater number of satisfied groundings than others, causing the variance of their counts to be correspondingly larger. This situation causes the standard gradient descent methods to converge very slowly, since there is no single appropriate learning rate for all soft-constrained clauses. An alternative approach to CLL function optimisation is max-margin training, which is better suited to problems where the goal is to maximise the classification accuracy \cite{huynh2009max,huynh2011online}. Instead of optimising the CLL function, max-margin training aims to maximise the ratio between the probability of the correct truth assignment of CEs to hold and the closest competing incorrect truth assignment. In this work we assess both the second-order Diagonal Newton algorithm \cite{singla2005discriminative} and the max-margin method proposed by \citeN{huynh2009max}.

\section{Compact Markov Network Construction} \label{sec:CompactMN}

The use of MLNs for inference and learning requires the grounding of the entire knowledge base used for event recognition, including the domain-independent axioms of the Event Calculus (axioms \eqref{axiom:fol_dec7a_holdsAt}{--}\eqref{axiom:fol_dec7a_not_holdsAt_inertia}). Unless optimized, this process leads to unmanageably large ground Markov Networks, where inference and learning become practically infeasible. This section presents our approach to addressing this problem.

\subsection{Simplified Representation} \label{sec:ECGrounding}

The choice of Event Calculus dialect, as presented in Section \ref{sec:EventCalculus}, has a significant impact on the grounding process. For example, Shanahan's Full Event Calculus \cite{shanahan1999event} employs axioms  that contain triply quantified time-point variables. As a result, the number of their groundings has a cubic relation to the number of time-points. Furthermore, that formalism contains existentially quantified variables over events and time-points. During MLN grounding existentially quantified formulas are replaced by the disjunction of their groundings \cite{Domingos2009markov}. This leads to a large number of disjunctions and a combinatorial explosion of the number of clauses, producing unmanageably large Markov networks.

In contrast, the proposed Event Calculus (\MLNEC) is based on the Discrete Event Calculus \cite{mueller2008event}, where the domain-independent axioms are defined over successive time-points. For example, axiom \eqref{axiom:fol_dec7a_holdsAt} produces one clause\footnote{In Conjunctional Normal Form.} and has two distinct variables $F$ and $T$. Therefore, the number of its groundings is determined by the Cartesian product of the corresponding variable-binding constraints, that is $|\mathcal{F}|{\times}|\mathcal{T}|$. Assuming that the domain of fluents $\mathcal{F}$ is relatively small compared to the domain of time-points $\mathcal{T}$, the number of groundings of axiom \eqref{axiom:fol_dec7a_holdsAt} grows linearly to the number of time-points. Furthermore, in \MLNEC\ the initiation and termination of fluents  --- representing CEs --- are only defined in terms of fluents and time-points (see the general form \eqref{dsa:fol_general_form}). This representation reduces further the number of variables and eliminates the existential quantification in the domain-independent axioms. As a result, \MLNEC\ produces a substantially smaller number of ground clauses, than many other dialects of Event Calculus.

\subsection{Knowledge Base Transformation}  \label{sec:kbc}

In addition to choosing an Event Calculus dialect that makes the number of ground clauses linearly dependent on the number of time-points, we can achieve significant improvements in the size of the ground Markov Networks, by making the Closed World Assumption. 

A knowledge base with domain-dependent rules in the form of \eqref{dsa:fol_general_form} describes explicitly the conditions in which fluents are initiated or terminated. It is usually impractical to define also when a fluent is \emph{not} initiated and \emph{not} terminated. However, the open-world semantics of first-order logic result in an inherent uncertainty about the value of a fluent for many time-points. In other words, if at a specific time-point  no event that terminates or initiates a fluent happens, we cannot rule out the possibility that the fluent has been initiated or terminated.  As a result, we cannot determine whether a fluent holds or not, leading to the loss of inertia.

This is a variant of the well-known frame problem and one solution for the Event Calculus in first-order logic is the use of circumscription \cite{mccarthy1980circ,lifschitz1994,shanahan1997solving,doherty1997computing,mueller2008event}.  The aim of circumscription is to automatically rule out all those conditions which are not explicitly entailed by the given formulas. Hence, circumscription introduces a closed-world assumption to first-order logic. 

Technically, we perform circumscription by predicate completion --- a syntactic transformation where formulas are translated into logically stronger ones. In particular, we perform a knowledge transformation procedure in which predicate completion is computed for both $\folinitiatedAt$ and $\folterminatedAt$ predicates. Due to the form of CE definitions (see formalisation \eqref{dsa:fol_general_form}), the result of predicate completion is applied to each CE separately, e.g.~$\pInitiatedAt{meeting(ID_1,\, ID_2)}{T}$, rather than to a generic  $\pInitiatedAt{F}{T}$ predicate. Similar to \citeN{mueller2008event}, we also eliminate the  $\folinitiatedAt$ and $\folterminatedAt$ predicates from the knowledge base, by exploiting the equivalences resulting from predicate completion. In cases where the definitions of the initiation or termination of a specific CE are missing, the corresponding initiation or termination is  considered \textit{False} for all time-points, e.g.~$\folterminatedAt(\mathit{fluent(X,Y)},$ $T)\Leftrightarrow \mathit{False}$.

To illustrate the form of the resulting knowledge base, consider the domain-dependent definition of \meeting\ --- i.e.~rules \eqref{dsa:init_meet1}{--}\eqref{dsa:term_meet3}. After predicate completion, these rules will be replaced by the following formulas:
\begin{align}
&\begin{mysplit} \label{circ:Initiation_example}
  \pInitiatedAt{meeting(ID_1,\, ID_2)}{T} \Leftrightarrow  \\
  \qquad \qquad \qquad \bigl(\pHappens{active(ID_1)}{T}\ \wedge \\
  \qquad \qquad \qquad \neg \pHappens{running(ID_2)}{T}\ \wedge\  \\
  \qquad \qquad \qquad \pClose{ID_1}{ID_2}{25}{T}\, \bigr)\ \bigvee \\
  \qquad \qquad \qquad \bigl(\pHappens{inactive(ID_1)}{T}\ \wedge \\
  \qquad \qquad \qquad \neg \pHappens{running(ID_2)}{T}\ \wedge \\
  \qquad \qquad \qquad \neg \pHappens{active(ID_2)}{T}\ \wedge \\
  \qquad \qquad \qquad \pClose{ID_1}{ID_2}{25}{T}\, \bigr)
\end{mysplit} \\
&\begin{mysplit} \label{circ:Termination_example}
\pTerminatedAt{meeting(ID_1,\, ID_2)}{T} \Leftrightarrow  \\
  \qquad \qquad \qquad \bigl( \pHappens{walking(ID_1)}{T}\ \wedge \\
  \qquad \qquad \qquad \neg \pClose{ID_1}{ID_2}{25}{T}\, \bigr)\ \bigvee \\
  \qquad \qquad \qquad \pHappens{running(ID_1)}{T}\  \bigvee \\
  \qquad \qquad \qquad \pHappens{exit(ID_1)}{T}
\end{mysplit}
\end{align}
\noindent The resulting rules \eqref{circ:Initiation_example} and \eqref{circ:Termination_example} define all conditions under which the \meeting\ CE is initiated or terminated. Any other event occurrence cannot affect this CE, as it cannot initiate the CE or terminate it. Based on the equivalence in formula  \eqref{circ:Initiation_example}, the domain-independent axiom \eqref{axiom:fol_dec7a_holdsAt} is automatically re-written into the following specialised form\footnote{This direct re-writing of \eqref{circ:Initiation_example} results to a single formula that contains the disjunction of formula \eqref{circ:Initiation_example}. However, for reasons that have to do with the handling of uncertainty in MLN and will be discussed in a later section, in \eqref{flattened:fol_dec7a_holdsAt-example} we choose to equivalently represent it using two separate formulas.}:
\begin{align}
\label{flattened:fol_dec7a_holdsAt-example}
&\begin{mysplit} 
  \pHoldsAt{meeting(ID_1,\, ID_2)}{T{+}1} \Leftarrow \\
  \qquad \qquad \qquad \pHappens{active(ID_1)}{T}\ \wedge \\
  \qquad \qquad \qquad \neg \pHappens{running(ID_2)}{T}\ \wedge\  \\
  \qquad \qquad \qquad \pClose{ID_1}{ID_2}{25}{T}\\ \ \\
  \pHoldsAt{meeting(ID_1,\, ID_2)}{T{+}1} \Leftarrow \\
  \qquad \qquad \qquad \pHappens{inactive(ID_1)}{T}\ \wedge \\
  \qquad \qquad \qquad \neg \pHappens{running(ID_2)}{T}\ \wedge \\
  \qquad \qquad \qquad \neg \pHappens{active(ID_2)}{T}\ \wedge \\
  \qquad \qquad \qquad \pClose{ID_1}{ID_2}{25}{T}
 \end{mysplit}
\end{align}

\begin{samepage}
\noindent Similarly, the inertia axiom \eqref{axiom:fol_dec7a_holdsAt_inertia} can be re-written according to \eqref{circ:Termination_example} as follows:
\begin{align}
\label{flattened:fol_dec7a_holdsAt_inertia-example}
&\begin{mysplit}
\ \ \  \pHoldsAt{meeting(ID_1,\, ID_2)}{T{+}1} \Leftarrow \\
  \qquad \qquad \qquad \pHoldsAt{meeting(ID_1,\, ID_2)}{T}\ \wedge \\
  \qquad \qquad \qquad \neg \Bigl(\, \bigl( \pHappens{walking(ID_1)}{T}\ \wedge \\
  \qquad \qquad \qquad \qquad \neg \pClose{ID_1}{ID_2}{25}{T}\, \bigr)\ \bigvee \\
  \qquad \qquad \qquad \qquad \pHappens{running(ID_1)}{T}\  \bigvee \\
  \qquad \qquad \qquad \qquad \pHappens{exit(ID_1)}{T}\, \Bigr)
\end{mysplit}
\end{align}
\end{samepage}

The result of this transformation procedure replaces the original set of domain-independent axioms and domain-dependent CE definitions with a logically stronger knowledge base. The rules in the resulting knowledge base form the template that MLNs will use to produce ground Markov networks. 
The transformed formulas produce considerably more compact ground Markov networks than the original ones, as the clauses to be grounded are reduced. Moreover, the predicates $\folinitiatedAt$ and $\folterminatedAt$ are eliminated and the corresponding random variables are not added to the network. This reduction decreases substantially the space of possible worlds, since the target random variables of the network ($Y$ in equation \eqref{eq:mln_full_cond_PD}) are limited only to the corresponding $\folholdsAt$ ground predicates. Specifically, the space of possible worlds is reduced from $2^{3{\times}|\mathcal{F}|{\times}|\mathcal{T}|}$ to $2^{|\mathcal{F}|{\times}|\mathcal{T}|}$ --- where $|\mathcal{T}|$ and $|\mathcal{F}|$ denote the number of distinct time-points and fluents, respectively. These reductions improve the computational performance of the probabilistic inference. Furthermore, due to the reduced space of possible worlds, the same number of sampling iterations results in better probability estimates.

Formally, the resulting knowledge base is composed of rules having the following form:
\begin{numcases}{\Sigma\, =}
 \begin{mysplit} \label{sigma:HoldsAt}
  \ \ \pHoldsAt{fluent_1(X,\, Y)}{T{+}1} \Leftarrow \\
    \qquad \pHappens{event_i(X)}{T} \wedge \ldots \wedge \CEConditions  \qquad \qquad \quad\ \\
  \dots \\
 \end{mysplit} \\
 \begin{mysplit} \label{sigma:not_HoldsAt}
  \neg \pHoldsAt{fluent_1(X,Y)}{T{+}1} \Leftarrow \\
    \qquad \pHappens{event_j(X)}{T} \wedge \ldots \wedge \CEConditions \\
  \dots \\
 \end{mysplit}
\end{numcases}
\begin{numcases}{\Sigma' = }
  \begin{mysplit} \label{sigma':HoldsAt}
   \ \ \pHoldsAt{fluent_1(X,Y)}{T{+}1} \Leftarrow \\
    \qquad \pHoldsAt{fluent_1(X,Y)}{T}\ \wedge \\
    \qquad \neg \bigl(\, (\, \pHappens{event_j(X)}{T} \wedge \ldots \wedge \CEConditions\, ) \bigvee \ldots\, \bigr) \\
  \dots \\
  \end{mysplit} \\
 \begin{mysplit} \label{sigma':not_HoldsAt}
 \neg \pHoldsAt{fluent_1(X,Y)}{T{+}1} \Leftarrow \\
    \qquad \neg \pHoldsAt{fluent_1(X,Y)}{T}\ \wedge \\
    \qquad \neg \bigl(\, (\, \pHappens{event_i(X)}{T} \wedge \ldots \wedge \CEConditions\, ) \bigvee \ldots\, \bigr) \\
  \dots \\
 \end{mysplit}
\end{numcases}
\noindent The rules in \eqref{sigma:HoldsAt}{--}\eqref{sigma':not_HoldsAt} can be separated into two subsets. The former set $\Sigma$ contains specialised definitions of axioms \eqref{axiom:fol_dec7a_holdsAt} and \eqref{axiom:fol_dec7a_not_holdsAt}, specifying when a fluent holds (or does not hold) when its initiation (or termination) conditions are met. The latter set $\Sigma^{'}$ contains specialised definitions of the inertia axioms \eqref{axiom:fol_dec7a_holdsAt_inertia} and \eqref{axiom:fol_dec7a_not_holdsAt_inertia}, specifying whether a specific fluent continues to hold or not at any instance of time.

The knowledge transformation procedure reduces the size of the produced network, based only on the rules of the knowledge base. Given a narrative of SDEs, further reduction can be achieved during the ground network construction. All ground predicates that appear in the given narrative are replaced by their truth value. Ground clauses that become tautological are safely removed, as they remain satisfied in all possible worlds \cite{singla2005discriminative,shavlik2009speeding}. Therefore, the resulting network comprises only the remaining ground clauses, containing ground predicates with unknown truth states --- i.e.~groundings of $\folholdsAt$.

\section{The Behaviour of the Probabilistic Event Calculus} \label{sec:inertia}

As mentioned in Section \ref{sec:MLN}, weighted formulas in MLNs define soft constraints, allowing some worlds that do not satisfy these formulas to become likely. For example, consider a knowledge base of Event Calculus axioms and CE definitions (e.g.~\meeting\ and \moving) compiled in the form of rules \eqref{sigma:HoldsAt}{--}\eqref{sigma':not_HoldsAt}. Given a narrative of SDEs, the probability of a CE to hold at a specific time-point is determined by the probabilities of the worlds in which this CE holds. Each world, in turn, has some probability which is proportional to the sum of the weights of the ground clauses that it satisfies. Consequently, the probability of a CE to hold at a specific instance of time depends on the corresponding constraints of the ground Markov network. Thus, by treating the rules in the $\Sigma$ and $\Sigma'$ sets as either hard or soft constraints, we can modify the behaviour of the Event Calculus. 

%
%
\subsection{Soft-constrained rules in $\Sigma$}

In order to illustrate how the probability of a CE is affected when its initiation or termination conditions are met, consider the case that the rules in $\Sigma$ are soft-constrained while the inertia rules in $\Sigma'$ remain hard-constrained. By soft-constraining the rules in $\Sigma$, the worlds violating their clauses become probable. This situation reduces the certainty with which a CE is recognised when its initiation or termination conditions are met. For example, assume that the initiation rules \eqref{flattened:fol_dec7a_holdsAt-example} of the \meeting\ CE are associated with weights. As a result, the \meeting\ activity is initiated with some certainty, causing the CE to hold with some probability. Depending on the strength of the weights, the worlds that violate these rules become more or less likely. Thus, we can control the level of certainty with which a CE holds or not under the same conditions.

When the initiation conditions are met, the probability of the CE to hold increases. Equivalently, when the termination conditions are satisfied, the probability of the CE decreases. At the same time, all worlds violating hard-constrained inertia rules in $\Sigma'$ are rejected. In the presence of SDEs leading to the partial satisfaction (i.e.~satisfaction of a possibly empty strict subset) of the initiation/termination conditions, the probability of a CE to hold is not affected. The inertia is retained deterministically  as in crisp logic. 

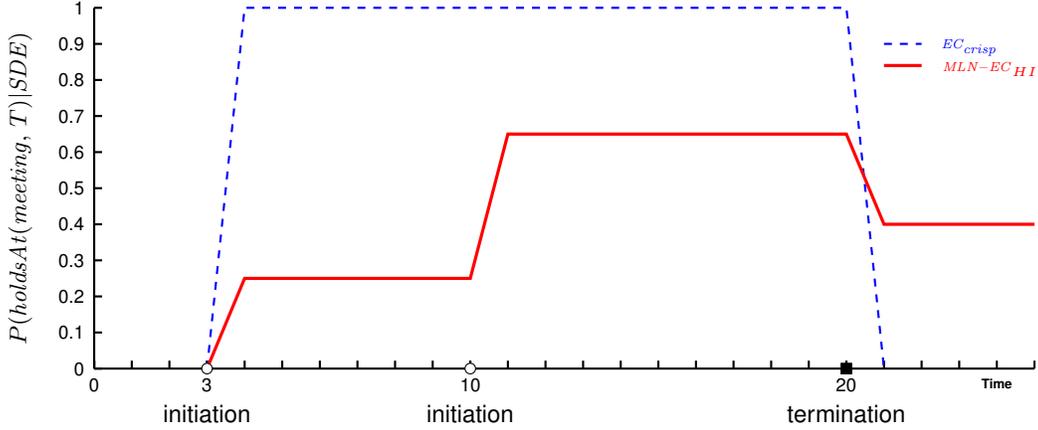
\begin{figure}[t]
\centering
\begin{tikzpicture}[y=4.8cm, x=.5cm,font=\sffamily, >=stealth] 
 	axis
 	\draw[-,thick] (0,0) -- (25,0) node[below, at={(24,0)}] {\tiny \textbf{Time}};
     	\draw[thick] (0,0) -- coordinate (y axis mid) (0,1);
    	ticks
    	\foreach \x in {0,3,10,20}
     		\draw[thick] (\x,3pt) -- (\x,0pt) node[anchor=north] {\scriptsize \x};

    	\foreach \x in {0,...,25}
	    \draw[thick] (\x,3pt) -- (\x,0pt);

    	\foreach \y in {0, 0.1, 0.2, 0.3, 0.4, 0.5, 0.6, 0.7, 0.8, 0.9, 1}
     		\draw[thick] (3pt,\y) -- (0pt,\y) node[anchor=east] {\scriptsize \y}; 
	labels      
	\node[rotate=90, above=0.7cm] at (y axis mid) {\small $P(\pHoldsAt{meeting}{T}|\mathit{SDE})$};

	plots
	\draw[blue,thick] node[at={(22, 0.89)},label=right:\scriptsize{\tiny  $\ECLP$}] {};
	\draw[blue,dashed,thick] plot coordinates {(21,0.9) (22,0.9)};
	\draw[blue,dashed,thick] plot coordinates {(3,0.0) (4,1.0) (20,1.0) (21,0.0)};

	\draw[red,very thick] node[at={(22, 0.83)},label=right:\scriptsize{\tiny $\PEC_{HI}$}] {};
	\draw[red,very thick] plot coordinates {(21, 0.84) (22, 0.84)};
	\draw[red,very thick] plot[id=mln] coordinates {(3,0.0) (4,0.25) (10,0.25) (11,0.65) (20,0.65) (21,0.4) (25,0.4)};

 	\draw node[at={(3,-0.05)},label=below:\small{initiation}] {};
	\draw plot[mark=*, mark options={fill=white}] coordinates {(3,0.0)};

	\draw node[at={(10,-0.05)},label=below:\small{initiation}] {};
	\draw plot[mark=*, mark options={fill=white}] coordinates {(10,0.0)};

	\draw node[at={(20,-0.05)},label=below:\small{termination}] {};
	\draw plot[mark=square*, mark options={fill=black}] coordinates {(20,0.0)};

\end{tikzpicture} 
\caption{The probability of the \meeting\ CE given some SDE narrative. $\ECLP$ is a crisp Event Calculus. $\HI$ is a probabilistic Event Calculus where rules in $\Sigma$ are soft-constrained, while the inertia rules in $\Sigma'$ remain hard-constrained.}
\label{fig:example_meeting-hi} 
\end{figure}

Figure \ref{fig:example_meeting-hi} illustrates this behaviour with the fluent \meeting\ that initially does not hold at time $0$. According to the narrative of SDEs, the \meeting\ activity is initiated at time-points $3$ and  $10$, e.g.~satisfying the constraints imposed by rules \eqref{dsa:init_meet1} and \eqref{dsa:init_meet2} respectively. At time $20$, the \meeting\ activity is terminated by the conditions of rule \eqref{dsa:term_meet1}. In crisp Event Calculus, denoted as $\ECLP$, after its first initiation the \meeting\ activity holds with absolute certainty. The second initiation at time $10$ does not cause any change and the CE continues to hold. The termination at time $20$ causes the CE to not hold, again with absolute certainty, for the remaining time-points. In $\HI$ (hard-constrained inertia rules), however, the rules in $\Sigma$ are soft-constrained. As a result, at time-point $4$ the probability of \meeting\ to hold increases to some value. Similar to $\ECLP$, the inertia is fully retained and the probability of \meeting\ deterministically persists in the interval $4$ to $10$. In contrast to $\ECLP$, the second initiation at time-point $10$ increases the certainty of \meeting\ to hold. As a result, the probability of \meeting\ is higher in the interval $11$ to $20$. In the same manner, the termination at $20$ reduces the probability of \meeting\ and the CE continues to hold with some reduced probability.

\ifdefined\TIKZEXTERNALIZE
  \tikzsetnextfilename{figure_example_meeting-hi}
\fi

%
%
\subsection{Soft-constrained inertia rules in $\Sigma'$}
%
%
To illustrate how the behaviour of inertia is affected by soft-constraining the corresponding rules in $\Sigma'$, consider that the rules in $\Sigma$ are hard-constrained. Consequently, when the initiation (or termination) conditions are met, a CE holds (or does not hold) with absolute certainty. 
The persistence of a CE depends on its inertia rules in $\Sigma'$. If the inertia of $\folholdsAt$ is hard-constrained, the worlds in which an initiated CE does not hold are rejected. Similarly, by keeping the inertia of $\neg \folholdsAt$ hard-constrained, all worlds in which a terminated CE holds are rejected. By soft-constraining these rules we control the strength of the inertia constraints. Thus, in the presence of SDEs leading to the partial satisfaction of the corresponding initiation/termination conditions, a CE may not persist with absolute certainty, as worlds that violate these constraints become likely. The persistence of $\folholdsAt$ and $\neg \folholdsAt$ is gradually lost over successive time-points. When allowing the constraints of $\folholdsAt$ inertia to be violated, the probability of a CE gradually drops. Similarly, by allowing the constraints representing the inertia of $\neg \folholdsAt$ to be violated, the probability of a CE gradually increases. The lower the value of the weight on the constraint, the more probable the worlds that violate the constraints become. In other words,  weight values in $\Sigma'$ cause CE to persist for longer or shorter time periods. 

Since the sum of the probabilities of $\folholdsAt$ and $\neg \folholdsAt$ for a specific CE is always equal to 1, the relative strength of $\folholdsAt$ and $\neg \folholdsAt$ rules in $\Sigma'$ determines the type of inertia in the model. The following two general cases can be distinguished.

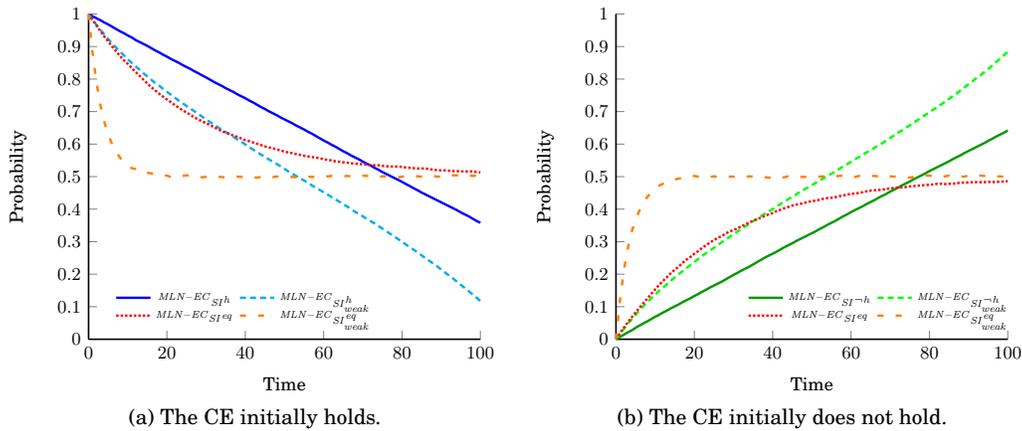
\begin{figure}[t]
\pgfplotsset{
every axis/.append style={thick},
tick label style={font=\small}
}
%
%
\ifdefined\TIKZEXTERNALIZE
  \tikzsetnextfilename{figure_eg_behaviour1_holdsAt}
\fi
\subfloat[The CE initially holds.\label{fig:eg_behaviour1_holds}]{
\begin{tikzpicture}[scale=0.76]
    \begin{axis}[
    legend columns=2,
    legend style={ at={(0.05,0.1)},anchor=west,draw=none},
    xlabel={\small Time},
    ylabel={\small Probability},
    ymin=0, ymax=1.0, xmin=0, xmax=100, 
    ytick={0, 0.1, 0.2, 0.3, 0.4, 0.5, 0.6, 0.7, 0.8, 0.9, 1},
    thick,
    axis x line*=bottom,
    axis y line*=left]

     \pgfplotstableread{One.data}\exampledata
    
     \addplot[color=blue,very thick] table[y=SIT,x=Time] {\exampledata};
 	\addlegendentry{\tiny $\SIT$}

     \addplot[color=cyan,very thick,densely dashed] table[y=SITW,x=Time] {\exampledata};
 	\addlegendentry{\tiny $\SITWeak$}
    
     \addplot[color=red, very thick,densely dotted] table[y=SI,x=Time] {\exampledata};
 	\addlegendentry{\tiny $\SIEq$}

     \addplot[color=orange,very thick,loosely dashed] table[y=SIW,x=Time] {\exampledata};
 	\addlegendentry{\tiny $\SIEqWeak$}

    \end{axis}
  \end{tikzpicture}
}
~
%
%
\ifdefined\TIKZEXTERNALIZE
  \tikzsetnextfilename{figure_eg_behaviour1_not-holdsAt}
\fi
\subfloat[The CE initially does not hold.\label{fig:eg_behaviour1_not-holds}]{ 
  \begin{tikzpicture}[scale=0.76]
    \begin{axis}[
    legend columns=2,
    legend style={ at={(1.03,0.1)},anchor=east,draw=none},
    xlabel={\small Time},
    ylabel={\small Probability},
    ymin=0, ymax=1.0, xmin=0, xmax=100, 
    ytick={0, 0.1, 0.2, 0.3, 0.4, 0.5, 0.6, 0.7, 0.8, 0.9, 1},
    thick,
    axis x line*=bottom,
    axis y line*=left]
     \pgfplotstableread{Zero.data}\exampledata    

     \addplot[color=green!60!black,very thick] table[y=SII,x=Time] {\exampledata};
 	\addlegendentry{\tiny $\SII$}

     \addplot[color=green,very thick,densely dashed] table[y=SIIW,x=Time] {\exampledata};
 	\addlegendentry{\tiny $\SIIWeak$}

     \addplot[color=red,very thick,densely dotted] table[y=SI,x=Time] {\exampledata};
 	\addlegendentry{\tiny $\SIEq$}

     \addplot[color=orange,very thick,loosely dashed] table[y=SIW,x=Time] {\exampledata};
 	\addlegendentry{\tiny $\SIEqWeak$}

    \end{axis}
  \end{tikzpicture}
}
\caption{In both figures SDEs occur leading to the partial satisfaction of the initiation/termination conditions of a CE in the interval 0 to 100. In the left figure the CE holds at time $0$ with absolute certainty, while in the right figure the CE does not hold at time $0$.}
\label{fig:eg_behaviour1}
\end{figure}

\paragraph{Equally strong inertia constraints} All rules in $\Sigma'$ are equally soft-constrained, i.e.~they are associated with the same weight value. Consequently, both inertia rules of $\folholdsAt$ and $\neg \folholdsAt$ for a particular CE impose constraints of equal importance, allowing worlds that violate them to become likely. As a result, in the absence of useful evidence, the probability of $\folholdsAt$ will tend to approximate the value $0.5$. For example, Figure \ref{fig:eg_behaviour1_holds} illustrates soft persistence for the \meeting\ CE when it holds with absolute certainty at time-point $0$, and thereafter nothing happens to initiate or terminate it. The curve $\SIEq$ (soft-constrained inertia rules with equal weights) shows the behaviour of inertia in this case. As time evolves, the probability of \meeting\ appears to gradually drop, converging to $0.5$. If we assign weaker weights to the inertia axioms, shown by the $\SIEqWeak$ curve, the probability of \meeting\ drops more sharply. Similarly, in Figure \ref{fig:eg_behaviour1_not-holds}, the \meeting\ CE is assumed to not hold initially. As time evolves, the probability of \meeting\ gradually increases up to the value $0.5$, as shown by the $\SIEq$ and $\SIEqWeak$ curves respectively.

%
%

\paragraph{Inertia constraints of different strength} When the inertia rules of $\folholdsAt$ and \linebreak $\neg \folholdsAt$ for a particular CE in $\Sigma'$ have different weights, the probability of the CE will no longer converge to $0.5$. Since the weights impose constraints with different confidence, worlds violating the stronger constraints become less likely than worlds violating the weaker ones. Depending on the relative strength of the weights, the probability of the CE may converge either to $1.0$ or $0.0$. The relative strength of the weights affects also the rate at which the probability of CE changes. As an extreme example, in Figure \ref{fig:eg_behaviour1_holds}, the rules for the inertia of $\neg \folholdsAt$ remain hard-constrained. By assigning weights to the rules for the inertia of $\folholdsAt$, the persistence of the CE is lost. Since the inertia constraints of $\folholdsAt$ are weaker than the constraints of $\neg \folholdsAt$, worlds violating the former set of constraints will always be more likely. As a result, the probability of the CE will continue to drop, even below $0.5$. The curves $\SIT$ (soft-constrained inertia of $\folholdsAt$) and $\SITWeak$ (weaker $\folholdsAt$ inertia constraints) illustrate how the probability of \meeting\ drops sharply towards $0.0$. The weaker the constraints ($\SITWeak$) the steeper the drop. In a similar manner, when the inertia constraints of $\neg \folholdsAt$ are weaker than the constraints of $\folholdsAt$, the probability of CE gradually increases and may reach values above $0.5$ --- presented by the $\SII$ (soft-constrained inertia of $\neg \folholdsAt$) and $\SIIWeak$ (weaker $\neg \folholdsAt$ inertia constraints) cases in Figure \ref{fig:eg_behaviour1_not-holds}. 

\vspace{0.6cm}

As explained in Section \ref{sec:kbc}, the inertia rule of a specific CE may consist of a large body of conditions, e.g.~rule \eqref{flattened:fol_dec7a_holdsAt_inertia-example}. Depending on the number of conditions involved, the inertia rule of a specific CE may be decomposed into several clauses, each corresponding to a different subset of conditions. For instance, the following two clauses are added to $\Sigma'$ by the inertia rule \eqref{flattened:fol_dec7a_holdsAt_inertia-example}:
%
%
\begin{align}
 &\begin{mysplit} \label{cnf:inertia_example1}
  \pHappens{walking(ID_1)}{T} \vee \pHappens{running(ID_1)}{T} \vee \pHappens{exit(ID_1)}{T}\ \vee\\
 \neg \pHoldsAt{meeting(ID_1,\, ID_2)}{T} \vee \pHoldsAt{meeting(ID_1,\, ID_2)}{T{+}1}
 \end{mysplit} \\
&\begin{mysplit} \label{cnf:inertia_example2}
  \neg \pClose{ID_1}{ID_2}{25}{T}  \vee  \pHappens{running(ID_1)}{T} \vee \pHappens{exit(ID_1)}{T}\ \vee\\
 \neg \pHoldsAt{meeting(ID_1,\, ID_2)}{T} \vee \pHoldsAt{meeting(ID_1,\, ID_2)}{T{+}1}
 \end{mysplit}
\end{align}
\noindent The above clauses contain literals from the termination rules of the \meeting\ CE. Often, when SDEs that lead to the partial satisfaction of the initiation/termination conditions occur, some of these clauses become trivially satisfied. For example, at time-point $10$ both persons $\mathit{ID_1}$ and $\mathit{ID_2}$ are \textit{active}, while their distance is above the threshold of $25$ pixels, i.e.~$\pClose{ID_1}{ID_2}{25}{10}{=}\mathit{False}$. Consequently, the grounding of clause \eqref{cnf:inertia_example2} at time-point $10$ is trivially satisfied for all possible worlds. Although the \textit{meeting} CE is not terminated at time-point $10$, because clause \eqref{cnf:inertia_example1} is not satisfied, the satisfaction of clause \eqref{cnf:inertia_example2} reduces the probability  of $\folholdsAt$ for the CE. This is because the inertia at time-point $10$ is now supported only by the satisfaction of the ground clause \eqref{cnf:inertia_example1}. In other words, the difference between the probabilities of worlds that violate the inertia of $\folholdsAt$ and worlds that do not, is reduced. 
 
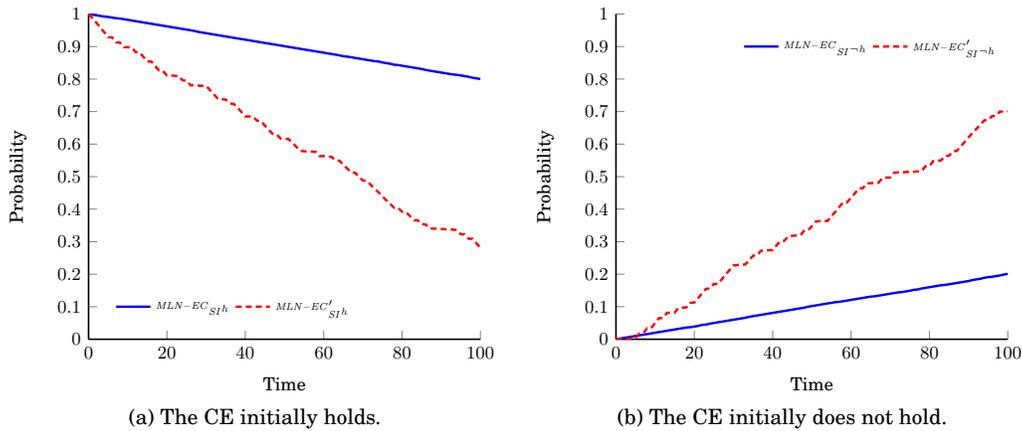
\begin{figure}[t]
\pgfplotsset{
every axis/.append style={thick},
tick label style={font=\small}
}
%
%
\ifdefined\TIKZEXTERNALIZE
 \tikzsetnextfilename{figure_eg_behaviour2_holdsAt}
\fi
\subfloat[The CE initially holds.\label{fig:eg_behaviour2_holds}]{
  \begin{tikzpicture}[scale=0.76]
    \begin{axis}[
    legend columns=2,
    legend style={ at={(0.05,0.1)},anchor=west,draw=none},
    xlabel={\small Time},
    ylabel={\small Probability},
    ymin=0, ymax=1.0, xmin=0, xmax=100, 
    ytick={0, 0.1, 0.2, 0.3, 0.4, 0.5, 0.6, 0.7, 0.8, 0.9, 1},
    thick,
    axis x line*=bottom,
    axis y line*=left]

     \pgfplotstableread{sim_one.data}\exampledata
    
     \addplot[color=blue,very thick] table[y=SIT0,x=Time] {\exampledata};
 	\addlegendentry{\tiny $\SIT$}

      \addplot[color=red,very thick,densely dashed] table[y=SIT1,x=Time] {\exampledata};
  	\addlegendentry{\tiny $\SIT'$}
  	
    \end{axis}
  \end{tikzpicture}
}
~
%
%
\ifdefined\TIKZEXTERNALIZE
  \tikzsetnextfilename{figure_eg_behaviour2_not-holdsAt}
\fi
\subfloat[The CE initially does not hold.\label{fig:eg_behaviour2_not-holds}]{
  \begin{tikzpicture}[scale=0.76]
    \begin{axis}[
    legend columns=2,
    legend style={ at={(1,0.9)},anchor=east,draw=none},
    xlabel={\small Time},
    ylabel={\small Probability},
    ymin=0, ymax=1.0, xmin=0, xmax=100, 
    ytick={0, 0.1, 0.2, 0.3, 0.4, 0.5, 0.6, 0.7, 0.8, 0.9, 1},
    thick,
    axis x line*=bottom,
    axis y line*=left]

     \pgfplotstableread{sim_zero.data}\exampledata
    
     \addplot[color=blue,very thick] table[y=SII0,x=Time] {\exampledata};
 	\addlegendentry{\tiny $\SII$}

      \addplot[color=red,very thick,densely dashed] table[y=SII1,x=Time] {\exampledata};
  	\addlegendentry{\tiny $\SII'$}

    \end{axis}
  \end{tikzpicture} 
}
\caption{In both figures no useful SDEs occur in the interval $0$ to $100$. In both $\SIT$ and $\SII$ none of the inertia clauses in $\Sigma'$ become trivially satisfied by the SDEs. However, in $\SIT'$ and $\SII'$ some inertia clauses are trivially satisfied by the SDE. In the left figure the CE holds at time $0$ with absolute certainty, while in the right figure the CE does not hold at time $0$.}
\label{fig:eg_behaviour2}
\end{figure}

To illustrate this phenomenon, consider the example cases in Figure \ref{fig:eg_behaviour2_holds} where only the rules about the inertia of $\folholdsAt$ are soft-constrained. Both $\SIT$ and $\SIT'$ cases share the same knowledge base. In the $\SIT$ case, the occurrence of SDEs causes none of the inertia clauses to become trivially satisfied. In the $\SIT'$ case, however, the SDEs are randomly generated and cause a different subset of inertia clauses to become trivially satisfied at each time-point. In both cases the probability of the CE is reduced. In contrast to $\SIT$, however, the inertia in $\SIT'$ drops more sharply, as some of the clauses in $\Sigma'$ are trivially satisfied by the given SDEs.  Additionally, the probability of the CE to hold in $\SIT'$ persists at a different level in each time-point, since different subsets of clauses become trivially satisfied each time. Similarly, in Figure \ref{fig:eg_behaviour2_not-holds} the rules about the inertia of $\neg \folholdsAt$ are soft-constrained. In contrast to $\SII$, the occurrence of SDEs leads to the partial satisfaction of the initiation conditions causing the inertia in $\SII'$ to persist with a different confidence at each time-point, increasing the probability of the CE to hold more sharply.

%
%

\vspace{0.6cm}

%
%
Having analysed the effect of softening the inertia rules, it is worth noting that in many real cases the entire knowledge base may be soft-constrained. In this case, since the rules in $\Sigma$ are soft-constrained, CEs are not being initiated or terminated with absolute certainty. At the same time, CEs do not persist with certainty, as the rules in $\Sigma'$ are also soft-constrained.

Depending on the requirements of the target application, various policies regarding the soft-constraining of the knowledge base may be adopted. This flexibility is one of the advantages of combining logic with probabilities in the proposed method. Furthermore, it should be stressed that in a typical event recognition application the knowledge base will contain a large number of clauses. The strength of a constraint imposed by a clause is also affected by the weights of other clauses with which it shares the same predicates. Due to these interdependencies, the manual setting of weights is bound to be suboptimal and cumbersome. Fortunately, the weights can be estimated automatically from training sets, using standard parameter optimisation methods. 

%

\section{Evaluation} \label{sec:experiments}

In this section we evaluate the proposed method (\MLNEC) in the domain of video activity recognition. As presented in Section \ref{sec:example}, we use the publicly available benchmark dataset of the CAVIAR project. The aim of the experiments is to assess the effectiveness of \MLNEC\ in recognising CEs that occur among people, based on imperfect CE definitions and in the presence of incomplete narratives of SDEs.

\MLNEC\ combines the benefits of logic-based representation (e.g.~direct expression of domain background knowledge) with probabilistic modeling (e.g.~uncertainty handling). For comparison purposes, we include in the experiments two approaches that are closely related to our method. First, we include the logic-based activity recognition method of \citeN{artikis2009behaviour}, which we call here \CREC. Like our method, \CREC\ employs a variant of the Event Calculus and uses the same definitions of CEs. Unlike \MLNEC, \CREC\ cannot perform probabilistic reasoning. Second, we include a pure probabilistic method that employs a linear-chain Conditional Random Field model \cite{LaffertyMP01CRF}, which we call here \LCRF. Similar to our method, \LCRF\ is a log-linear model that performs probabilistic reasoning over an undirected probabilistic network. On the other hand, \LCRF\ does not employ a logic-based representation. 

%
%
\subsection{Setup}

From the $28$ videos of the CAVIAR dataset, we have extracted $19$ sequences that are annotated with the \meeting\ and/or \moving\ CEs. The rest of the sequences in the dataset are ignored, as they do not contain examples of the two target CEs. Out of $19$ sequences, $8$ are annotated with both \moving\ and \meeting\ activities, $9$ are annotated only with \moving\ and $2$ only with \meeting. The total length of the extracted sequences is $12869$ frames.  Each frame is annotated with the occurrence or not of a CE and is considered an example instance. The whole dataset contains  a total of $25738$ annotated example instances. There are $6272$ example instances in which \moving\ occurs and $3622$ in which \meeting\ occurs. For both CEs, consequently, the number of negative examples is significantly larger than the number of positive examples, $19466$ for moving and $22116$ for meeting.

%
%
The input of all three methods consists of a sequence of SDEs, i.e.~\textit{active, inactive, walking, running, enter} and \textit{exit}. In both \MLNEC\ and \LCRF, the spatial constraints $\folclose$ and $\folOM$ are precomputed and their truth value is provided as input. In situations where no event occurs or the distance of the involved persons is above the highest predefined threshold, the tags \textit{none} and \textit{far} are given to \LCRF, respectively.

%
%
The output of the \CREC\ method consists of a sequence of ground $\folholdsAt$ predicates, indicating which CEs are recognised. Since \CREC\ performs crisp reasoning, all CEs are recognised with absolute certainty. On the other hand, the output of both probabilistic methods depends on the inference type, i.e.~maximum a-posteriori inference (MAP) or marginal. Given a sequence of SDEs, MAP inference outputs the most probable instantiations of CEs for all time-points. On the other hand, marginal inference outputs CEs associated with some probability for all time-points.

Table \ref{tbl:training-set} presents the structure of the training sequences for the probabilistic methods. In particular, Table \ref{tbl:training-set-MLN} shows an example training sequence for \MLNEC. Each sequence is composed of input SDEs (ground $\folhappens$), precomputed spatial constraints between pairs of people (ground $\folclose$ and $\folOM$), as well as the corresponding CE annotations (ground $\folholdsAt$). Negated predicates in the training sequence state that the truth value of the corresponding predicate is \textit{False}. Table \ref{tbl:training-set-CRF} shows the equivalent input training data for \LCRF. The random variables \textit{Person A} and \textit{Person B} represent the events that the two persons may perform at each time-point and variables \textit{Close} and \textit{Orientation Move} represent the spatial constraints between the two persons. Similar to \MLNEC, \LCRF\ does not contain any hidden variables between input SDEs and output CEs. As a result, training is fully supervised for both methods.

\begin{table}
\tbl{Example training sets for CE \moving.\label{tbl:training-set}}{
\begin{minipage}{1\textwidth}
\centering
\subfloat[Input narrative for \MLNEC.\label{tbl:training-set-MLN}]{
\begin{tabular}{|l|r|}
\hline
\multicolumn{1}{|c|}{Simple Derived Events} & \multicolumn{1}{c|}{Composite Events} \\\hline
\multicolumn{1}{|c|}{$\dots$} & \multicolumn{1}{c|}{$\dots$} \\
$\pHappens{walking(id_1)}{100}$ &  \\
 $\pHappens{walking(id_2)}{100}$ & $\pHoldsAt{moving(id_1,\, id_2)}{100}$ \\
 $\pOrientationMove{id_1}{id_2}{100}$ & $\pHoldsAt{moving(id_2,\, id_1)}{100}$ \\
 $\pClose{id_1}{id_2}{24}{100}$ &  \\
  \multicolumn{1}{|c|}{$\dots$} & \multicolumn{1}{c|}{$\dots$} \\
$\pHappens{active(id_1)}{101}$ & \\
 $\pHappens{walking(id_2)}{101}$ & $\pHoldsAt{moving(id_1,\, id_2)}{101}$ \\ 
 $\pOrientationMove{id_1}{id_2}{101}$ & $\pHoldsAt{moving(id_2,\, id_1)}{101}$ \\
 $\neg \pClose{id_1}{id_2}{24}{101}$ &  \\
  \multicolumn{1}{|c|}{$\dots$} & \multicolumn{1}{c|}{$\dots$} \\
 $\pHappens{walking(id_1)}{200}$ &  \\
 $\pHappens{running(id_2)}{200}$ &  $\neg \pHoldsAt{moving(id_1,\, id_2)}{200}$ \\
 $\neg \pOrientationMove{id_1}{id_2}{200}$ &  $\neg \pHoldsAt{moving(id_2,\, id_1)}{200}$ \\
 $\neg \pClose{id_1}{id_2}{24}{200}$ &   \\
  \multicolumn{1}{|c|}{$\dots$} & \multicolumn{1}{c|}{$\dots$} \\\hline
\end{tabular}
}

\hfill

\subfloat[Input sequence for \LCRF.\label{tbl:training-set-CRF}]{
\begin{tabular}{|l|l|l|l|r|}
\hline
    \multicolumn{1}{|c|}{Person A} & \multicolumn{1}{c|}{Person B} & \multicolumn{1}{c|}{Close} & \multicolumn{1}{c|}{Orientation Move} &\multicolumn{1}{c|}{Composite Events} \\\hline
     \multicolumn{1}{|c|}{$\dots$} & \multicolumn{1}{c|}{$\dots$} & \multicolumn{1}{c|}{$\dots$} & \multicolumn{1}{c|}{$\dots$} & \multicolumn{1}{c|}{$\dots$} \\\hline
     walking & walking & 24 & True & Moving\\\hline
     active & walking & Far & True & NotMoving\\\hline
     \multicolumn{1}{|c|}{$\dots$} & \multicolumn{1}{c|}{$\dots$} & \multicolumn{1}{c|}{$\dots$} & \multicolumn{1}{c|}{$\dots$} & \multicolumn{1}{c|}{$\dots$} \\\hline
     walking & running & Far & False & NotMoving\\\hline
     \multicolumn{1}{|c|}{$\dots$} & \multicolumn{1}{c|}{$\dots$} & \multicolumn{1}{c|}{$\dots$} & \multicolumn{1}{c|}{$\dots$} & \multicolumn{1}{c|}{$\dots$} \\\hline
  \end{tabular}
}
\end{minipage}
}
\begin{tabnote}
\Note{}{Table \ref{tbl:training-set-MLN} shows a training set for \MLNEC. The first column is composed of a narrative of SDEs and precomputed spatial constraints for \MLNEC, while the second column contains the CE annotation in the form of ground $\folholdsAt$ predicates. Table \ref{tbl:training-set-CRF} shows the equivalent training set for \LCRF.  Columns \textit{Person A} to \textit{Orientation Move} contain the input SDEs and spatial constraints, while the last column contains the annotation.}
\end{tabnote}
\end{table}

To estimate the weights in \LCRF, we use the quasi-Newton optimisation algorithm L-BFGS \cite{byrd1994representations}. For MAP and marginal inference we use the  \textit{Viterbi} and \textit{Forward-Backward} algorithms \cite{culotta2004confidence,sutton2006introduction}. In \MLNEC\ we use the second-order Diagonal Newton method of  \citeN{lowd2007efficient} and perform marginal inference with the \textit{MC-SAT} algorithm \cite{poon2006sound}, taking $1000$ samples for each sequence. We additionally perform max-margin training and MAP inference, using the method of \citeN{huynh2009max}. In the experiments we use the open-source software packages {\sc Alchemy}~\shortcite{alchemy05} and {\sc CRF++}\footnote{\url{http://code.google.com/p/crfpp}}.

\MLNEC\ is tested under three different scenarios ($\HI$, $\SIT$ and $\SI$, see Table \ref{tbl:inertia-configurations} for a description). In all three variants of \MLNEC, the rules in $\Sigma$ are soft-constrained while the inertia rules in $\Sigma'$ are either soft or hard.

Throughout the experimental analysis, the results for marginal inference are presented in terms of $F_1$ score for threshold values ranging between $0.0$ and $1.0$. Any CE with probability above the threshold is considered to be recognised. A snapshot of the performance using the threshold value $0.5$, is presented in terms of true positives (TP), false positives (FP), false negatives (FN), precision, recall and $F_1$ score. Additionally, the overall performance for marginal inference is measured in terms of area under precision-recall curve (AUPRC). The number of true negatives in our experiments is significantly larger than the number of true positives. Similar to $F_1$ score, precision and recall, the AUPRC is insensitive to the number of true negatives. The evaluation results using MAP inference are presented in terms of true positives (TP), false positives (FP), false negatives (FN), precision, recall and $F_1$. All reported experiment statistics are micro-averaged over the instances of recognised CEs in the $10$ folds.

\begin{table}[h]
\centering
\small
\tbl{Variants of \MLNEC, using hard and soft inertia rules in $\Sigma'$.\label{tbl:inertia-configurations}}{
\begin{tabular}{|l|l|}
\hline
\multicolumn{1}{|c|}{Scenarios} & \multicolumn{1}{c|}{Description} \\\hline
$\HI$  & All inertia rules in $\Sigma'$ are hard-constrained.  \\\hline
\multirow{2}{*}{$\SIT$} & The inertia rules of $\folholdsAt$ are soft-constrained, while the rest of $\Sigma'$\\
       &  remains hard-constrained. \\\hline
$\SI$  & All inertia rules in $\Sigma'$ are soft-constrained. \\\hline
\end{tabular}
}
\end{table}

%
%
\subsection{The Methods Being Compared}

%
%
Both \CREC\ and \MLNEC\ employ a logic-based representation, implement a variant of the Event Calculus and contain equivalent definitions of CEs. The CE definitions of \meeting\ and \moving\ of \CREC\ are translated into first-order logic for \MLNEC\ using the formulation proposed in Section \ref{sec:EventCalculus}. The definition of \meeting\ is given by formulas \eqref{dsa:init_meet1}{--}\eqref{dsa:term_meet3}, while that of \moving\ is given by formulas \eqref{dsa:init_move}{--}\eqref{dsa:term_move5}. In contrast to \CREC, each clause in \MLNEC\ may be associated with a weight value, indicating a degree of confidence.

%
%

Similar to \MLNEC, \LCRF\ is a discriminative probabilistic graphical model. The relationship among CEs at successive time-points is modelled as a Markov network, conditioned on the input evidence of SDEs. A CE at any time-point in the sequence is represented by a Boolean random variable, stating whether the CE holds or not. For example, the random variables representing the \moving\ CE may take either the tag value \textit{Moving} or \textit{NotMoving} at some time-point in the sequence.

However, there are also several differences between the two probabilistic methods. In \LCRF, the input SDEs and the spatial constraints are represented by multivariate random variables. For instance, the input SDEs for a particular person are represented by a single random variable that can take any SDE tag value, e.g.~\textit{active, inactive, walking}, etc. The relationship among random variables is defined by two types of features. The former type associates input SDEs and spatial constraints with output CEs at the same time-point, creating features for all possible instantiations. The latter type associates successive CEs, in order to form linear chains. In particular, features are constructed for each possible pair of CE instantiations at successive time-points. All features in \LCRF\ are associated with weights and thus all relationships are soft-constrained. 

On the other hand, \MLNEC\ employs a logic-based representation and all features are produced from ground clauses. Domain knowledge is combined with the Event Calculus axioms, in order to form the structure of the network. For example, the relations between successive CE instantiations are formed by the inertia axioms and the corresponding initiation and termination rules. Moreover, the \MLNEC\ provides control over the behaviour of CE persistence, by allowing the inertia clauses to be defined either as soft or as hard constraints. The probabilistic inference in \MLNEC\ can deal with both deterministic (i.e.~hard-constrained clauses) and probabilistic (i.e.~soft-constrained clauses) dependencies, as well as arbitrary structured networks. On the other hand, the structural simplicity of \LCRF\ allows for specialised and significantly faster inference and learning methods.

%
%
\subsection{Experimental Results of the \CREC} \label{sec:results-Crisp}

The CE definitions are domain dependent rules that together with the domain-independent
axioms of the Event Calculus represent common sense and background knowledge. This knowledge may deviate from that implied by an annotated dataset, resulting in errors when recognising events. Therefore, regarding the annotation of the datasets, the CE definitions are imperfect. Such issues can be clearly shown by analysing the performance of \CREC, which uses the CE definitions also used in \MLNEC\ and does not involve the representation of probabilistic knowledge.

As shown in Figure \ref{fig:fmeasure}, \CREC\ achieves a similar $F_1$ score for both activities. However, in terms of precision and recall the situation is quite different, revealing two different cases of imperfect CE definitions. The precision for \moving\ is $22$ percentage points higher than that of \meeting. The opposite holds for recall, with the recall for \moving\ being $21.6$ percentage points lower than that of \meeting.  The lower recall values for \moving\ indicate a larger number of unrecognised \moving\ activities (FN). In some example sequences \moving\ is being  initiated late, producing many false negatives. Additionally, the termination rules of \moving\ cause the CE to be prematurely terminated in some cases. For example, when the distance of two persons that are moving together becomes greater than $34$ pixels for a few frames, rule \eqref{dsa:term_move1} terminates \moving. On the other hand, compared to \moving, the definition of \meeting\ results in a larger number of erroneously recognised \meeting\ activities (FP). The initiation rule \eqref{dsa:init_meet2}, for example, causes the \meeting\ activity to be initiated earlier than it should. 

Another issue caused by the definitions of \meeting\ and \moving\ is that the two CEs may overlap. According to rules \eqref{dsa:init_meet1}{--}\eqref{dsa:term_move5}, the initiation of \moving\ does not cause the termination of \meeting. Consider, for example, a situation where two people meet for a while and thereafter they move together. During the interval in which \moving\ is detected, \meeting\ will also remain detected, as it is not terminated and the law of inertia holds. However, according to the annotation of the CAVIAR team these activities do not happen concurrently. Furthermore, \meeting\ appears to be annotated in cases where the relative distance of both interacting persons is greater than that used in the initiation rules \eqref{dsa:init_meet1} and \eqref{dsa:init_meet2}.

On the other hand, the background knowledge may describe situations that are not included in the dataset.  For example, by allowing the \meeting\ CE to be initiated when the two persons are not very close to each other, one may achieve better results in this subset of the dataset, but erroneously recognise the  \meeting\ CE in other situations, e.g.~people passing by each other. Additionally, the domain-independent property of inertia, which is included in the background knowledge, helps the method to continue to recognise the occurrence of a CE even when the narrative of SDEs is temporally incomplete, e.g.~due to camera failures.

%
%
\subsection{Experimental Results of the Probabilistic Methods}

The experiments for the probabilistic methods are organised into two tasks\footnote{MLN and \LCRF\ formatted version of the dataset, CE definitions of \MLNEC, template files of \LCRF and the results of both probabilistic methods can be found at: \url{http://www.iit.demokritos.gr/~anskarl/pub/mlnec}}. In the first task, both probabilistic methods are trained discriminitavely and their performance is evaluated using $10$-fold cross-validation. In the second task, we asses the value of inertia by erasing  input evidence from randomly chosen successive time-points. We use the trained models from the first task, but the testing sequences are incomplete narratives of SDEs.

\subsubsection{Task I}

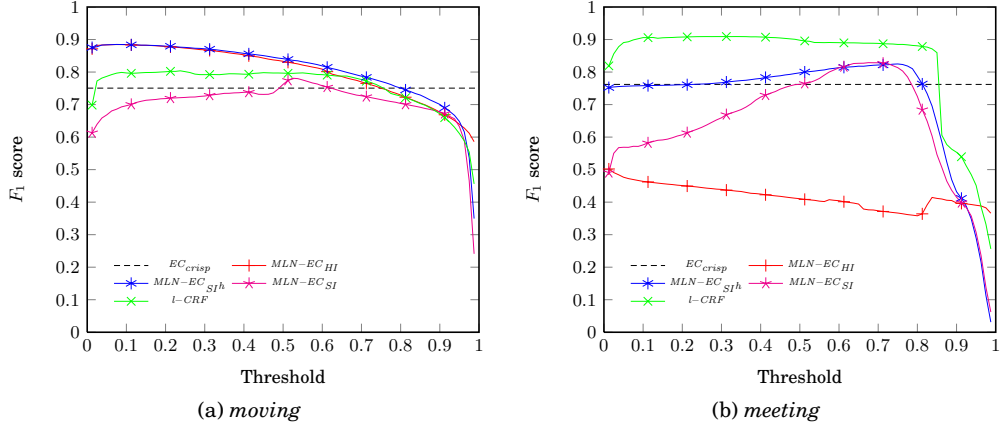
\begin{figure}[t]
\centering
\tikzset{every mark/.append style={scale=2}}
\pgfplotsset{
every axis/.append style={thick},
tick label style={font=\small}
}
%
%
\subfloat[\moving\label{fig:fmeasure-moving}]{
\ifdefined\TIKZEXTERNALIZE
  \tikzsetnextfilename{figure_task1_fmeasure-moving}
\fi
\begin{tikzpicture}[scale=0.76]
    \begin{axis}[
    legend columns=2,
    legend style={ at={(0.05,0.15)},anchor=west,draw=none},
    xlabel={\small Threshold},
    ylabel={\small $F_{1}$ score},
    ymin=0, ymax=1.0, xmin=0, xmax=1.0, 
    ytick={0, 0.1, 0.2, 0.3, 0.4, 0.5, 0.6, 0.7, 0.8, 0.9, 1}, 
    xtick={0, 0.1, 0.2, 0.3, 0.4, 0.5, 0.6, 0.7, 0.8, 0.9, 1},
    mark size = 1.5]

    \pgfplotstableread{marginal.move-fmeasure.data}\fmeasuredata

    \addplot[color=black,thin,densely dashed] coordinates { (0,0.7505618) (1,0.7505618)};
	\addlegendentry{\tiny $\ECLP$}
  
    \addplot[color=red,mark=+,thin,mark repeat={8}] table[y=PEC_HI,x=Threshold] {\fmeasuredata};
	\addlegendentry{\tiny $\HI$}
	
    \addplot[color=blue,mark=asterisk,thin,mark repeat={8}] table[y=PEC_SIT,x=Threshold] {\fmeasuredata};
	\addlegendentry{\tiny $\SIT$}
    
    \addplot[color=magenta,mark=star,thin,mark repeat={8}] table[y=PEC_SI,x=Threshold] {\fmeasuredata};
	\addlegendentry{\tiny $\SI$}
	
    \addplot[color=green,mark=x,thin,mark repeat={8}] table[y=crfpp_move,x=Threshold] {\fmeasuredata};
	\addlegendentry{\tiny $\CRF$}

    \end{axis}
\end{tikzpicture}
}
~
%
%
\subfloat[\meeting\label{fig:fmeasure-meeting}]{
\ifdefined\TIKZEXTERNALIZE
  \tikzsetnextfilename{figure_task1_fmeasure-meeting}
\fi
\begin{tikzpicture}[scale=0.76]
    \begin{axis}[
    legend columns=2,
    legend style={ at={(0.05,0.15)},anchor=west,draw=none},
    xlabel={\small Threshold},
    ylabel={\small $F_1$ score},
    ymin=0, ymax=1.0, xmin=0, xmax=1, 
    ytick={0, 0.1, 0.2, 0.3, 0.4, 0.5, 0.6, 0.7, 0.8, 0.9, 1}, 
    xtick={0, 0.1, 0.2, 0.3, 0.4, 0.5, 0.6, 0.7, 0.8, 0.9, 1},
    mark size = 1.5]
    \pgfplotstableread{marginal.meet-fmeasure.data}\fmeasuredata
    
    \addplot[color=black,thin,densely dashed] coordinates { (0,0.7619867) (1,0.7619867)};
	\addlegendentry{\tiny $\ECLP$}
  
    \addplot[color=red,mark=+,thin,mark repeat={8}] table[y=PEC_HI,x=Threshold] {\fmeasuredata};
	\addlegendentry{\tiny $\HI$}
     
      \addplot[,color=blue,mark=asterisk,thin,mark repeat={8}] table[y=PEC_SIT,x=Threshold] {\fmeasuredata};
	\addlegendentry{\tiny $\SIT$}
%
    \addplot[color=magenta,mark=star,thin,mark repeat={8}] table[y=PEC_SI,x=Threshold] {\fmeasuredata};
	\addlegendentry{\tiny $\SI$}
	
    \addplot[color=green,mark=x,thin,mark repeat={8}] table[y=crfpp_meet,x=Threshold] {\fmeasuredata};
	\addlegendentry{\tiny $\CRF$}
    \end{axis}
\end{tikzpicture}
}
\caption{$F_1$ scores using different threshold values for the \moving\ and \meeting\ CE.}
\label{fig:fmeasure}
\end{figure}

In contrast to $\ECLP$, \LCRF\ is a probabilistic model and cannot employ background knowledge in the form of common sense rules. The recognition of a CE is affected from all input SDEs that are detected  at a particular time-point, as well as from the adjacent CEs that have been recognised. The parameters of the model are estimated from the training set and thus the model is completely data driven. Compared to $\ECLP$, \LCRF\ achieves better performance for both \moving\ and \meeting. 
Using marginal inference, \LCRF\ gives higher $F_1$ scores for most threshold values, as shown in Figures \ref{fig:fmeasure-moving} and \ref{fig:fmeasure-meeting}. For threshold $0.5$, the $F_1$ score is higher than of $\ECLP$ by $4.6$ and $13.5$ percentage points for \moving\ and \meeting, respectively (see Tables \ref{tbl:wlearning-moving} and \ref{tbl:wlearning-meeting}). The recall of \moving\ is higher by $18.4$ percentage points, while the precision is lower by $13.7$ percentage points. \LCRF\ recognises a larger number of \moving\ activities, increasing the number of both true and false positives. As noted in Section \ref{sec:results-Crisp}, this can be achieved by looser spatial constraints. Recall and precision scores for \meeting\ are higher by $1.7$ and $23.7$ percentage points, respectively. 
In the case of MAP inference, \LCRF\ gives almost the same $F_1$ score as $\ECLP$ for \moving\ (see Tables \ref{tbl:wlearning-moving-mm} and \ref{tbl:wlearning-meeting-mm}). 
In general \LCRF\ outperforms $\ECLP$ in both CE. Unlike $\ECLP$, \LCRF\ interrupts the recognition of \meeting\ when \moving\ starts. 

%
%

%
%
\begin{table}
\tbl{Results for the \moving\ and \meeting\ CE using marginal inference.}{
 \begin{minipage}{1\textwidth}
 \centering
%
%
\begin{tabular}{|l|} 
\hline
\multicolumn{1}{|c|}{Method} \\\hline
$\ECLP$ \\\hline
$\HI$ \\\hline
$\SIT$ \\\hline
$\SI$ \\\hline
$\CRF$ \\\hline
\end{tabular}
\hspace*{-0.6em}
\subfloat[\moving, threshold $0.5$.\label{tbl:wlearning-moving}]{
\begin{tabular}{|r|r|r|r|r|r|r|} 
\hline
 \multicolumn{1}{|c|}{TP} & \multicolumn{1}{c|}{TN} & \multicolumn{1}{c|}{FP} & \multicolumn{1}{c|}{FN} & \multicolumn{1}{c|}{Precision} & \multicolumn{1}{c|}{Recall} & \multicolumn{1}{c|}{F$_{1}$ score} \\\hline
 4008 & 19086 & 400 & 2264 & \textbf{0.9093} & 0.6390 & 0.7506\\\hline
 5121 & 18584 & 902 & 1151 & 0.8502 & 0.8165 & 0.8330 \\\hline
 5233 & 18542 & 944 & 1039 & 0.8472 & \textbf{0.8343} & \textbf{0.8407} \\\hline
 4938 & 17760 & 1726 & 1334 & 0.7410 & 0.7873 & 0.7635 \\\hline
 5160 & 17964 & 1522 & 1112 & 0.7722 & 0.8227 & 0.7967 \\\hline
\end{tabular}
}
\hspace*{-0.6em}
\subfloat[\moving\label{tbl:auprc-moving}]{
\begin{tabular}{|r|} 
\hline
\multicolumn{1}{|c|}{AUPRC} \\\hline
\\\hline
\textbf{0.8847} \\\hline
0.8597 \\\hline
0.8280 \\\hline
0.8358 \\\hline
\end{tabular}
}

\hfill

%
%
\begin{tabular}{|l|} 
\hline
\multicolumn{1}{|c|}{Method} \\\hline
$\ECLP$ \\\hline
$\HI$ \\\hline
$\SIT$ \\\hline
$\SI$ \\\hline
$\CRF$ \\\hline
\end{tabular}
\hspace*{-0.6em}
\subfloat[\meeting, threshold $0.5$. \label{tbl:wlearning-meeting}]{
\begin{tabular}{|r|r|r|r|r|r|r|}
\hline
 \multicolumn{1}{|c|}{TP} & \multicolumn{1}{c|}{TN} & \multicolumn{1}{c|}{FP} & \multicolumn{1}{c|}{FN} & \multicolumn{1}{c|}{Precision} & \multicolumn{1}{c|}{Recall} & \multicolumn{1}{c|}{F$_{1}$ score} \\\hline
 3099 & 20723 & 1413 & 523 & 0.6868 & 0.8556 & 0.7620 \\\hline
 1284 & 20786 & 1350 & 2338 & 0.4875 & 0.3545 & 0.4105 \\\hline
 3060 & 21147 & 989 & 562 & 0.7557 & 0.8448 & 0.7978 \\\hline
 3052 & 20800 & 1336 & 570 & 0.6955 & 0.8426 & 0.7620 \\\hline
 3160 & 21874 &  262 &  462 & \textbf{0.9234} & \textbf{0.8724} & \textbf{0.8972} \\\hline
\end{tabular}
}
\hspace*{-0.6em}
\subfloat[\meeting\label{tbl:auprc-meeting}]{
\begin{tabular}{|r|} 
\hline
\multicolumn{1}{|c|}{AUPRC} \\\hline
 \\\hline
0.5160 \\\hline
0.7559 \\\hline
0.7730 \\\hline
\textbf{0.8937} \\\hline
\end{tabular}
}
\end{minipage}
}
\end{table}

\begin{table}
\tbl{Results for the \moving\ and \meeting\ CE using MAP inference}{
\begin{minipage}{1\textwidth}
\centering
%
%
\subfloat[\moving\label{tbl:wlearning-moving-mm}]{
 \begin{tabular}{|l|r|r|r|r|r|r|r|} 
\hline
   \multicolumn{1}{|c|}{Method} & \multicolumn{1}{c|}{TP} & \multicolumn{1}{c|}{TN} & \multicolumn{1}{c|}{FP} & \multicolumn{1}{c|}{FN} & \multicolumn{1}{c|}{Precision} & \multicolumn{1}{c|}{Recall} & \multicolumn{1}{c|}{F$_{1}$ score} \\\hline
  $\ECLP$ & 4008 & 19086 & 400 & 2264 & \textbf{0.9093} & 0.6390 & 0.7506\\\hline
  $\HI$   & 5598 & 18358 & 1128 & 674 & 0.8323 & 0.8925 & 0.8614 \\\hline
  $\SIT$  & 5902 & 18398 & 1088 & 370 & 0.8443 & \textbf{0.9410} & \textbf{0.8901} \\\hline
  $\SI$   & 4040 & 17911 & 1575 & 2232 & 0.7195 & 0.6441 & 0.6797 \\\hline
  $\CRF$	& 4716 & 17848 & 1638 & 1556 & 0.7422 & 0.7519 & 0.7470 \\\hline
\end{tabular}
}
\hfill
%
%
\subfloat[\meeting\label{tbl:wlearning-meeting-mm}]{
  \begin{tabular}{|l|r|r|r|r|r|r|r|}
\hline
  \multicolumn{1}{|c|}{Method} & \multicolumn{1}{c|}{TP} & \multicolumn{1}{c|}{TN} & \multicolumn{1}{c|}{FP} & \multicolumn{1}{c|}{FN} & \multicolumn{1}{c|}{Precision} & \multicolumn{1}{c|}{Recall} & \multicolumn{1}{c|}{F$_{1}$ score} \\\hline
  $\ECLP$ & 3099 & 20723 & 1413 & 523 & 0.6868 & 0.8556 & 0.7620 \\\hline
  $\HI$ 	& 3099 & 20739 & 1397 & 523 & 0.6893 & 0.8556 & 0.7635 \\\hline
  $\SIT$ 	& 3067 & 21825 & 311 & 555 & 0.9079 & 0.8468 & 0.8763 \\\hline
  $\SI$ 	& 1083 & 21641 & 495 & 2539 & 0.6863 & 0.2990 & 0.4165 \\\hline
  $\CRF$  & 3154 & 21906 & 230 & 468 & \textbf{0.9320} & \textbf{0.8708} & \textbf{0.9004}  \\\hline
  \end{tabular}
}
\end{minipage}
}
\end{table}

%
%
In the first \MLNEC\ scenario, indicated by $\HI$, rules in $\Sigma$ are soft-constrained, i.e.~they are associated with a weight value after training. Those weights control the certainty with which a CE holds when its initiation or termination conditions are satisfied. The rules in $\Sigma'$, however, remain hard-constrained and thus the behaviour of inertia for both CEs is preserved deterministically and cannot be adjusted. 
Compared to \CREC, $\HI$ achieves a higher $F_1$ score using marginal inference for the \moving\ CE, for most threshold values (see Figure \ref{fig:fmeasure-moving}). For threshold $0.5$, the recall of \MLNEC\ is higher by $17.7$ percentage points than \CREC\ while precision is lower by $6$ points, leading the $F_1$ score of \MLNEC\ to be higher than \CREC\ by $8.2$ points (Table \ref{tbl:wlearning-moving}). This improvement in recall is caused by the weights that are learned for the termination conditions, which prevent the \moving\ CE from terminating prematurely. Compared to \LCRF, $\HI$ achieves better $F1$ scores for many thresholds and higher AUPRC by $4.8$ percentage points (see Tables \ref{tbl:wlearning-moving} and \ref{tbl:auprc-moving}). Using MAP inference, $\HI$ achives higher $F_1$ score by $11$ percentage points than both $\ECLP$ and \LCRF\ (see Table \ref{tbl:wlearning-moving-mm}). Compared to \CREC, the recall of $\HI$ is improved by $25.3$ percentage points, while its precision drops by $7$ percentage points. $\HI$ achieves higher recall and precision scores than \LCRF, by $14$ and $9$ percentage points, respectively.
However, in the case of \meeting, $\HI$ performs worse than \CREC\ and \LCRF\ in marginal inference, as shown in Figure \ref{fig:fmeasure-meeting} and Table \ref{tbl:wlearning-meeting}. The combination of hard-constrained inertia rules with the fact that \meeting\ does not terminate when \moving\ starts, push the weights of the initiation rules to very low values during training. This situation results in many unrecognised \meeting\ instances and low precision and recall values. Max-margin training in this scenario is not affected as much as the Diagonal Newton weight learning method, leading to a model with similar behaviour and performance as \CREC, as shown in Table \ref{tbl:wlearning-meeting-mm}.

%
%
In the $\SIT$ scenario, while $\Sigma$ remains soft-constrained, the inertia rules of $\folholdsAt$ in $\Sigma'$ are also soft-constrained. As a result, the probability of a CE tends to decrease, even when the required termination conditions are not met and nothing relevant is happening. This scenario is more suitable to our target activity recognition task and \MLNEC\ learns a model with a high $F_1$ score for both CEs. In order to explain the effect of soft constraining the inertia of $\folholdsAt$, we will use again the example of \meeting\ being recognised and thereafter \moving\ being also recognised. Since \meeting\ is not terminated, it continues to hold and overlaps with \moving. During the overlap, all occurring SDE are irrelevant with respect to \meeting\ and cannot cause any initiation or termination. As a result, the recognition probability of \meeting\ cannot be reinforced, by re-initiation. As shown in Section \ref{sec:inertia}, in such circumstances the recognition probability of \meeting\ gradually decreases. 

For the \moving\ CE, the performance of $\SIT$ using marginal inference is similar to $\HI$, as shown in Figure \ref{fig:fmeasure-moving} and Table \ref{tbl:wlearning-moving}. Using a threshold of $0.5$, recall is higher than that of $\ECLP$ by $19.5$ percentage points while precision is lower by $6.2$ points, resulting in $9$ points increase in $F_1$ measure. Compared to \LCRF, $\HI$ achieves higher $F_1$ scores for many thresholds (see Figure \ref{fig:fmeasure-moving}), as well as higher AUPRC by $2.4$ percentage points (see Table \ref{tbl:auprc-moving}). In the case of MAP inference, $\SIT$ increases further its performance (Table \ref{tbl:wlearning-moving-mm}). Compared to \CREC, recall is higher by $30$ percentage points and precision drops only by $6.5$ percentage points, resulting in $14$ percentage points higher $F_1$ score. $\SIT$ achieves higher $F_1$ score, precision and recall than \LCRF\ by $14.3$, $10$ and $18.9$ percentage points, respectively. 

For the \meeting\ CE, the performance of $\SIT$ using marginal inference is significantly better than of $\HI$ (see Figure \ref{fig:fmeasure-meeting}). At the $0.5$ threshold value, precision increases by $6.9$ percentage points over \CREC, while recall falls by only $1$ point and thus $F_1$ is higher by $3.5$ points (Table \ref{tbl:wlearning-meeting}). However, the $F_1$ scores of $\SIT$ remain lower than those of \LCRF\ and its AUPRC is lower by $13.8$ points (Table \ref{tbl:auprc-meeting}). Using MAP inference, the performance of \MLNEC\ improves, but remains worse than \LCRF, by $2.4$ percentage points in terms of $F_1$ (Table \ref{tbl:wlearning-meeting-mm}). $\SIT$ performs similarly to \LCRF\ for \meeting. $\SIT$ misses the recognition of \meeting\ at time-points where the persons involved are not sufficiently close to each other according to the initiation rules \eqref{dsa:init_meet1} and \eqref{dsa:init_meet2}, i.e.~they have a distance greater than $25$ pixels.

%
%
Finally, in the $\SI$ scenario, the entire knowledge base is soft-constrained. The weights in $\Sigma$ allow full control over the confidence that a CE holds when its initiation or termination conditions are met. Additionally, by soft-constraining the rules in $\Sigma'$, $\SI$ provides fully probabilistic inertia. However, this flexibility comes at the cost of an increase in the number of parameters to be estimated from data, as all clauses in the knowledge base are now soft-constrained. As a result, $\SI$ requires more training data. Using marginal inference $\SI$ performs almost the same as \CREC\ in terms of $F_1$ score, but worse than $\SIT$ for both CEs. In the case of MAP inference, $\SI$ performs worse than \CREC, as well as \LCRF\ for both CEs.

%
%
The three variants of $\PEC$ used in the above experiments, illustrated the potential benefits of softening the constraints and performing probabilistic inference in event recognition. In contrast to \CREC, an important characteristic of \MLNEC\ is that multiple successive initiations (or terminations) can increase (or decrease) the recognition probability of a CE. By softening the CE definitions, premature initiation or termination can be avoided. In particular, as explained above, the weights learned for the termination definitions of the \moving\ CE reduced the number of unrecognised \moving\ activities. 

The choice of rules to be softened affects significantly the event recognition accuracy. In the presented application, for example, the $\SIT$ setting is the best choice, as softening the inertia of $\folholdsAt$ provides advantages over crisp recognition. Depending on the target application and the availability of training data, different types of inertia rules may be softened, varying the inertia behaviour from deterministic to completely probabilistic. This is a key feature of \MLNEC.

\subsubsection{Task II}

An important difference between the proposed logic-based approach and its purely data-driven competitors, such as \LCRF, is that it is less dependent on the peculiarities of the training data. By incorporating background knowledge about the task and the domain, in terms of logic, it can make the recognition process more robust to variations in the data. Such variations are very common in practice, particularly in dynamic environments, such as the ones encountered in event recognition. The common assumption made in machine learning that the training and test data share the same statistical properties is often violated in these situations. In order to measure the benefits that we can gain by the combination of background knowledge and learning in MLNs, we examine  the recognition of CEs  under such situations. In particular, in this second task, we are comparing the performance of \MLNEC\ and \LCRF\ in cases where the input evidence is missing from a number of successive time-points in the test data. The models are not retrained and thus their weights remain as they were estimated in Task I. 

The incomplete test sequences are generated artificially by erasing successive input SDEs and associated spatial relations at random starting points. We have generated two variants of incomplete test sequences, one containing random blank intervals of length $10$ and $20$ time-points respectively. The starting time-points of the blank intervals are chosen randomly with probability $0.01$, drawn from a uniform distribution. Since the target CEs require the interaction of two persons, erasing events involving a single person cannot affect the performance of the recognition methods that we compare. Therefore, blank intervals are created only from time-points where both persons are involved in some SDEs. This process of artificially generating incomplete test sequences is repeated five times, generating corresponding test sets.

The recognition performance can be affected in various ways, depending on the position where evidence is erased  from  the test sequences. For example, when the beginning of an activity is erased, its recognition will be delayed, increasing the number of false negatives. Similarly, erasing information at the end of an activity will delay the termination of a CE, resulting in a higher number of false positives. In cases where missing information appears during an activity, the recognition of the CE may be interrupted, increasing the number of false negatives. The recognition performance may even be improved in some cases, due to the removal of SDEs that would cause false positives or false negatives.

Out of all the methods examined in Task I the performance of \CREC\ and $\HI$ is bound to be affected less by the missing data, due to the use of deterministic inertia. This is because the erased evidence will often be in the interval that a CE is (not) being recognised. In these cases, the erased evidence will not affect the inertia of the CE and the CE will remain (not) recognised. \CREC\ and $\HI$ are only affected  when the evidence is erased at the beginning or at the end of an activity, which is less frequent. For this reason, we chose to exclude these two methods from Task II. Furthermore we exclude the $\SI$, which performed significantly worse than $\SIT$ in Task I. Therefore, in this task we compare only $\SIT$ against \LCRF.

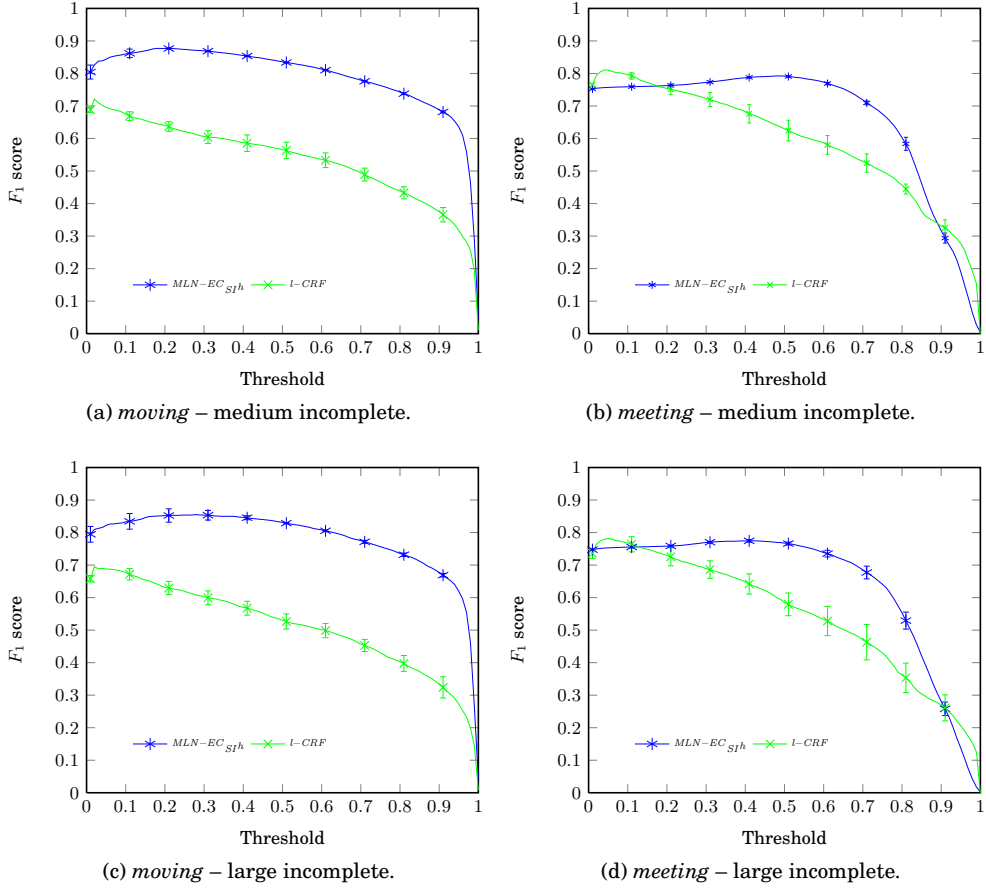
\begin{figure}
\centering
\tikzset{every mark/.append style={scale=2}}
\pgfplotsset{
every axis/.append style={thick},
tick label style={font=\small}
}
%
%
\subfloat[\moving\ -- medium incomplete.\label{fig:task2_l10_fmeasure-moving}]{
\ifdefined\TIKZEXTERNALIZE
  \tikzsetnextfilename{figure_task2_l10_fmeasure-moving}
\fi
\begin{tikzpicture}[scale=0.76]
    \begin{axis}[
    legend columns=2,
    legend style={ at={(0.1,0.15)},anchor=west,draw=none},
    xlabel={\small Threshold},
    ylabel={\small $F_{1}$ score},
    ymin=0, ymax=1.0, xmin=0, xmax=1.0, 
    ytick={0, 0.1, 0.2, 0.3, 0.4, 0.5, 0.6, 0.7, 0.8, 0.9, 1}, 
    xtick={0, 0.1, 0.2, 0.3, 0.4, 0.5, 0.6, 0.7, 0.8, 0.9, 1},
    mark size = 1.5]
    
    \pgfplotstableread{reduced_p001_l10_marginal.move-fmeasure.data}\fmeasuredata

  
%
    \addplot[forget plot,each nth point=10,filter discard warning=false,unbounded coords=discard,color=blue,mark=.,only marks,error bars/.cd,y dir=plus,y explicit] 
	table[y=PEC_SIT-avg,x=Threshold,y error expr=\thisrow{PEC_SIT-max}-\thisrow{PEC_SIT-avg}] {\fmeasuredata};
    \addplot[forget plot,each nth point=10,filter discard warning=false,unbounded coords=discard,color=blue,mark=.,only marks,error bars/.cd,y dir=minus,y explicit] 
	table[y=PEC_SIT-avg,x=Threshold,y error expr=\thisrow{PEC_SIT-avg}-\thisrow{PEC_SIT-min}] {\fmeasuredata};
    \addplot[color=blue,thin,mark=asterisk,mark repeat={10}] 
	table[y=PEC_SIT-avg,x=Threshold] {\fmeasuredata};
 	\addlegendentry{\tiny $\SIT$}
 

    \addplot[forget plot,each nth point=10,filter discard warning=false,unbounded coords=discard,color=green,mark=.,only marks,error bars/.cd,y dir=plus,y explicit] 
	table[y=CRF-avg,x=Threshold,y error expr=\thisrow{CRF-max}-\thisrow{CRF-avg}] {\fmeasuredata};
    \addplot[forget plot,each nth point=10,filter discard warning=false,unbounded coords=discard,color=green,mark=.,only marks,error bars/.cd,y dir=minus,y explicit] 
	table[y=CRF-avg,x=Threshold,y error expr=\thisrow{CRF-avg}-\thisrow{CRF-min}] {\fmeasuredata};
    \addplot[color=green,thin,mark=x,mark repeat={10}] 
	table[y=CRF-avg,x=Threshold] {\fmeasuredata};
 	\addlegendentry{\tiny $\CRF$}
	
    \end{axis}
\end{tikzpicture}
}
%
%
\subfloat[\meeting\ -- medium incomplete.\label{fig:task2_l10_fmeasure-meeting}]{ 
\ifdefined\TIKZEXTERNALIZE
  \tikzsetnextfilename{figure_task2_l10_fmeasure-meeting}
\fi
\begin{tikzpicture}[scale=0.76]
    \begin{axis}[
    legend columns=2,
    legend style={ at={(0.1,0.15)},anchor=west,draw=none},
    xlabel={\small Threshold},
    ylabel={\small $F_{1}$ score},
    ymin=0, ymax=1.0, xmin=0, xmax=1.0, 
    ytick={0, 0.1, 0.2, 0.3, 0.4, 0.5, 0.6, 0.7, 0.8, 0.9, 1}, 
    xtick={0, 0.1, 0.2, 0.3, 0.4, 0.5, 0.6, 0.7, 0.8, 0.9, 1},
    mark size = 1]
    
    \pgfplotstableread{reduced_p001_l10_marginal.meet-fmeasure.data}\fmeasuredata

  
%
    \addplot[forget plot,each nth point=10,filter discard warning=false,unbounded coords=discard,color=blue,mark=.,only marks,error bars/.cd,y dir=plus,y explicit] 
	table[y=PEC_SIT-avg,x=Threshold,y error expr=\thisrow{PEC_SIT-max}-\thisrow{PEC_SIT-avg}] {\fmeasuredata};
    \addplot[forget plot,each nth point=10,filter discard warning=false,unbounded coords=discard,color=blue,mark=.,only marks,error bars/.cd,y dir=minus,y explicit] 
	table[y=PEC_SIT-avg,x=Threshold,y error expr=\thisrow{PEC_SIT-avg}-\thisrow{PEC_SIT-min}] {\fmeasuredata};
    \addplot[color=blue,thin,mark=asterisk,mark repeat={10}] 
	table[y=PEC_SIT-avg,x=Threshold] {\fmeasuredata};
 	\addlegendentry{\tiny $\SIT$}
 

    \addplot[forget plot,each nth point=10,filter discard warning=false,unbounded coords=discard,color=green,mark=.,only marks,error bars/.cd,y dir=plus,y explicit] 
	table[y=CRF-avg,x=Threshold,y error expr=\thisrow{CRF-max}-\thisrow{CRF-avg}] {\fmeasuredata};
    \addplot[forget plot,each nth point=10,filter discard warning=false,unbounded coords=discard,color=green,mark=.,only marks,error bars/.cd,y dir=minus,y explicit] 
	table[y=CRF-avg,x=Threshold,y error expr=\thisrow{CRF-avg}-\thisrow{CRF-min}] {\fmeasuredata};
    \addplot[color=green,thin,mark=x,mark repeat={10}] 
	table[y=CRF-avg,x=Threshold] {\fmeasuredata};
 	\addlegendentry{\tiny $\CRF$}

    \end{axis}
\end{tikzpicture}
}

%
%
%
\subfloat[\moving\ -- large incomplete.\label{fig:task2_l20_fmeasure-moving}]{
\ifdefined\TIKZEXTERNALIZE
  \tikzsetnextfilename{figure_task2_l20_fmeasure-moving}
\fi
\begin{tikzpicture}[scale=0.76]
    \begin{axis}[
    legend columns=2,
    legend style={ at={(0.1,0.15)},anchor=west,draw=none},
    xlabel={\small Threshold},
    ylabel={\small $F_{1}$ score},
    ymin=0, ymax=1.0, xmin=0, xmax=1.0, 
    ytick={0, 0.1, 0.2, 0.3, 0.4, 0.5, 0.6, 0.7, 0.8, 0.9, 1}, 
    xtick={0, 0.1, 0.2, 0.3, 0.4, 0.5, 0.6, 0.7, 0.8, 0.9, 1},
    mark size = 1.5]

    \pgfplotstableread{reduced_p001_l20_marginal.move-fmeasure.data}\fmeasuredata

  
%
    \addplot[forget plot,each nth point=10,filter discard warning=false,unbounded coords=discard,color=blue,mark=.,only marks,error bars/.cd,y dir=plus,y explicit] 
	table[y=PEC_SIT-avg,x=Threshold,y error expr=\thisrow{PEC_SIT-max}-\thisrow{PEC_SIT-avg}] {\fmeasuredata};
    \addplot[forget plot,each nth point=10,filter discard warning=false,unbounded coords=discard,color=blue,mark=.,only marks,error bars/.cd,y dir=minus,y explicit] 
	table[y=PEC_SIT-avg,x=Threshold,y error expr=\thisrow{PEC_SIT-avg}-\thisrow{PEC_SIT-min}] {\fmeasuredata};
    \addplot[color=blue,thin,mark=asterisk,mark repeat={10}] 
	table[y=PEC_SIT-avg,x=Threshold] {\fmeasuredata};
 	\addlegendentry{\tiny $\SIT$}
 

    \addplot[forget plot,each nth point=10,filter discard warning=false,unbounded coords=discard,color=green,mark=.,only marks,error bars/.cd,y dir=plus,y explicit] 
	table[y=CRF-avg,x=Threshold,y error expr=\thisrow{CRF-max}-\thisrow{CRF-avg}] {\fmeasuredata};
    \addplot[forget plot,each nth point=10,filter discard warning=false,unbounded coords=discard,color=green,mark=.,only marks,error bars/.cd,y dir=minus,y explicit] 
	table[y=CRF-avg,x=Threshold,y error expr=\thisrow{CRF-avg}-\thisrow{CRF-min}] {\fmeasuredata};
    \addplot[color=green,thin,mark=x,mark repeat={10}] 
	table[y=CRF-avg,x=Threshold] {\fmeasuredata};
 	\addlegendentry{\tiny $\CRF$}
          
    \end{axis}
\end{tikzpicture}
}
%
%
\subfloat[\meeting\ -- large incomplete.\label{fig:task2_l20_fmeasure-meeting}]{ 
\ifdefined\TIKZEXTERNALIZE
  \tikzsetnextfilename{figure_task2_l20_fmeasure-meeting}
\fi
\begin{tikzpicture}[scale=0.76]
    \begin{axis}[
    legend columns=2,
    legend style={ at={(0.1,0.15)},anchor=west,draw=none},
    xlabel={\small Threshold},
    ylabel={\small $F_{1}$ score},
    ymin=0, ymax=1.0, xmin=0, xmax=1.0, 
    ytick={0, 0.1, 0.2, 0.3, 0.4, 0.5, 0.6, 0.7, 0.8, 0.9, 1}, 
    xtick={0, 0.1, 0.2, 0.3, 0.4, 0.5, 0.6, 0.7, 0.8, 0.9, 1},
    mark size = 1.5]

    \pgfplotstableread{reduced_p001_l20_marginal.meet-fmeasure.data}\fmeasuredata


 	
    \addplot[forget plot,each nth point=10,filter discard warning=false,unbounded coords=discard,color=blue,mark=.,only marks,error bars/.cd,y dir=plus,y explicit] 
	table[y=PEC_SIT-avg,x=Threshold,y error expr=\thisrow{PEC_SIT-max}-\thisrow{PEC_SIT-avg}] {\fmeasuredata};
    \addplot[forget plot,each nth point=10,filter discard warning=false,unbounded coords=discard,color=blue,mark=.,only marks,error bars/.cd,y dir=minus,y explicit] 
	table[y=PEC_SIT-avg,x=Threshold,y error expr=\thisrow{PEC_SIT-avg}-\thisrow{PEC_SIT-min}] {\fmeasuredata};
    \addplot[color=blue,thin,mark=asterisk,mark repeat={10}] 
	table[y=PEC_SIT-avg,x=Threshold] {\fmeasuredata};
 	\addlegendentry{\tiny $\SIT$}
 

    \addplot[forget plot,each nth point=10,filter discard warning=false,unbounded coords=discard,color=green,mark=.,only marks,error bars/.cd,y dir=plus,y explicit] 
	table[y=CRF-avg,x=Threshold,y error expr=\thisrow{CRF-max}-\thisrow{CRF-avg}] {\fmeasuredata};
    \addplot[forget plot,each nth point=10,filter discard warning=false,unbounded coords=discard,color=green,mark=.,only marks,error bars/.cd,y dir=minus,y explicit] 
	table[y=CRF-avg,x=Threshold,y error expr=\thisrow{CRF-avg}-\thisrow{CRF-min}] {\fmeasuredata};
    \addplot[color=green,thin,mark=x,mark repeat={10}] 
	table[y=CRF-avg,x=Threshold] {\fmeasuredata};
 	\addlegendentry{\tiny $\CRF$}
 
    \end{axis}
\end{tikzpicture}
}
\caption{Average $F_1$ scores over five runs when data are removed. Marginal inference is used.}
\label{fig:task2_fmeasure}
\end{figure}

\begin{figure}
\centering
\pgfplotsset{xbar,
enlarge y limits=0.4,
every axis legend/.append style={at={(0.5,1.0)},anchor=south,draw=none},
legend columns=3,
reverse legend,
area legend,
width=0.5\textwidth,
height=0.3\textwidth,
tick label style={font=\small},
xmin=0,xmax=1.0,
xtick={0, 0.1, 0.2, 0.3, 0.4, 0.5, 0.6, 0.7, 0.8, 0.9, 1},
ytick={1,2},
yticklabels={\small $\CRF$, \small $\SIT$},
axis x line*=bottom,
axis y line*=left}
\subfloat[\moving\label{fig:task2_fmeasure-mm-moving}]{
\ifdefined\TIKZEXTERNALIZE
  \tikzsetnextfilename{figure_task2_fmeasure-mm-moving}
\fi
\begin{tikzpicture}[scale=0.86]
\begin{axis}[xlabel={$F_{1}$ score}] 

\addplot[draw=brown,pattern color=brown!70,pattern=crosshatch,xbar,error bars/.cd,x dir=both,x explicit]
coordinates {
  (0.4112,1) +-(0.0491,0.0491)
  (0.7974,2) +-(0.0107,0.0107)
}; 
  \addlegendentry{\tiny large}

\addplot[draw=blue,pattern color=blue!50,pattern=crosshatch dots,xbar,error bars/.cd,x dir=both,x explicit]
coordinates {
  (0.4316,1) +-(0.0376,0.0376)
  (0.8078,2) +-(0.0108,0.0108)
};
  \addlegendentry{\tiny medium} 

\addplot[draw=black,pattern=north west lines]
coordinates {
  (0.7470,1)
  (0.8901,2)
};
  \addlegendentry{\tiny original} 
  
\node[right,fill=white,inner sep=1.5pt] at (axis cs:0.02,0.75) {\tiny 0.4112};
\node[right,fill=white,inner sep=1.5pt] at (axis cs:0.02,1) {\tiny 0.4316};
\node[right,fill=white,inner sep=1.5pt] at (axis cs:0.02,1.25) {\tiny 0.7470};

\node[right,fill=white,inner sep=1.5pt] at (axis cs:0.02,1.75) {\tiny 0.7974};
\node[right,fill=white,inner sep=1.5pt] at (axis cs:0.02,2) {\tiny 0.8078};
\node[right,fill=white,inner sep=1.5pt] at (axis cs:0.02,2.25) {\tiny 0.8901};

\end{axis}
\end{tikzpicture}
}
\subfloat[\meeting\label{fig:task2_fmeasure-mm-meeting}]{
\ifdefined\TIKZEXTERNALIZE
  \tikzsetnextfilename{figure_task2_fmeasure-mm-meeting}
\fi
\begin{tikzpicture}[scale=0.86]
\begin{axis}[xlabel={$F_{1}$ score}] 

\addplot[draw=brown,pattern color=brown!50,pattern=crosshatch,xbar,error bars/.cd,x dir=both,x explicit]
coordinates {
  (0.3319,1) +-(0.0507,0.0507)
  (0.7763,2) +-(0.0159,0.0159)
};
  \addlegendentry{\tiny large} 

\addplot[draw=blue,pattern color=blue!70,pattern=crosshatch dots,xbar,error bars/.cd,x dir=both,x explicit]
coordinates {
  (0.4101,1) +-(0.0295,0.0295)
  (0.8222,2) +-(0.0025,0.0025)
}; 
  \addlegendentry{\tiny medium}

  \addplot[draw=black,pattern=north west lines]
coordinates {
  (0.9004,1)
  (0.8763,2)
};
  \addlegendentry{\tiny original} 
  
\node[right,fill=white,inner sep=1.5pt] at (axis cs:0.02,0.75) {\tiny 0.3319};
\node[right,fill=white,inner sep=1.5pt] at (axis cs:0.02,1) {\tiny 0.4101};
\node[right,fill=white,inner sep=1.5pt] at (axis cs:0.02,1.25) {\tiny 0.9004};

\node[right,fill=white,inner sep=1.5pt] at (axis cs:0.02,1.75) {\tiny 0.7763};
\node[right,fill=white,inner sep=1.5pt] at (axis cs:0.02,2) {\tiny 0.8222};
\node[right,fill=white,inner sep=1.5pt] at (axis cs:0.02,2.25) {\tiny 0.8763};

\end{axis}
\end{tikzpicture}
}
\caption{Average $F_1$ scores over five runs when data are removed. MAP inference is used.}
\label{fig:task2_fmeasure-mm}
\end{figure}
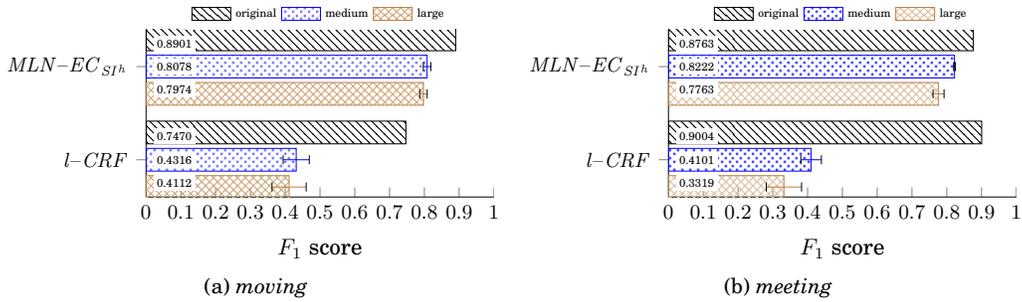

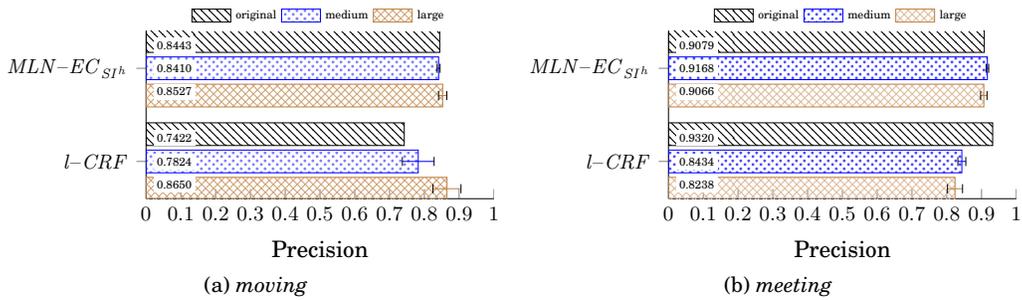
\begin{figure}
\centering
\pgfplotsset{xbar,
enlarge y limits=0.4,
every axis legend/.append style={at={(0.5,1.0)},anchor=south,draw=none},
legend columns=3,
reverse legend,
area legend,
width=0.5\textwidth,
height=0.3\textwidth,
tick label style={font=\small},
xmin=0,xmax=1.0,
xtick={0, 0.1, 0.2, 0.3, 0.4, 0.5, 0.6, 0.7, 0.8, 0.9, 1},
ytick={1,2},
yticklabels={\small $\CRF$, \small $\SIT$},
axis x line*=bottom,
axis y line*=left}
\subfloat[\moving\label{fig:task2_precision-mm-moving}]{
\ifdefined\TIKZEXTERNALIZE
  \tikzsetnextfilename{figure_task2_precision-mm-moving}
\fi
\begin{tikzpicture}[scale=0.86]
\begin{axis}[xlabel={Precision}] 

\addplot[draw=brown,pattern color=brown!70,pattern=crosshatch,xbar,error bars/.cd,x dir=both,x explicit]
coordinates {
  (0.8650,1) +-(0.0403,0.0403)
  (0.8527,2) +-(0.0118,0.0118)
}; 
  \addlegendentry{\tiny large}

\addplot[draw=blue,pattern color=blue!50,pattern=crosshatch dots,xbar,error bars/.cd,x dir=both,x explicit]
coordinates {
  (0.7824,1) +-(0.0456,0.0456)
  (0.8410,2) +-(0.0042,0.0042)
};
  \addlegendentry{\tiny medium} 
  
\addplot[draw=black,pattern=north west lines]
coordinates {
  (0.7422,1)
  (0.8443,2)
};
  \addlegendentry{\tiny original} 
  
\node[right,fill=white,inner sep=1.5pt] at (axis cs:0.02,0.75) {\tiny 0.8650};
\node[right,fill=white,inner sep=1.5pt] at (axis cs:0.02,1) {\tiny 0.7824};
\node[right,fill=white,inner sep=1.5pt] at (axis cs:0.02,1.25) {\tiny 0.7422};

\node[right,fill=white,inner sep=1.5pt] at (axis cs:0.02,1.75) {\tiny 0.8527};
\node[right,fill=white,inner sep=1.5pt] at (axis cs:0.02,2) {\tiny 0.8410};
\node[right,fill=white,inner sep=1.5pt] at (axis cs:0.02,2.25) {\tiny 0.8443};

\end{axis}
\end{tikzpicture}
}
\subfloat[\meeting\label{fig:task2_precision-mm-meeting}]{
\ifdefined\TIKZEXTERNALIZE
  \tikzsetnextfilename{figure_task2_precision-mm-meeting}
\fi
\begin{tikzpicture}[scale=0.86]
\begin{axis}[xlabel={Precision}] 

\addplot[draw=brown,pattern color=brown!50,pattern=crosshatch,xbar,error bars/.cd,x dir=both,x explicit]
coordinates {
  (0.8238,1) +-(0.0216,0.0216)
  (0.9066,2) +-(0.0092,0.0092)
};
  \addlegendentry{\tiny large} 

\addplot[draw=blue,pattern color=blue!70,pattern=crosshatch dots,xbar,error bars/.cd,x dir=both,x explicit]
coordinates {
  (0.8434,1) +-(0.0114,0.0114)
  (0.9168,2) +-(0.0037,0.0037)
}; 
  \addlegendentry{\tiny medium}
  
  \addplot[draw=black,pattern=north west lines]
coordinates {
  (0.9320,1)
  (0.9079,2)
};
  \addlegendentry{\tiny original} 

\node[right,fill=white,inner sep=1.5pt] at (axis cs:0.02,0.75) {\tiny 0.8238};
\node[right,fill=white,inner sep=1.5pt] at (axis cs:0.02,1) {\tiny 0.8434};
\node[right,fill=white,inner sep=1.5pt] at (axis cs:0.02,1.25) {\tiny 0.9320};

\node[right,fill=white,inner sep=1.5pt] at (axis cs:0.02,1.75) {\tiny 0.9066};
\node[right,fill=white,inner sep=1.5pt] at (axis cs:0.02,2) {\tiny 0.9168};
\node[right,fill=white,inner sep=1.5pt] at (axis cs:0.02,2.25) {\tiny 0.9079};

\end{axis}
\end{tikzpicture}
}
\caption{Average precision over five runs when data are removed. MAP inference is used.}
\label{fig:task2_precision-mm}
\end{figure}

\begin{figure}
\centering
\pgfplotsset{xbar,
enlarge y limits=0.4,
every axis legend/.append style={at={(0.5,1.0)},anchor=south,draw=none},
legend columns=3,
reverse legend,
area legend,
width=0.5\textwidth,
height=0.3\textwidth,
tick label style={font=\small},
xmin=0,xmax=1.0,
xtick={0, 0.1, 0.2, 0.3, 0.4, 0.5, 0.6, 0.7, 0.8, 0.9, 1},
ytick={1,2},
yticklabels={\small $\CRF$, \small $\SIT$},
axis x line*=bottom,
axis y line*=left}
\subfloat[\moving\label{fig:task2_recall-mm-moving}]{
\ifdefined\TIKZEXTERNALIZE
  \tikzsetnextfilename{figure_task2_recall-mm-moving}
\fi
\begin{tikzpicture}[scale=0.86]
\begin{axis}[xlabel={Recall}] 

\addplot[draw=brown,pattern color=brown!70,pattern=crosshatch,xbar,error bars/.cd,x dir=both,x explicit]
coordinates {
  (0.2697,1) +-(0.0421,0.0421)
  (0.7489,2) +-(0.0163,0.0163)
};
  \addlegendentry{\tiny large}

\addplot[draw=blue,pattern color=blue!50,pattern=crosshatch dots,xbar,error bars/.cd,x dir=both,x explicit]
coordinates {
  (0.2980,1) +-(0.0307,0.0307)
  (0.7771,2) +-(0.0183,0.0183)
}; 
  \addlegendentry{\tiny medium} 
  
  \addplot[draw=black,pattern=north west lines]
coordinates {
  (0.7519,1)
  (0.9410,2)
};
  \addlegendentry{\tiny original} 
  
\node[right,fill=white,inner sep=1.5pt] at (axis cs:0.02,0.75) {\tiny 0.2697};
\node[right,fill=white,inner sep=1.5pt] at (axis cs:0.02,1) {\tiny 0.2980};
\node[right,fill=white,inner sep=1.5pt] at (axis cs:0.02,1.25) {\tiny 0.7519};

\node[right,fill=white,inner sep=1.5pt] at (axis cs:0.02,1.75) {\tiny 0.7489};
\node[right,fill=white,inner sep=1.5pt] at (axis cs:0.02,2) {\tiny 0.7771};
\node[right,fill=white,inner sep=1.5pt] at (axis cs:0.02,2.25) {\tiny 0.9410};

\end{axis}
\end{tikzpicture}
}
\subfloat[\meeting\label{fig:task2_recall-mm-meeting}]{
\ifdefined\TIKZEXTERNALIZE
  \tikzsetnextfilename{figure_task2_recall-mm-meeting}
\fi
\begin{tikzpicture}[scale=0.86]
\begin{axis}[xlabel={Recall}] 

\addplot[draw=brown,pattern color=brown!50,pattern=crosshatch,xbar,error bars/.cd,x dir=both,x explicit]
coordinates {
  (0.2078,1) +-(0.0380,0.0380)
  (0.6788,2) +-(0.0214,0.0214)
};
  \addlegendentry{\tiny large} 

\addplot[draw=blue,pattern color=blue!70,pattern=crosshatch dots,xbar,error bars/.cd,x dir=both,x explicit]
coordinates {
  (0.2709,1) +-(0.0249,0.0249)
  (0.7453,2) +-(0.0020,0.0020)
}; 
  \addlegendentry{\tiny medium}
  
\addplot[draw=black,pattern=north west lines]
coordinates {
  (0.8708,1)
  (0.8468,2)
};
  \addlegendentry{\tiny original} 
  
\node[right,fill=white,inner sep=1.5pt] at (axis cs:0.02,0.75) {\tiny 0.2078};
\node[right,fill=white,inner sep=1.5pt] at (axis cs:0.02,1) {\tiny 0.2709};
\node[right,fill=white,inner sep=1.5pt] at (axis cs:0.02,1.25) {\tiny 0.8708};

\node[right,fill=white,inner sep=1.5pt] at (axis cs:0.02,1.75) {\tiny 0.6788};
\node[right,fill=white,inner sep=1.5pt] at (axis cs:0.02,2) {\tiny 0.7453};
\node[right,fill=white,inner sep=1.5pt] at (axis cs:0.02,2.25) {\tiny 0.8468};

\end{axis}
\end{tikzpicture}
}
\caption{Average recall over five runs when data are removed. MAP inference is used.}
\label{fig:task2_recall-mm}
\end{figure}
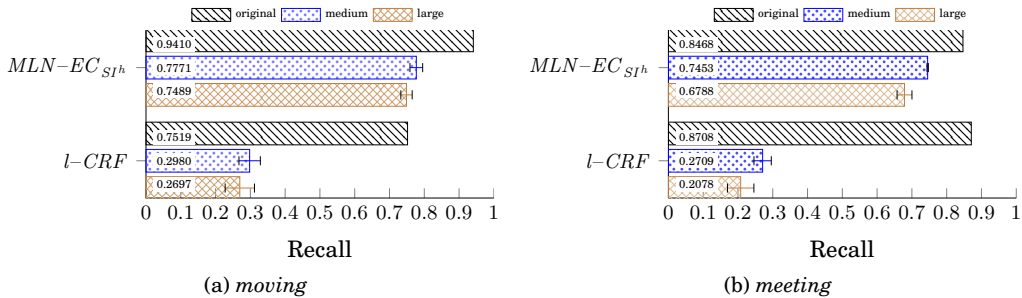

In the rest of this section we will denote \textit{medium} and \textit{large incomplete sequences} the test sequences that contain random blank intervals of $10$ and $20$ time-point duration, respectively. The evaluation measures are the same as in Task I. Figure \ref{fig:task2_fmeasure} presents the results in terms of $F_1$ score for the two methods, using marginal inference. The bar charts of Figure \ref{fig:task2_auprc}, on the other hand, present the average AUPRC of the two methods compared also to the AUPRC when no data is removed (from Tables \ref{tbl:auprc-moving} and \ref{tbl:auprc-meeting} of Task I). The threshold values range between $0.0$ and $1.0$. Using similar illustration, Figures \ref{fig:task2_fmeasure-mm}, \ref{fig:task2_precision-mm} and \ref{fig:task2_recall-mm} present the results of the two methods using MAP inference. All results are averaged over five runs and error bars display the standard deviation.

Unlike $\SIT$, \LCRF\ appears to be affected significantly  from incomplete evidence. 
Using marginal inference on the \moving\ CE in the original test sequences (see Table \ref{tbl:wlearning-moving}) \LCRF\ achieved an $F_1$ score of $0.7967$. At the same threshold value, the average $F_1$ score of \LCRF\ drops to $0.566$ and $0.53$ for medium and large incomplete sequences, respectively. $\SIT$ is affected much less, achieving $F_1$ scores of $0.836$ and $0.832$, for medium and large incomplete sequences, respectively. In terms of AUPRC (Figure \ref{fig:task2_auprc-moving}), the performance of \LCRF\ also drops by $13$ points, while $\SIT$  is almost unaffected. 
When MAP inference is used, the effect of the removal of data seems to be larger.
The recall of \LCRF\ falls by more than $45$ percentage points, causing the $F_1$ score to drop by more than $30$ percentage points (Figures \ref{fig:task2_precision-mm-moving} and \ref{fig:task2_fmeasure-mm-moving}). The number of recognised \moving\ activities is reduced, resulting in an increase in precision, but with high variance (Figure \ref{fig:task2_precision-mm-moving}). The precision of $\SIT$ remains close to the original test set, with a small variance. However, its recall drops 
and  causes the reduction of $F_1$ score by $8$ and $9$ percentage points for medium and large incomplete sequences, respectively. 

In the case of the \meeting\ CE, $\SIT$ also seems to resist more than \LCRF\ to the distortion of the data. The $F_1$ score is higher than that of \LCRF\ for many threshold values, using marginal inference (see Figures \ref{fig:task2_l10_fmeasure-meeting} and \ref{fig:task2_l20_fmeasure-meeting}). For threshold $0.5$ in the original test sequences \LCRF\ achieved an $F_1$ score that was higher by $10$ percentage points than that of $\SIT$. However, when data are removed its $F_1$ score for the same threshold drops much  lower than that of $\SIT$ (a difference of more than $15$ percentage points). The AUPRC of \LCRF\ also drops much more (more than $10$ points) than that of $\SIT$ (see Figure \ref{fig:task2_auprc-meeting}).
The effect is even higher when MAP inference is used for \meeting\ CE. In particular, the recall of \LCRF\ drops more than $60$ percentage points (see Figure \ref{fig:task2_recall-mm-meeting}), while that of $\SIT$ drops much less. Thus, the $F_1$ score of $\SIT$ reduces by less than $10$ percentage points, while that of \LCRF\ is $50$  percentage points lower than in the original data (see Figure \ref{fig:task2_fmeasure-mm-meeting}).


%
%

%
%
\begin{figure}
\centering
\pgfplotsset{xbar,
enlarge y limits=0.4,
every axis legend/.append style={at={(0.5,1.0)},anchor=south,draw=none},
legend columns=3,
reverse legend,
area legend,
width=0.5\textwidth,
height=0.3\textwidth,
tick label style={font=\small},
xmin=0,xmax=1.0,
xtick={0, 0.1, 0.2, 0.3, 0.4, 0.5, 0.6, 0.7, 0.8, 0.9, 1},
ytick={1,2},
yticklabels={\small $\CRF$, \small $\SIT$},
axis x line*=bottom,
axis y line*=left
}
\subfloat[\moving\label{fig:task2_auprc-moving}]{
\ifdefined\TIKZEXTERNALIZE
  \tikzsetnextfilename{figure_task2_auprc-moving}
\fi
\begin{tikzpicture}[scale=0.86]
\begin{axis}[xlabel={AUPRC}] 

\addplot[draw=brown,pattern color=brown!70,pattern=crosshatch,xbar,error bars/.cd,x dir=both,x explicit]
coordinates {
  (0.7080,1) +-(0.0341,0.0341)
  (0.8418,2) +-(0.0153,0.0153)
}; 
  \addlegendentry{\tiny large}

\addplot[draw=blue,pattern color=blue!50,pattern=crosshatch dots,xbar,error bars/.cd,x dir=both,x explicit]
coordinates {
  (0.7066,1) +-(0.0266,0.0266)
  (0.8548,2) +-(0.0063,0.0063)
  };
  \addlegendentry{\tiny medium} 

\addplot[draw=black,pattern=north west lines]
coordinates {
  ( 0.8358,1)
  (0.8597,2)
};
  \addlegendentry{\tiny original} 
  
\node[right,fill=white,inner sep=1.5pt] at (axis cs:0.02,0.75) {\tiny 0.7080};
\node[right,fill=white,inner sep=1.5pt] at (axis cs:0.02,1) {\tiny 0.7066};
\node[right,fill=white,inner sep=1.5pt] at (axis cs:0.02,1.25) {\tiny 0.8358};

\node[right,fill=white,inner sep=1.5pt] at (axis cs:0.02,1.75) {\tiny 0.8418};
\node[right,fill=white,inner sep=1.5pt] at (axis cs:0.02,2) {\tiny 0.8548};
\node[right,fill=white,inner sep=1.5pt] at (axis cs:0.02,2.25) {\tiny 0.8597};
  
\end{axis}
\end{tikzpicture}
}
\subfloat[\meeting\label{fig:task2_auprc-meeting}]{
\ifdefined\TIKZEXTERNALIZE
  \tikzsetnextfilename{figure_task2_auprc-meeting}
\fi
\begin{tikzpicture}[scale=0.86]
\begin{axis}[xlabel={AUPRC}] 

\addplot[draw=brown,pattern color=brown!50,pattern=crosshatch,xbar,error bars/.cd,x dir=both,x explicit]
coordinates {
  (0.7337,1) +-(0.0464,0.0464)
  (0.7119,2) +-(0.0324,0.0324)
};
  \addlegendentry{\tiny large} 

\addplot[draw=blue,pattern color=blue!70,pattern=crosshatch dots,xbar,error bars/.cd,x dir=both,x explicit]
coordinates {
  (0.7912,1) +-(0.0202,0.0202)
  (0.7195,2) +-(0.0132,0.0132)
}; 
  \addlegendentry{\tiny medium}

  \addplot[draw=black,pattern=north west lines]
coordinates {
  (0.8937,1)
  (0.7559,2)
};
  \addlegendentry{\tiny original} 
  
\node[right,fill=white,inner sep=1.5pt] at (axis cs:0.02,0.75) {\tiny 0.7337};
\node[right,fill=white,inner sep=1.5pt] at (axis cs:0.02,1) {\tiny 0.7912};
\node[right,fill=white,inner sep=1.5pt] at (axis cs:0.02,1.25) {\tiny 0.8937};

\node[right,fill=white,inner sep=1.5pt] at (axis cs:0.02,1.75) {\tiny 0.7119};
\node[right,fill=white,inner sep=1.5pt] at (axis cs:0.02,2) {\tiny 0.7195};
\node[right,fill=white,inner sep=1.5pt] at (axis cs:0.02,2.25) {\tiny 0.7559};

\end{axis}
\end{tikzpicture}
}
\caption{Average AUPRC scores over five runs when data are removed. Marginal inference is used.}
\label{fig:task2_auprc}
\end{figure}
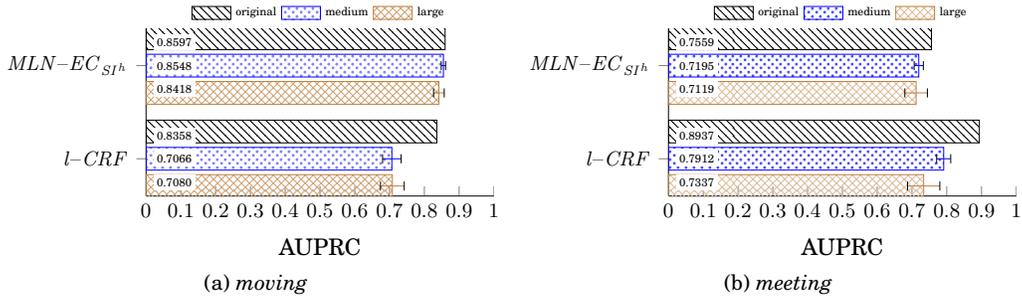


%
%

%
%

%
%

In summary, this task showed that the logic-based \MLNEC\ method is more robust than its purely statistical \LCRF\ counterpart, when data are removed from the test set, rendering it less similar to the training set. This is due to the fact that \LCRF\ is completely data driven and does not employ any background knowledge. On the other hand, \MLNEC\ employs background knowledge, including the domain independent axioms of inertia. Consequently, the persistence of a CE is modeled differently by \LCRF\ and \MLNEC. \LCRF\ learns to maintain the state of CEs under some circumstances that appear in the training data. However it does not model the inertia explicitly. Therefore, when the circumstances change its performance is hurt significantly. \MLNEC\ on the other hand enjoys the explicit modelling of inertia, provided as background knowledge. Even when this inertia is softened, it remains a strong bias in the model. As a result, \MLNEC\ avoids overfitting the training data and behaves robustly when the data changes.  

\section{Related work} \label{sec:related_work}

%
%
Event Calculus is related to other formalisms in the literature of commonsense reasoning, such as the Situation Calculus \cite{mccarthy1968some,reiter2001knowledge}, the Fluent Calculus \cite{thielscher1999situation,thielscher2001qualification}, the action language $\mathcal{C}+$ \cite{giunchiglia2004nonmonotonic,akman2004representing} and Temporal Action Logics \cite{doherty1998tal,Kv05talplanner}. Action formalisms provide domain-independent axioms in order to represent and reason about the effects of events and support the property of inertia. Comparisons and proofs of equivalence between formalisms for commonsense reasoning can be found in
\citeN{kowalski1997reconciling}, \citeN{van1997relation}, \citeN{chittaro2000temporal}, \citeN{miller2002some}, \citeN{schiffel2006reconciling}, \citeN[Chapter~15]{mueller2006commonsense}, \citeN{craven2006execution} and \citeN{paschke2009rule}. 

Probabilistic extensions of the Situation Calculus have been proposed in the literature, in order to support noisy input from sensors, stochastic events and model Markov Decision Processes, e.g.~see \citeN{bacchus1995reasoning}, \citeN{pinto2000non}, \citeN{mateus2001probabilistic}, \citeN{HajishirziA08} and \citeN[Chapter~12]{reiter2001knowledge}. Furthermore, \citeN{holldobler2006flucap} proposed a probabilistic extension of the Fluent Calculus. Both Situation Calculus and Fluent Calculus, as well as in their probabilistic variants use a tree model of time, in which each event may give rise to a different possible future. A point in time is represented by a situation, which is a possible sequence of events. As a result, events are represented to occur sequentially and atemporally. 
 
In the Event Calculus, as well as in $\mathcal{C}+$ and Temporal Action Logics, there is a single time line on which events occur. This is a more suitable model for event recognition, where the task is to recognise CEs of interest in a time-stamped sequence of observed SDEs. 

$\mathcal{PC}+$ is a probabilistic generalisation of $\mathcal{C}+$ that incorporates   probabilistic knowledge about the effects of events \cite{eiter2002probabilistic}. $\mathcal{PC}+$ supports non-deterministic and probabilistic effects of events, as well as probabilistic uncertainty about the initial state of the application. Similar to the aforementioned probabilistic variants of commonsense reasoning languages, the method focuses on planning under uncertainty while inertia remains deterministic. 

In \MLNEC\ we can have customisable inertia behaviour by adjusting the weights of the inertia axioms, as shown in Sections \ref{sec:inertia} and \ref{sec:experiments}. To deal with uncertainty we employ the framework of Markov Logic Networks for probabilistic modelling and automatically estimate the weights from training data.

A related approach that we have developed in parallel is that of \citeN{anskarl2013PLPEC}. The method employs an Event Calculus formalism that is based on probabilistic logic programming and handles noise in the input data. Input SDEs are assumed to be independent and are associated with detection probabilities. The Event Calculus axioms and CE definitions in the knowledge base remain hard-constrained. Given a narrative of SDEs, a CE may be recognised with some probability. Any initiation or termination caused by the given SDEs increases or decreases the probability of a CE to hold.
Inertia is modelled by the closed-world semantics of logic programming and is restricted to be deterministic. It is worth-noting that the MLN approach presented in this paper does not make any independence assumption about the input SDEs.

%
%
\citeN{ShetNRD07} proposed an activity recognition method that is based on logic programming and handles uncertainty using the Bilattice framework \cite{ginsberg1988multivalued}. The knowledge base consists of domain-specific rules, expressing CEs in terms of SDEs. Each CE or SDE is associated with two uncertainty values, indicating a degree of information and confidence respectively. The underlying idea of the method is that the more confident information is provided, the stronger the belief about the corresponding CE becomes. Another logic-based method that recognises user activities over noisy or incomplete data is proposed by \citeN{filippaki2011}. The method recognises CEs from SDEs using rules that impose temporal and spatial constraints between SDEs. Some of the constraints in CE definitions are optional. As a result, a CE can be recognised from incomplete information, but with lower confidence. The confidence of a CE increases when more of the optional SDEs are recognised. Due to noisy or incomplete information, the recognised CEs may be logically inconsistent with each other. The method resolves those inconsistencies using the confidence, duration and number of involved SDEs. In contrast to these methods, our work employs MLNs that have formal probabilistic semantics, as well as an Event Calculus formalism to represent complex CEs.

%
%
Probabilistic graphical models have been successfully applied to a variety of event recognition tasks where a significant amount of uncertainty exists. Since event recognition requires the processing of streams of time-stamped SDE, numerous event recognition methods are based on sequential variants of probabilistic graphical models, such as Hidden Markov Models (HMM) \cite{rabiner1986HMM}, Dynamic Bayesian Networks (DBN) \cite{murphy2002DBN} and linear-chain Conditional Random Fields (CRF) \cite{LaffertyMP01CRF}. Such models can naturally handle uncertainty but their propositional structure provides limited representation capabilities. To overcome this limitation, graphical models have been extended to model interactions between multiple entities \cite{BrandOP97,gong2003,wu2007joint,vail2007conditional}, to capture long-term dependencies between states \cite{hongeng2003large} and to model the hierarchical composition of events \cite{natarajan2007hierarchical,liao2007hierarchical}. However, the lack of a formal representation language makes the definition of structured CEs complicated and the use of background knowledge very hard.

%
%
Recently, statistical relational learning (SRL) methods have been applied to event recognition. These methods combine logic with probabilistic models, in order to represent complex relational structures and perform reasoning under uncertainty. Using a declarative language as a template, SRL methods specify probabilistic models at an abstract level. Given an input stream of SDE observations, the template is partially or completely instantiated, creating lifted or propositional graphical models on which probabilistic inference is performed \cite{BrazAR08,RaedtK10}.

%
%
Among others, HMMs have been extended in order to represent states and transitions using logical expressions \cite{kersting2006lohmm,natarajan2008LoHiHMM}. In contrast to standard HMM, the logical representation allows the model to represent compactly probability distributions over sequences of logical atoms, rather than propositional symbols. Similarly, DBNs have been extended using first-order logic \cite{manfredotti2009RBN,manfredotti2010RBN}. A tree structure is used, where each node corresponds to a first-order logic expression, e.g.~a predicate representing a CE, and can be related to nodes of the same or previous time instances. Compared to their propositional counterparts, the extended HMM and DBN methods can compactly represent CE that involve various entities. 

Our method is based on Markov Logic Networks (MLNs), which is a more general and expressive model. The knowledge base of weighted first-order logic formulas in MLNs defines an arbitrarily structured undirected graphical model. Therefore, MLNs provide a generic SRL framework, which subsumes various graphical models, e.g.~HMM, CRF, etc., and can be used with expressive logic-based formalisms, such as the Event Calculus. The inertia axioms of our method allow the model to capture long-term dependencies between events. Additionally, adopting a discriminative model, the method avoids common independence assumptions over the input SDEs.

%
%
Markov Logic Networks have been used for event recognition in the literature. \citeN{BiswasTF07} combine the information provided by different low-level classifiers with the use of MLNs, in order to recognise CEs. \citeN{TranD08};~\citeN{KembhaviYD10} take into account the confidence value of the input SDEs, which may be due to noisy sensors. A more expressive approach that can represent persistent and concurrent CEs, as well as their starting and ending points, is proposed by \citeN{HelaouiNS11}. However, that method has a quadratic complexity to the number of time-points. 

%
%
%
\citeN{morariu11} proposed an MLN-based method that uses interval relations. The method determines the most consistent sequence of CEs, based on the observations of low-level classifiers. Similar to \citeN{TranD08};~\citeN{KembhaviYD10} the method expresses CEs in first-order logic, but it employs temporal relations from the  Interval Algebra \cite{Allen83}. In order to avoid the combinatorial explosion of possible intervals, as well as to eliminate the existential quantifiers in CE definitions, a bottom-up process eliminates the unlikely CE hypotheses. The elimination process can only be applied to domain-dependent axioms, as it is guided by the observations and the Interval Algebra relations. A different approach to interval-based activity recognition, is the Probabilistic Event Logic (PEL) \cite{brendel2011PEL,selman2011PEL}. Similar to MLNs, the method defines a log-linear model from a set of weighted formulas, but the formulas are represented in Event Logic \cite{siskind2001EL}. Each formula defines a soft constraint over some events, using interval relations that are represented by the \textit{spanning intervals} data structure. The method performs inference via a local-search algorithm (based on  MaxWalkSAT of \citeN{kautz1997MaxWalkSAT}), but using the spanning intervals it avoids  grounding all possible time intervals. In our work, we address the combinatorial explosion problem in a more generic manner, through the efficient representation of the domain-independent axioms. Additionally, we use a transformation procedure to further simplify the structure of the Markov network. The transformation is performed at the level of the knowledge base and is independent of the input SDEs.

\citeN{sadilek2012} employ hybrid-MLNs \cite{wang2008hybrid} in order to recognise successful and failed interactions between humans, using noisy location data from GPS devices. The method uses hybrid formulas that denoise the location data. Hybrid formulas are defined as normal soft-constrained formulas, but their weights are also associated with a real-valued function, e.g.~the distance of two persons. As a result, the strength of the constraint that a hybrid rule imposes is defined by both its weight and function --- e.g.~the closer the distance, the stronger the constraint. The weights are estimated from training data. However, the method does not employ any generic formalism for representing the events and their effects and thus it uses only domain-dependent CE definitions. On the other hand, the use of a hybrid approach for numeric constraints is an interesting alternative to the discretisation adopted by our method.

\section{Conclusions} \label{sec:conclusions}

We addressed the issue of imperfect CE definitions that stems from the uncertainty that naturally exists in event recognition. We proposed a probabilistic version of the Event Calculus based on Markov Logic Networks (\MLNEC). The method has declarative and formal (probabilistic) semantics, inheriting the properties of the Event Calculus. We placed particular emphasis on the efficiency and effectiveness of our approach. By simplifying the axioms of the Event Calculus, as well as following a knowledge transformation procedure, the method produces compact Markov networks with reduced complexity. Consequently, the performance of probabilistic inference is improved, as it takes place on a simpler model. \MLNEC\ supports flexible CE persistence, ranging from deterministic to probabilistic, in order to meet the requirements of different applications.  Due to the use of MLNs, the method lends itself naturally to learning the weights of event definitions from data, as the manual setting of weights is sub-optimal and cumbersome. \MLNEC\ is trained discriminatively, using a supervised learning technique. In the experimental evaluation, \MLNEC\ outperforms its crisp equivalent on a benchmark data. \MLNEC\ matches the performance of a linear-chain Conditional Random Fields method. Furthermore, due to the use of the Event Calculus, \MLNEC\ is affected less by missing data in the test sequences than its probabilistic akin.

There are several directions in which we would like to extend our work. In many applications the input SDE observations are accompanied by a degree of confidence, usually in the form of probability. Therefore, we consider extending our method in order to exploit data that involves such confidence values, either in the form of additional clauses (e.g.~\citeN{TranD08}, \citeN{morariu11}), or by employing different inference algorithms (e.g.~\citeN{jain2010soft}). Furthermore, we would like to address the problems that involve numerical constraints by adopting a hybrid-MLN (e.g.~\citeN{sadilek2012}) or a similar approach. We also consider extending our formalism in order to support temporal interval relations, using preprocessing techniques (e.g.~\citeN{morariu11}), or by employing different representation and inference methods (e.g.~\citeN{brendel2011PEL}, \citeN{selman2011PEL}). As shown in Section \ref{sec:experiments}, the \MLNEC\ with soft-constrained inertia performs well. We would like to extend our method to automatically soften the right subset of inertia axioms. Finally, we would like to examine structure learning/refinement methods for the CE definitions, since they are often hard to acquire from experts.

\begin{acks}
The authors wish to thank the anonymous reviewers of the 5th International RuleML Symposium on Rules, as well as Professor Larry S.~Davis and Dr.~Vlad I.~Morariu, for their valuable comments and suggestions.
\end{acks}


\begin{thebibliography}{00}


\ifx \showCODEN    \undefined \def \showCODEN     #1{\unskip}     \fi
\ifx \showDOI      \undefined \def \showDOI       #1{{\tt DOI:}\penalty0{#1}\ }
  \fi
\ifx \showISBNx    \undefined \def \showISBNx     #1{\unskip}     \fi
\ifx \showISBNxiii \undefined \def \showISBNxiii  #1{\unskip}     \fi
\ifx \showISSN     \undefined \def \showISSN      #1{\unskip}     \fi
\ifx \showLCCN     \undefined \def \showLCCN      #1{\unskip}     \fi
\ifx \shownote     \undefined \def \shownote      #1{#1}          \fi
\ifx \showarticletitle \undefined \def \showarticletitle #1{#1}   \fi
\ifx \showURL      \undefined \def \showURL       #1{#1}          \fi

\bibitem[\protect\citeauthoryear{Akman, Selim T.~Erdo{\u{g}}an, Lifschitz, and
  Turner}{Akman et~al\mbox{.}}{2004}]%
        {akman2004representing}
{Varol Akman}, {Joohyung~Lee Selim T.~Erdo{\u{g}}an}, {Vladimir Lifschitz},
  {and} {Hudson Turner}. 2004.
\newblock \showarticletitle{{Representing the Zoo World and the Traffic World
  in the language of the Causal Calculator}}.
\newblock {\em {Artificial Intelligence}\/} {153}, 1 (2004), 105--140.
\newblock


\bibitem[\protect\citeauthoryear{Allen}{Allen}{1983}]%
        {Allen83}
{James~F. Allen}. 1983.
\newblock \showarticletitle{{Maintaining Knowledge about Temporal Intervals}}.
\newblock {\em {Communications of the ACM}\/} {26}, 11 (1983), 832--843.
\newblock


\bibitem[\protect\citeauthoryear{Artikis, Sergot, and Paliouras}{Artikis
  et~al\mbox{.}}{2010a}]%
        {artikis2010logic}
{Alexander Artikis}, {Marek Sergot}, {and} {Georgios Paliouras}. 2010a.
\newblock \showarticletitle{{A Logic Programming Approach to Activity
  Recognition}}. In {\em Proceedings of the 2nd International Workshop on
  Events in Multimedia (EiMM)}. ACM, 3--8.
\newblock


\bibitem[\protect\citeauthoryear{Artikis, Skarlatidis, and Paliouras}{Artikis
  et~al\mbox{.}}{2010b}]%
        {artikis2009behaviour}
{Alexander Artikis}, {Anastasios Skarlatidis}, {and} {Georgios Paliouras}.
  2010b.
\newblock \showarticletitle{{Behaviour Recognition from Video Content: a Logic
  Programming Approach}}.
\newblock {\em {International Journal on Artificial Intelligence Tools
  (JAIT)}\/} {19}, 2 (2010), 193--209.
\newblock


\bibitem[\protect\citeauthoryear{Artikis, Skarlatidis, Portet, and
  Paliouras}{Artikis et~al\mbox{.}}{2012}]%
        {LBER_KER12}
{Alexander Artikis}, {Anastasios Skarlatidis}, {Fran\c{c}ois Portet}, {and}
  {Georgios Paliouras}. 2012.
\newblock \showarticletitle{{Logic-based event recognition}}.
\newblock {\em {Knowledge Engineering Review}\/} {27}, 4 (2012), 469--506.
\newblock


\bibitem[\protect\citeauthoryear{Bacchus, Halpern, and Levesque}{Bacchus
  et~al\mbox{.}}{1995}]%
        {bacchus1995reasoning}
{Fahiem Bacchus}, {Joseph~Y. Halpern}, {and} {Hector~J. Levesque}. 1995.
\newblock \showarticletitle{{Reasoning about Noisy Sensors in the Situation
  Calculus}}. In {\em Proceedings of the 14th International Joint Conference on
  Artificial Intelligence (IJCAI)}. Morgan Kaufmann, 1933--1940.
\newblock


\bibitem[\protect\citeauthoryear{Biba, Xhafa, Esposito, and Ferilli}{Biba
  et~al\mbox{.}}{2011}]%
        {biba2011engineering}
{Marenglen Biba}, {Fatos Xhafa}, {Floriana Esposito}, {and} {Stefano Ferilli}.
  2011.
\newblock \showarticletitle{{Engineering SLS Algorithms for Statistical
  Relational Models}}. In {\em Proceedings of the International Conference on
  Complex, Intelligent and Software Intensive Systems (CISIS)}. IEEE Computer
  Society, 502--507.
\newblock


\bibitem[\protect\citeauthoryear{Biswas, Thrun, and Fujimura}{Biswas
  et~al\mbox{.}}{2007}]%
        {BiswasTF07}
{Rahul Biswas}, {Sebastian Thrun}, {and} {Kikuo Fujimura}. 2007.
\newblock \showarticletitle{{Recognizing Activities with Multiple Cues}}. In
  {\em Proceedings of the 2nd Workshop on Human Motion - Understanding,
  Modeling, Capture and Animation} {\em (Lecture Notes in Computer Science)}.
  Springer, 255--270.
\newblock


\bibitem[\protect\citeauthoryear{Blockeel}{Blockeel}{2011}]%
        {blockeel2011statistical}
{Hendrik Blockeel}. 2011.
\newblock \showarticletitle{{Statistical Relational Learning}}.
\newblock In {\em Handbook on Neural Information Processing}, {Monica
  Bianchini}, {Marco Maggini}, {and} {Lakhmi Jain} (Eds.). Springer.
\newblock


\bibitem[\protect\citeauthoryear{Brand, Oliver, and Pentland}{Brand
  et~al\mbox{.}}{1997}]%
        {BrandOP97}
{Matthew Brand}, {Nuria Oliver}, {and} {Alex Pentland}. 1997.
\newblock \showarticletitle{{Coupled hidden Markov models for complex action
  recognition}}. In {\em Proceedings of the Conference on Computer Vision and
  Pattern Recognition (CVPR)}. IEEE Computer Society, 994--999.
\newblock


\bibitem[\protect\citeauthoryear{Brendel, Fern, and Todorovic}{Brendel
  et~al\mbox{.}}{2011}]%
        {brendel2011PEL}
{William Brendel}, {Alan Fern}, {and} {Sinisa Todorovic}. 2011.
\newblock \showarticletitle{{Probabilistic Event Logic for Interval-based Event
  Recognition}}. In {\em Proceedings of the Conference on Computer Vision and
  Pattern Recognition (CVPR)}. IEEE Computer Society, 3329--3336.
\newblock


\bibitem[\protect\citeauthoryear{Byrd, Nocedal, and Schnabel}{Byrd
  et~al\mbox{.}}{1994}]%
        {byrd1994representations}
{Richard~H Byrd}, {Jorge Nocedal}, {and} {Robert~B Schnabel}. 1994.
\newblock \showarticletitle{{Representations of quasi-Newton matrices and their
  use in limited memory methods}}.
\newblock {\em {Mathematical Programming}\/} {63}, 1 (1994), 129--156.
\newblock


\bibitem[\protect\citeauthoryear{Chittaro and Montanari}{Chittaro and
  Montanari}{2000}]%
        {chittaro2000temporal}
{Luca Chittaro} {and} {Angelo Montanari}. 2000.
\newblock \showarticletitle{{Temporal Representation and Reasoning in
  Artificial Intelligence: Issues and Approaches}}.
\newblock {\em {Annals of Mathematics and Artificial Intelligence}\/} {28}, 1
  (2000), 47--106.
\newblock


\bibitem[\protect\citeauthoryear{Collins}{Collins}{2002}]%
        {collins2002discriminative}
{Michael Collins}. 2002.
\newblock \showarticletitle{{Discriminative training methods for Hidden Markov
  Models: Theory and Experiments with Perceptron Algorithms}}. In {\em
  Proceedings of the ACL Conference on Empirical Methods in Natural Language
  Processing (EMNLP)}, Vol.~10. Association for Computational Linguistics,
  1--8.
\newblock


\bibitem[\protect\citeauthoryear{Craven}{Craven}{2006}]%
        {craven2006execution}
{Robert Craven}. 2006.
\newblock {\em {Execution mechanisms for the action language C+}}.
\newblock Ph.D. Dissertation. University of London.
\newblock


\bibitem[\protect\citeauthoryear{Culotta and McCallum}{Culotta and
  McCallum}{2004}]%
        {culotta2004confidence}
{Aron Culotta} {and} {Andrew McCallum}. 2004.
\newblock \showarticletitle{{Confidence estimation for information
  extraction}}. In {\em Proceedings of HLT-NAACL 2004: Short Papers}.
  Association for Computational Linguistics, 109--112.
\newblock


\bibitem[\protect\citeauthoryear{de~Salvo~Braz, Amir, and Roth}{de~Salvo~Braz
  et~al\mbox{.}}{2008}]%
        {BrazAR08}
{Rodrigo de Salvo~Braz}, {Eyal Amir}, {and} {Dan Roth}. 2008.
\newblock \showarticletitle{{A Survey of First-Order Probabilistic Models}}.
\newblock In {\em Innovations in Bayesian Networks: Theory and Applications}.
  Studies in Computational Intelligence, Vol. 156. Springer, 289--317.
\newblock


\bibitem[\protect\citeauthoryear{Doherty, Gustafsson, Karlsson, and
  Kvarnstr{\"o}m}{Doherty et~al\mbox{.}}{1998}]%
        {doherty1998tal}
{Patrick Doherty}, {Joakim Gustafsson}, {Lars Karlsson}, {and} {Jonas
  Kvarnstr{\"o}m}. 1998.
\newblock \showarticletitle{{TAL: Temporal Action Logics Language Specification
  and Tutorial}}.
\newblock {\em {Electronic Transactions on Artificial Intelligence}\/} {2},
  3--4 (1998), 273--306.
\newblock


\bibitem[\protect\citeauthoryear{Doherty, Lukaszewicz, and Szalas}{Doherty
  et~al\mbox{.}}{1997}]%
        {doherty1997computing}
{Patrick Doherty}, {Witold Lukaszewicz}, {and} {Andrzej Szalas}. 1997.
\newblock \showarticletitle{{Computing Circumscription Revisited: A Reduction
  Algorithm}}.
\newblock {\em {Journal of Automated Reasoning}\/} {18}, 3 (1997), 297--336.
\newblock


\bibitem[\protect\citeauthoryear{Domingos and Lowd}{Domingos and Lowd}{2009}]%
        {Domingos2009markov}
{Pedro Domingos} {and} {Daniel Lowd}. 2009.
\newblock {\em {Markov Logic: An Interface Layer for Artificial Intelligence}}.
\newblock Morgan {\&} Claypool Publishers.
\newblock


\bibitem[\protect\citeauthoryear{Eiter and Lukasiewicz}{Eiter and
  Lukasiewicz}{2003}]%
        {eiter2002probabilistic}
{Thomas Eiter} {and} {Thomas Lukasiewicz}. 2003.
\newblock \showarticletitle{{Probabilistic Reasoning about Actions in
  Nonmonotonic Causal Theories}}. In {\em UAI}, {Christopher Meek} {and} {Uffe
  Kj{\ae}rulff} (Eds.). Morgan Kaufmann, 192--199.
\newblock
\showISBNx{0-127-05664-5}


\bibitem[\protect\citeauthoryear{Etzion and Niblett}{Etzion and
  Niblett}{2010}]%
        {etzion2010event}
{Opher Etzion} {and} {Peter Niblett}. 2010.
\newblock {\em {Event Processing in Action}}.
\newblock Manning Publications Company. I--XXIV, 1--360 pages.
\newblock
\showISBNx{978-1-935182-21-4}


\bibitem[\protect\citeauthoryear{Filippaki, Antoniou, and
  Tsamardinos}{Filippaki et~al\mbox{.}}{2011}]%
        {filippaki2011}
{Chrysi Filippaki}, {Grigoris Antoniou}, {and} {Ioannis Tsamardinos}. 2011.
\newblock \showarticletitle{{Using Constraint Optimization for Conflict
  Resolution and Detail Control in Activity Recognition}}. In {\em Proceedings
  of the 2nd International Joint Conference on Ambient Intelligence (AmI)} {\em
  (Lecture Notes in Computer Science)}, Vol. 7040. Springer, 51--60.
\newblock


\bibitem[\protect\citeauthoryear{Gal, Wasserkrug, and Etzion}{Gal
  et~al\mbox{.}}{2011}]%
        {gal2011event}
{Avigdor Gal}, {Segev Wasserkrug}, {and} {Opher Etzion}. 2011.
\newblock \showarticletitle{{Event Processing over Uncertain Data}}.
\newblock In {\em Reasoning in Event-Based Distributed Systems}, {Sven Helmer},
  {Alexandra Poulovassilis}, {and} {Fatos Xhafa} (Eds.). Studies in
  Computational Intelligence, Vol. 347. Springer, 279--304.
\newblock
\showISBNx{978-3-642-19723-9}


\bibitem[\protect\citeauthoryear{Ginsberg}{Ginsberg}{1988}]%
        {ginsberg1988multivalued}
{Matthew~L. Ginsberg}. 1988.
\newblock \showarticletitle{{Multivalued logics: a uniform approach to
  reasoning in artificial intelligence}}.
\newblock {\em {Computational Intelligence}\/}  {4} (1988), 265--316.
\newblock


\bibitem[\protect\citeauthoryear{Giunchiglia, Lee, Lifschitz, McCain, and
  Turner}{Giunchiglia et~al\mbox{.}}{2004}]%
        {giunchiglia2004nonmonotonic}
{Enrico Giunchiglia}, {Joohyung Lee}, {Vladimir Lifschitz}, {Norman McCain},
  {and} {Hudson Turner}. 2004.
\newblock \showarticletitle{{Nonmonotonic Causal Theories}}.
\newblock {\em {Artificial Intelligence}\/} {153}, 1 (2004), 49--104.
\newblock


\bibitem[\protect\citeauthoryear{Gong and Xiang}{Gong and Xiang}{2003}]%
        {gong2003}
{Shaogang Gong} {and} {Tao Xiang}. 2003.
\newblock \showarticletitle{{Recognition of Group Activities using Dynamic
  Probabilistic Networks}}. In {\em Proceedings of the 9th International
  Conference on Computer Vision (ICCV)}, Vol.~2. IEEE Computer Society,
  742--749.
\newblock


\bibitem[\protect\citeauthoryear{Hajishirzi and Amir}{Hajishirzi and
  Amir}{2008}]%
        {HajishirziA08}
{Hannaneh Hajishirzi} {and} {Eyal Amir}. 2008.
\newblock \showarticletitle{{Sampling First Order Logical Particles}}. In {\em
  Proceedings of the 24th Conference in Uncertainty in Artificial Intelligence
  (UAI), Helsinki, Finland}. AUAI Press, 248--255.
\newblock


\bibitem[\protect\citeauthoryear{Helaoui, Niepert, and Stuckenschmidt}{Helaoui
  et~al\mbox{.}}{2011}]%
        {HelaouiNS11}
{Rim Helaoui}, {Mathias Niepert}, {and} {Heiner Stuckenschmidt}. 2011.
\newblock \showarticletitle{{Recognizing Interleaved and Concurrent Activities:
  A Statistical-Relational Approach}}. In {\em Proceedings of the 9th Annual
  International Conference on Pervasive Computing and Communications (PerCom)}.
  IEEE Computer Society, 1--9.
\newblock


\bibitem[\protect\citeauthoryear{H\"{o}lldobler, Karabaev, and
  Skvortsova}{H\"{o}lldobler et~al\mbox{.}}{2006}]%
        {holldobler2006flucap}
{Steffen H\"{o}lldobler}, {Eldar Karabaev}, {and} {Olga Skvortsova}. 2006.
\newblock \showarticletitle{{FLUCAP: a heuristic search planner for first-order
  MDPs}}.
\newblock {\em {Journal of Artificial Intelligence Research (JAIR)}\/} {27}, 1
  (Dec. 2006), 419--439.
\newblock
\showISSN{1076-9757}


\bibitem[\protect\citeauthoryear{Hongeng and Nevatia}{Hongeng and
  Nevatia}{2003}]%
        {hongeng2003large}
{Somboon Hongeng} {and} {Ramakant Nevatia}. 2003.
\newblock \showarticletitle{{Large-Scale Event Detection Using Semi-Hidden
  Markov Models}}. In {\em Proceedings of the 9th International Conference on
  Computer Vision (ICCV)}, Vol.~2. IEEE Computer Society, 1455--1462.
\newblock


\bibitem[\protect\citeauthoryear{Huynh and Mooney}{Huynh and Mooney}{2009}]%
        {huynh2009max}
{Tuyen~N. Huynh} {and} {Raymond~J. Mooney}. 2009.
\newblock \showarticletitle{{Max-Margin Weight Learning for Markov Logic
  Networks}}. In {\em Proceedings of the European Conference on Machine
  Learning and Principles and Practice of Knowledge Discovery in Databases
  (ECML PKDD)} {\em (Lecture Notes in Computer Science)}, Vol. 5781. Springer,
  564--579.
\newblock


\bibitem[\protect\citeauthoryear{Huynh and Mooney}{Huynh and Mooney}{2011}]%
        {huynh2011online}
{Tuyen~N. Huynh} {and} {Raymond~J. Mooney}. 2011.
\newblock \showarticletitle{{Online Max-Margin Weight Learning for Markov Logic
  Networks}}. In {\em Proceedings of the 11th SIAM International Conference on
  Data Mining (SDM11)}. Mesa, Arizona, USA, 642--651.
\newblock


\bibitem[\protect\citeauthoryear{Jain and Beetz}{Jain and Beetz}{2010}]%
        {jain2010soft}
{Dominik Jain} {and} {Michael Beetz}. 2010.
\newblock \showarticletitle{{Soft Evidential Update via Markov Chain Monte
  Carlo Inference}}. In {\em Proceedings of the 33rd Annual German Conference
  on AI (KI)} {\em (Lecture Notes in Computer Science)}, Vol. 6359. Springer,
  280--290.
\newblock


\bibitem[\protect\citeauthoryear{Kautz, Selman, and Jiang}{Kautz
  et~al\mbox{.}}{1997}]%
        {kautz1997MaxWalkSAT}
{Henry Kautz}, {Bart Selman}, {and} {Yueyen Jiang}. 1997.
\newblock \showarticletitle{{A General Stochastic Approach to Solving Problems
  with Hard and Soft Constraints}}.
\newblock In {\em The Satisfiability Problem: Theory and Applications},
  {Dingzhu Gu}, {Jun Du}, {and} {Panos Pardalos} (Eds.). DIMACS Series in
  Discrete Mathematics and Theoretical Computer Science, Vol.~35. AMS,
  573--586.
\newblock


\bibitem[\protect\citeauthoryear{Kembhavi, Yeh, and Davis}{Kembhavi
  et~al\mbox{.}}{2010}]%
        {KembhaviYD10}
{Aniruddha Kembhavi}, {Tom Yeh}, {and} {Larry~S. Davis}. 2010.
\newblock \showarticletitle{{Why Did the Person Cross the Road (There)? Scene
  Understanding Using Probabilistic Logic Models and Common Sense Reasoning}}.
  In {\em Proceedings of the 11th European Conference on Computer Vision
  (ECCV)} {\em (Lecture Notes in Computer Science)}, Vol. 6312. Springer,
  693--706.
\newblock


\bibitem[\protect\citeauthoryear{Kersting, Ahmadi, and Natarajan}{Kersting
  et~al\mbox{.}}{2009}]%
        {kersting2009counting}
{Kristian Kersting}, {Babak Ahmadi}, {and} {Sriraam Natarajan}. 2009.
\newblock \showarticletitle{{Counting Belief Propagation}}. In {\em Proceedings
  of the 25th Conference on Uncertainty in Artificial Intelligence (UAI)}. AUAI
  Press, 277--284.
\newblock


\bibitem[\protect\citeauthoryear{Kersting, Raedt, and Raiko}{Kersting
  et~al\mbox{.}}{2006}]%
        {kersting2006lohmm}
{Kristian Kersting}, {Luc~De Raedt}, {and} {Tapani Raiko}. 2006.
\newblock \showarticletitle{{Logical Hidden Markov Models}}.
\newblock {\em {Journal of Artificial Intelligence Research (JAIR)}\/} {25}, 1
  (2006), 425--456.
\newblock


\bibitem[\protect\citeauthoryear{Kok, Singla, Richardson, Domingos, Sumner,
  Poon, and Lowd}{Kok et~al\mbox{.}}{2005}]%
        {alchemy05}
{Stanley Kok}, {Parag Singla}, {Matthew Richardson}, {Pedro Domingos}, {Marc
  Sumner}, {Hoifung Poon}, {and} {Daniel Lowd}. 2005.
\newblock {\em {The Alchemy system for statistical relational AI}}.
\newblock {T}echnical {R}eport. Department of Computer Science and Engineering,
  University of Washington, Seattle, WA.
\newblock
\newblock
\shownote{\linebreak \url{ http://alchemy.cs.washington.edu}.}


\bibitem[\protect\citeauthoryear{Kowalski and Sadri}{Kowalski and
  Sadri}{1997}]%
        {kowalski1997reconciling}
{Robert Kowalski} {and} {Fariba Sadri}. 1997.
\newblock \showarticletitle{{Reconciling the Event Calculus with the Situation
  Calculus}}.
\newblock {\em {The Journal of Logic Programming}\/} {31}, 1 (1997), 39--58.
\newblock


\bibitem[\protect\citeauthoryear{Kowalski and Sergot}{Kowalski and
  Sergot}{1986}]%
        {kowalski1986logic}
{Robert Kowalski} {and} {Marek Sergot}. 1986.
\newblock \showarticletitle{{A Logic-based Calculus of Events}}.
\newblock {\em {New Generation Computing}\/} {4}, 1 (1986), 67--95.
\newblock


\bibitem[\protect\citeauthoryear{Kvarnstr{\"o}m}{Kvarnstr{\"o}m}{2005}]%
        {Kv05talplanner}
{Jonas Kvarnstr{\"o}m}. 2005.
\newblock {\em {TALplanner and Other Extensions to Temporal Action Logic}}.
\newblock Ph.D. Dissertation. Link{\"o}ping.
\newblock


\bibitem[\protect\citeauthoryear{Lafferty, McCallum, and Pereira}{Lafferty
  et~al\mbox{.}}{2001}]%
        {LaffertyMP01CRF}
{John~D. Lafferty}, {Andrew McCallum}, {and} {Fernando C.~N. Pereira}. 2001.
\newblock \showarticletitle{{Conditional Random Fields: Probabilistic Models
  for Segmenting and Labeling Sequence Data}}. In {\em Proceedings of the 18th
  International Conference on Machine Learning (ICML)}. Morgan Kaufmann,
  282--289.
\newblock


\bibitem[\protect\citeauthoryear{Liao, Fox, and Kautz}{Liao
  et~al\mbox{.}}{2005}]%
        {liao2007hierarchical}
{Lin Liao}, {Dieter Fox}, {and} {Henry~A. Kautz}. 2005.
\newblock \showarticletitle{{Hierarchical Conditional Random Fields for
  GPS-Based Activity Recognition}}. In {\em International Symposium of Robotics
  Research (ISRR)} {\em (Springer Tracts in Advanced Robotics (STAR))},
  Vol.~28. Springer, 487--506.
\newblock


\bibitem[\protect\citeauthoryear{Lifschitz}{Lifschitz}{1994}]%
        {lifschitz1994}
{Vladimir Lifschitz}. 1994.
\newblock \showarticletitle{Circumscription}.
\newblock In {\em {Handbook of logic in Artificial Intelligence and Logic
  Programming}}. Vol.~3. {Oxford University Press, Inc.}, 297--352.
\newblock


\bibitem[\protect\citeauthoryear{Lowd and Domingos}{Lowd and Domingos}{2007}]%
        {lowd2007efficient}
{Daniel Lowd} {and} {Pedro Domingos}. 2007.
\newblock \showarticletitle{{Efficient Weight Learning for Markov Logic
  Networks}}. In {\em Proceedings of the 11th European Conference on Principles
  and Practice of Knowledge Discovery in Databases (PKDD)} {\em (Lecture Notes
  in Computer Science)}, Vol. 4702. Springer, 200--211.
\newblock


\bibitem[\protect\citeauthoryear{Luckham}{Luckham}{2002}]%
        {luckham2002power}
{David~C. Luckham}. 2002.
\newblock {\em {The Power of Events: An Introduction to Complex Event
  Processing in Distributed Enterprise Systems}}.
\newblock Addison-Wesley Longman Publishing Co., Inc.
\newblock


\bibitem[\protect\citeauthoryear{Luckham}{Luckham}{2011}]%
        {LuckhamEPB}
{David~C. Luckham}. 2011.
\newblock {\em {Event Processing for Business: Organizing the Real-Time
  Enterprise}}.
\newblock Wiley.
\newblock
\showISBNx{9781118171851}
\showLCCN{2011029933}


\bibitem[\protect\citeauthoryear{Manfredotti}{Manfredotti}{2009}]%
        {manfredotti2009RBN}
{Cristina Manfredotti}. 2009.
\newblock \showarticletitle{{Modeling and Inference with Relational Dynamic
  Bayesian Networks}}.
\newblock In {\em Advances in Artificial Intelligence}, {Yong Gao} {and}
  {Nathalie Japkowicz} (Eds.). Lecture Notes in Computer Science, Vol. 5549.
  Springer Berlin / Heidelberg, 287--290.
\newblock
\showISBNx{978-3-642-01817-6}


\bibitem[\protect\citeauthoryear{Manfredotti, Hamilton, and Zilles}{Manfredotti
  et~al\mbox{.}}{2010}]%
        {manfredotti2010RBN}
{Cristina Manfredotti}, {Howard Hamilton}, {and} {Sandra Zilles}. 2010.
\newblock \showarticletitle{{Learning RDBNs for Activity Recognition}}. In {\em
  NIPS Workshop on Learning and Planning from Batch Time Series Data}.
\newblock


\bibitem[\protect\citeauthoryear{Mateus, Pacheco, Pinto, Sernadas, and
  Sernadas}{Mateus et~al\mbox{.}}{2001}]%
        {mateus2001probabilistic}
{Paulo Mateus}, {Ant{\'o}nio Pacheco}, {Javier Pinto}, {Am{\'\i}lcar Sernadas},
  {and} {Cristina Sernadas}. 2001.
\newblock \showarticletitle{{Probabilistic Situation Calculus}}.
\newblock {\em {Annals of Mathematics and Artificial Intelligence}\/} {32}, 1
  (2001), 393--431.
\newblock


\bibitem[\protect\citeauthoryear{McCarthy}{McCarthy}{1980}]%
        {mccarthy1980circ}
{John McCarthy}. 1980.
\newblock \showarticletitle{{Circumscription - A Form of Non-Monotonic
  Reasoning}}.
\newblock {\em {Artificial Intelligence}\/} {13}, 1-2 (1980), 27--39.
\newblock


\bibitem[\protect\citeauthoryear{McCarthy and Hayes}{McCarthy and
  Hayes}{1968}]%
        {mccarthy1968some}
{John McCarthy} {and} {Patrik~J. Hayes}. 1968.
\newblock {\em {Some philosophical problems from the standpoint of artificial
  intelligence}}.
\newblock Stanford University.
\newblock


\bibitem[\protect\citeauthoryear{Miller and Shanahan}{Miller and
  Shanahan}{2002}]%
        {miller2002some}
{Rob Miller} {and} {Murray Shanahan}. 2002.
\newblock \showarticletitle{{Some Alternative Formulations of the Event
  Calculus}}. In {\em Computational Logic: Logic Programming and Beyond, Essays
  in Honour of Robert A. Kowalski, Part II} {\em (Lecture Notes in Computer
  Science)}. Springer, 452--490.
\newblock


\bibitem[\protect\citeauthoryear{Minka}{Minka}{2005}]%
        {minka2005discriminative}
{Tom Minka}. 2005.
\newblock {\em {Discriminative models, not discriminative training}}.
\newblock {T}echnical {R}eport. Microsoft Research.
\newblock
\newblock
\shownote{Available at:
  \url{http://research.microsoft.com/pubs/70229/tr-2005-144.pdf}.}


\bibitem[\protect\citeauthoryear{Morariu and Davis}{Morariu and Davis}{2011}]%
        {morariu11}
{Vlad~I. Morariu} {and} {Larry~S. Davis}. 2011.
\newblock \showarticletitle{{Multi-agent event recognition in structured
  scenarios}}. In {\em Proceedings of the Conference on Computer Vision and
  Pattern Recognition (CVPR)}. IEEE Computer Society, 3289--3296.
\newblock


\bibitem[\protect\citeauthoryear{Mueller}{Mueller}{2006}]%
        {mueller2006commonsense}
{Erik~T. Mueller}. 2006.
\newblock {\em {Commonsense Reasoning}}.
\newblock Morgan Kaufmann.
\newblock


\bibitem[\protect\citeauthoryear{Mueller}{Mueller}{2008}]%
        {mueller2008event}
{Erik~T. Mueller}. 2008.
\newblock \showarticletitle{{Event Calculus}}.
\newblock In {\em Handbook of Knowledge Representation}. Foundations of
  Artificial Intelligence, Vol.~3. Elsevier, 671--708.
\newblock


\bibitem[\protect\citeauthoryear{Murphy}{Murphy}{2002}]%
        {murphy2002DBN}
{Kevin~P. Murphy}. 2002.
\newblock {\em {Dynamic Bayesian Networks: representation, inference and
  learning}}.
\newblock Ph.D. Dissertation. University of California.
\newblock


\bibitem[\protect\citeauthoryear{Natarajan and Nevatia}{Natarajan and
  Nevatia}{2007}]%
        {natarajan2007hierarchical}
{Pradeep Natarajan} {and} {Ramakant Nevatia}. 2007.
\newblock \showarticletitle{{Hierarchical Multi-channel Hidden Semi Markov
  Models}}. In {\em Proceedings of the 20th International Joint Conference on
  Artificial Intelligence (IJCAI)}. 2562--2567.
\newblock


\bibitem[\protect\citeauthoryear{Natarajan, Bui, Tadepalli, Kersting, and
  Wong}{Natarajan et~al\mbox{.}}{2008}]%
        {natarajan2008LoHiHMM}
{Sriraam Natarajan}, {Hung~Hai Bui}, {Prasad Tadepalli}, {Kristian Kersting},
  {and} {Weng-Keen Wong}. 2008.
\newblock \showarticletitle{{Logical Hierarchical Hidden Markov Models for
  Modeling User Activities}}. In {\em Proceedings of the 18th International
  Conference Inductive Logic Programming (ILP)} {\em (Lecture Notes in Computer
  Science)}, Vol. 5194. Springer, 192--209.
\newblock


\bibitem[\protect\citeauthoryear{Paschke and Kozlenkov}{Paschke and
  Kozlenkov}{2009}]%
        {paschke2009rule}
{Adrian Paschke} {and} {Alexander Kozlenkov}. 2009.
\newblock \showarticletitle{{Rule-Based Event Processing and Reaction Rules}}.
  In {\em Proceedings of the 3rd International Symposium on Rules (RuleML)}
  {\em (Lecture Notes in Computer Science)}, Vol. 5858. Springer, 53--66.
\newblock


\bibitem[\protect\citeauthoryear{Pinto, Sernadas, Sernadas, and Mateus}{Pinto
  et~al\mbox{.}}{2000}]%
        {pinto2000non}
{J. Pinto}, {A. Sernadas}, {C. Sernadas}, {and} {P. Mateus}. 2000.
\newblock \showarticletitle{{Non-determinism and uncertainty in the Situation
  Calculus}}.
\newblock {\em {International Journal of Uncertainty, Fuzziness and
  Knowledge-Based Systems}\/} {8}, 2 (2000), 127--149.
\newblock


\bibitem[\protect\citeauthoryear{Poon and Domingos}{Poon and Domingos}{2006}]%
        {poon2006sound}
{Hoifung Poon} {and} {Pedro Domingos}. 2006.
\newblock \showarticletitle{{Sound and Efficient Inference with Probabilistic
  and Deterministic Dependencies}}. In {\em Proceedings of the 21st AAAI
  Conference on Artificial Intelligence}. AAAI Press, 458--463.
\newblock


\bibitem[\protect\citeauthoryear{Rabiner and Juang}{Rabiner and Juang}{1986}]%
        {rabiner1986HMM}
{Lawrence~R. Rabiner} {and} {Biing-Hwang Juang}. 1986.
\newblock \showarticletitle{{An introduction to Hidden Markov Models}}.
\newblock {\em {Acoustics, Speech, and Signal Processing Magazine (ASSP)}\/}
  {3}, 1 (1986), 4--16.
\newblock


\bibitem[\protect\citeauthoryear{Raedt and Kersting}{Raedt and
  Kersting}{2010}]%
        {RaedtK10}
{Luc~De Raedt} {and} {Kristian Kersting}. 2010.
\newblock \showarticletitle{{Statistical Relational Learning}}.
\newblock In {\em Encyclopedia of Machine Learning}, {Claude Sammut} {and}
  {Geoffrey~I. Webb} (Eds.). Springer, 916--924.
\newblock
\showISBNx{978-0-387-30768-8}


\bibitem[\protect\citeauthoryear{Reiter}{Reiter}{2001}]%
        {reiter2001knowledge}
{Raymond Reiter}. 2001.
\newblock {\em {Knowledge in Action: Logical Foundations for Specifying and
  Implementing Dynamical Systems}}.
\newblock MIT Press.
\newblock


\bibitem[\protect\citeauthoryear{Riedel}{Riedel}{2008}]%
        {riedel2008improving}
{Sebastian Riedel}. 2008.
\newblock \showarticletitle{{Improving the Accuracy and Efficiency of MAP
  Inference for Markov Logic}}. In {\em Proceedings of the 24th Conference in
  Uncertainty in Artificial Intelligence (UAI)}. AUAI Press, 468--475.
\newblock


\bibitem[\protect\citeauthoryear{Sadilek and Kautz}{Sadilek and Kautz}{2012}]%
        {sadilek2012}
{Adam Sadilek} {and} {Henry~A. Kautz}. 2012.
\newblock \showarticletitle{{Location-Based Reasoning about Complex Multi-Agent
  Behavior}}.
\newblock {\em {Journal of Artificial Intelligence Research (JAIR)}\/}  {43}
  (2012), 87--133.
\newblock


\bibitem[\protect\citeauthoryear{Schiffel and Thielscher}{Schiffel and
  Thielscher}{2006}]%
        {schiffel2006reconciling}
{Stephan Schiffel} {and} {Michael Thielscher}. 2006.
\newblock \showarticletitle{{Reconciling Situation Calculus and Fluent
  Calculus}}. In {\em Proceedings of the National Conference on Artificial
  Intelligence}, Vol.~21. AAAI Press, 287.
\newblock


\bibitem[\protect\citeauthoryear{Selman, Amer, Fern, and Todorovic}{Selman
  et~al\mbox{.}}{2011}]%
        {selman2011PEL}
{Joseph Selman}, {Mohamed~R. Amer}, {Alan Fern}, {and} {Sinisa Todorovic}.
  2011.
\newblock \showarticletitle{{PEL-CNF: Probabilistic event logic conjunctive
  normal form for video interpretation}}. In {\em Proceedings of the
  International Conference on Computer Vision Workshops (ICCVW)}. IEEE Computer
  Society, 680--687.
\newblock


\bibitem[\protect\citeauthoryear{Shanahan}{Shanahan}{1997}]%
        {shanahan1997solving}
{Murray Shanahan}. 1997.
\newblock {\em {Solving the Frame Problem: A Mathematical Investigation of the
  Common Sense Law of Inertia}}.
\newblock MIT Press.
\newblock


\bibitem[\protect\citeauthoryear{Shanahan}{Shanahan}{1999}]%
        {shanahan1999event}
{Murray Shanahan}. 1999.
\newblock \showarticletitle{{The Event Calculus Explained}}.
\newblock In {\em Artificial Intelligence Today}, {Michael Wooldridge} {and}
  {Manuela Veloso} (Eds.). Lecture Notes in Computer Science, Vol. 1600.
  Springer, 409--430.
\newblock


\bibitem[\protect\citeauthoryear{Shavlik and Natarajan}{Shavlik and
  Natarajan}{2009}]%
        {shavlik2009speeding}
{Jude~W. Shavlik} {and} {Sriraam Natarajan}. 2009.
\newblock \showarticletitle{Speeding Up Inference in Markov Logic Networks by
  Preprocessing to Reduce the Size of the Resulting Grounded Network}. In {\em
  Proceedings of the 21st International Joint Conference on Artificial
  Intelligence (IJCAI)}. 1951--1956.
\newblock


\bibitem[\protect\citeauthoryear{Shet, Neumann, Ramesh, and Davis}{Shet
  et~al\mbox{.}}{2007}]%
        {ShetNRD07}
{Vinay~D. Shet}, {Jan Neumann}, {Visvanathan Ramesh}, {and} {Larry~S. Davis}.
  2007.
\newblock \showarticletitle{{Bilattice-based Logical Reasoning for Human
  Detection}}. In {\em Proceedings of the Conference on Computer Vision and
  Pattern Recognition (CVPR)}. IEEE Computer Society, 1--8.
\newblock


\bibitem[\protect\citeauthoryear{Singla and Domingos}{Singla and
  Domingos}{2005}]%
        {singla2005discriminative}
{Parag Singla} {and} {Pedro Domingos}. 2005.
\newblock \showarticletitle{{Discriminative Training of Markov Logic
  Networks}}. In {\em Proceedings of the 20th National Conference on Artificial
  Intelligence}. AAAI Press / The MIT Press, 868--873.
\newblock


\bibitem[\protect\citeauthoryear{Singla and Domingos}{Singla and
  Domingos}{2006}]%
        {singla2006memory}
{Parag Singla} {and} {Pedro Domingos}. 2006.
\newblock \showarticletitle{{Memory-Efficient Inference in Relational
  Domains}}. In {\em Proceedings of the 21st AAAI Conference on Artificial
  Intelligence}. AAAI Press, 488--493.
\newblock


\bibitem[\protect\citeauthoryear{Singla and Domingos}{Singla and
  Domingos}{2008}]%
        {singla2008lifted}
{Parag Singla} {and} {Pedro Domingos}. 2008.
\newblock \showarticletitle{{Lifted First-Order Belief Propagation}}. In {\em
  Proceedings of the 23rd AAAI Conference on Artificial Intelligence}. AAAI
  Press, 1094--1099.
\newblock


\bibitem[\protect\citeauthoryear{Siskind}{Siskind}{2001}]%
        {siskind2001EL}
{Jeffrey~Mark Siskind}. 2001.
\newblock \showarticletitle{{Grounding the Lexical Semantics of Verbs in Visual
  Perception using Force Dynamics and Event Logic}}.
\newblock {\em {Journal of Artificial Intelligence Research (JAIR)}\/}  {15}
  (2001), 31--90.
\newblock


\bibitem[\protect\citeauthoryear{Skarlatidis, Artikis, Filippou, and
  Paliouras}{Skarlatidis et~al\mbox{.}}{2013}]%
        {anskarl2013PLPEC}
{Anastasios Skarlatidis}, {Alexander Artikis}, {Jason Filippou}, {and}
  {Georgios Paliouras}. 2013.
\newblock \showarticletitle{{A Probabilistic Logic Programming Event
  Calculus}}.
\newblock {\em {Journal of Theory and Practice of Logic Programming (TPLP)}\/}
  (2013).
\newblock


\bibitem[\protect\citeauthoryear{Skarlatidis, Paliouras, Vouros, and
  Artikis}{Skarlatidis et~al\mbox{.}}{2011}]%
        {anskarl2011RML}
{Anastaios Skarlatidis}, {Georgios Paliouras}, {George Vouros}, {and}
  {Alexander Artikis}. 2011.
\newblock \showarticletitle{{Probabilistic Event Calculus Based on Markov Logic
  Networks}}. In {\em Proceedings of the 5th International Symposium on Rules
  (RuleML)} {\em (Lecture Notes in Computer Science)}, Vol. 7018. Springer,
  155--170.
\newblock


\bibitem[\protect\citeauthoryear{Sutton and McCallum}{Sutton and
  McCallum}{2007}]%
        {sutton2006introduction}
{Charles Sutton} {and} {Andrew McCallum}. 2007.
\newblock \showarticletitle{{An Introduction to Conditional Random Fields for
  Relational Learning}}.
\newblock In {\em Introduction to Statistical Relational Learning}, {Lise
  Getoor} {and} {Ben Taskar} (Eds.). MIT Press, 93--127.
\newblock


\bibitem[\protect\citeauthoryear{Thielscher}{Thielscher}{1999}]%
        {thielscher1999situation}
{Michael Thielscher}. 1999.
\newblock \showarticletitle{{From Situation Calculus to Fluent Calculus: State
  update axioms as a solution to the inferential frame problem}}.
\newblock {\em {Artificial intelligence}\/} {111}, 1 (1999), 277--299.
\newblock


\bibitem[\protect\citeauthoryear{Thielscher}{Thielscher}{2001}]%
        {thielscher2001qualification}
{Michael Thielscher}. 2001.
\newblock \showarticletitle{{The qualification problem: A solution to the
  problem of anomalous models}}.
\newblock {\em {Artificial Intelligence}\/} {131}, 1 (2001), 1--37.
\newblock


\bibitem[\protect\citeauthoryear{Tran and Davis}{Tran and Davis}{2008}]%
        {TranD08}
{Son~Dinh Tran} {and} {Larry~S. Davis}. 2008.
\newblock \showarticletitle{{Event Modeling and Recognition Using Markov Logic
  Networks}}. In {\em Proceedings of the 10th European Conference on Computer
  Vision (ECCV)} {\em (Lecture Notes in Computer Science)}, Vol. 5303.
  Springer, 610--623.
\newblock


\bibitem[\protect\citeauthoryear{Vail, Veloso, and Lafferty}{Vail
  et~al\mbox{.}}{2007}]%
        {vail2007conditional}
{Douglas~L. Vail}, {Manuela~M. Veloso}, {and} {John~D. Lafferty}. 2007.
\newblock \showarticletitle{{Conditional Random Fields for Activity
  Recognition}}. In {\em Proceedings of the 6th International Joint Conference
  on Autonomous Agents and Multiagent Systems (AAMAS)}. IFAAMAS, 1331--1338.
\newblock


\bibitem[\protect\citeauthoryear{Van~Belleghem, Denecker, and
  De~Schreye}{Van~Belleghem et~al\mbox{.}}{1997}]%
        {van1997relation}
{K. Van~Belleghem}, {M. Denecker}, {and} {D. De~Schreye}. 1997.
\newblock \showarticletitle{{On the relation between Situation Calculus and
  Event Calculus}}.
\newblock {\em {The Journal of Logic Programming}\/} {31}, 1 (1997), 3--37.
\newblock


\bibitem[\protect\citeauthoryear{Wang and Domingos}{Wang and Domingos}{2008}]%
        {wang2008hybrid}
{Jue Wang} {and} {Pedro Domingos}. 2008.
\newblock \showarticletitle{{Hybrid Markov Logic Networks}}. In {\em
  Proceedings of the 23rd AAAI Conference on Artificial Intelligence}. AAAI
  Press, 1106--1111.
\newblock


\bibitem[\protect\citeauthoryear{Wu, Lian, and Hsu}{Wu et~al\mbox{.}}{2007}]%
        {wu2007joint}
{Tsu-yu Wu}, {Chia-chun Lian}, {and} {Jane Yung-jen Hsu}. 2007.
\newblock \showarticletitle{{Joint Recognition of Multiple Concurrent
  Activities using Factorial Conditional Random Fields}}.
\newblock In {\em Proceedings of the Workshop on Plan, Activity, and Intent
  Recognition (PAIR)}. AAAI Press, 82--88.
\newblock


\end{thebibliography}

\end{document}